\documentclass[lettersize,journal]{IEEEtran}

\usepackage{graphicx}
\usepackage{amsmath}
\usepackage{amssymb}
\usepackage{amsthm}
\usepackage{bbm}
\usepackage{subfigure}
\usepackage{times}
\usepackage{latexsym}
\usepackage{microtype}
\usepackage{color}
\usepackage{float}
\usepackage{tabularx,multirow}
\usepackage{hyperref}
\usepackage{breakurl}
\makeatletter
\def\UrlAlphabet{%
      \do\a\do\b\do\c\do\d\do\e\do\f\do\g\do\h\do\i\do\j%
      \do\k\do\l\do\m\do\n\do\o\do\p\do\q\do\r\do\s\do\t%
      \do\u\do\v\do\w\do\x\do\y\do\z\do\A\do\B\do\C\do\D%
      \do\E\do\F\do\G\do\H\do\I\do\J\do\K\do\L\do\M\do\N%
      \do\O\do\P\do\Q\do\R\do\S\do\T\do\U\do\V\do\W\do\X%
      \do\Y\do\Z}
\def\UrlDigits{\do\1\do\2\do\3\do\4\do\5\do\6\do\7\do\8\do\9\do\0}
\g@addto@macro{\UrlBreaks}{\UrlOrds}
\g@addto@macro{\UrlBreaks}{\UrlAlphabet}
\g@addto@macro{\UrlBreaks}{\UrlDigits}

\usepackage{makecell}
\usepackage{booktabs}

\usepackage{bm}

\usepackage{ulem}

\usepackage{multirow}
\usepackage{multicol}
\usepackage{array}

\usepackage{amsmath}
  \usepackage{multirow}
  \usepackage{array}
  \usepackage{algorithm}
  \usepackage{algorithmic}
  \usepackage[table]{xcolor}
  \usepackage{makecell}
  
\usepackage{graphicx}
\usepackage{subfigure}
 
\usepackage{float}  % forge figure position
\usepackage{textcomp}
\usepackage{booktabs}
\usepackage{diagbox}  
\usepackage{bbm}
\usepackage{verbatim}

\ifCLASSOPTIONcompsoc
  \usepackage[nocompress]{cite}
\else
  \usepackage{cite}
\fi
\ifCLASSINFOpdf
\else
\fi
\newtheorem{definition}{Definition}[section]

\newtheorem{theorem}{Theorem}[section]
\newtheorem{assumption}{Assumption}[section]
\newtheorem{corollary}{Corollary}[section]

\begin{document}

%%%%%%%%%%%%%%%%%%%%%%%%%% Title %%%%%%%%%%%%%%%%%%%%%%%%%%%%%%%%%

\title{Neuromodulated Meta-Learning}

%%%%%%%%%%%%%%%%%%%%%%%%%% Authors %%%%%%%%%%%%%%%%%%%%%%%%%%%%%%%%%
\author{Jingyao~Wang,
           Huijie~Guo,
           Wenwen~Qiang,
           Jiangmeng~Li,
           Changwen~Zheng,
        Hui~Xiong,~\IEEEmembership{Fellow,~IEEE},
Gang~Hua,~\IEEEmembership{Fellow,~IEEE}      
\IEEEcompsocitemizethanks{\IEEEcompsocthanksitem J. Wang, W. Qiang, J. Li, and C. Zheng are with the University of Chinese Academy of Sciences, Beijing, China. They are also with the National Key Laboratory of Space Integrated Information System, Institute of Software Chinese Academy of Sciences, Beijing, China. E-mail: \{wangjingyao2023, qiangwenwen, jiangmeng2019, changwen\}@iscas.ac.cn.
\IEEEcompsocthanksitem H. Guo is with
              Beijing Advanced Innovation Center for Big Data and Brain Computing, Beihang University, Beijing, China. Email: guo\_hj@buaa.edu.cn            
\IEEEcompsocthanksitem H. Xiong is with the Thrust of Artificial Intelligence, Hong Kong University of Science and Technology, Guangzhou, China. He is also with the
              Department of Computer Science and Engineering, the Hong Kong University of Science and Technology, Hong Kong SAR, China. E-mail: xionghui@ust.hk.
\IEEEcompsocthanksitem G. Hua is with Wormpex AI Research
 LLC, Bellevue, WA 98004 USA. E-mail: ganghua@gmail.com.
\IEEEcompsocthanksitem Jingyao Wang and Huijie Guo have contributed equally to this work. Corresponding author: Wenwen Qiang
              
}
%\thanks{An earlier version of this work was published in \cite{sun-etal-2021-inconsistency-matters}}
}

% The paper headers
% The paper headers
\markboth{IEEE TRANSACTIONS ON PATTERN ANALYSIS AND MACHINE INTELLIGENCE}{Wang \MakeLowercase{\textit{et al.}}: Neuromodulated Meta-Learning}

% \IEEEpubid{0000--0000/00\$00.00~\copyright~2021 IEEE}
% % Remember, if you use this you must call \IEEEpubidadjcol in the second
% % column for its text to clear the IEEEpubid mark.

\IEEEtitleabstractindextext{%
\begin{abstract}
Humans excel at adapting perceptions and actions to diverse environments, enabling efficient interaction with the external world. This adaptive capability relies on the biological nervous system (BNS), which activates different brain regions for distinct tasks. Meta-learning similarly trains machines to handle multiple tasks but relies on a fixed network structure, not as flexible as BNS. To investigate the role of flexible network structure (FNS) in meta-learning, we conduct extensive empirical and theoretical analyses, finding that model performance is tied to structure, with no universally optimal pattern across tasks. This reveals the crucial role of FNS in meta-learning, ensuring meta-learning to generate the optimal structure for each task, thereby maximizing the performance and learning efficiency of meta-learning. Motivated by this insight, we propose to define, measure, and model FNS in meta-learning. First, we define that an effective FNS should possess frugality, plasticity, and sensitivity. Then, to quantify FNS in practice, we present three measurements for these properties, collectively forming the \emph{structure constraint} with theoretical supports. Building on this, we finally propose Neuromodulated Meta-Learning (NeuronML) to model FNS in meta-learning. It utilizes bi-level optimization to update both weights and structure with the structure constraint. Extensive theoretical and empirical evaluations demonstrate the effectiveness of NeuronML on various tasks. Code is publicly available at \href{https://github.com/WangJingyao07/NeuronML}{https://github.com/WangJingyao07/NeuronML}.
\end{abstract}

% Note that keywords are not normally used for peerreview papers.
}

% make the title area
\maketitle

% To allow for easy dual compilation without having to reenter the
% abstract/keywords data, the \IEEEtitleabstractindextext text will
% not be used in maketitle, but will appear (i.e., to be "transported")
% here as \IEEEdisplaynontitleabstractindextext when the compsoc 
% or transmag modes are not selected <OR> if conference mode is selected 
% - because all conference papers position the abstract like regular
% papers do.
\IEEEdisplaynontitleabstractindextext
% \IEEEdisplaynontitleabstractindextext has no effect when using
% compsoc or transmag under a non-conference mode.

% For peer review papers, you can put extra information on the cover
% page as needed:
% \ifCLASSOPTIONpeerreview
% \begin{center} \bfseries EDICS Category: 3-BBND \end{center}
% \fi
%
% For peerreview papers, this IEEEtran command inserts a page break and
% creates the second title. It will be ignored for other modes.
\IEEEpeerreviewmaketitle

%%
%% This is file `sample-sigconf.tex',
%% generated with the docstrip utility.
%%
%% The original source files were:
%%
%% samples.dtx  (with options: `sigconf')
%% 
%% IMPORTANT NOTICE:
%% 
%% For the copyright see the source file.
%% 
%% Any modified versions of this file must be renamed
%% with new filenames distinct from sample-sigconf.tex.
%% 
%% For distribution of the original source see the terms
%% for copying and modification in the file samples.dtx.
%% 
%% This generated file may be distributed as long as the
%% original source files, as listed above, are part of the
%% same distribution. (The sources need not necessarily be
%% in the same archive or directory.)
%%
%%
%% Commands for TeXCount
%TC:macro \cite [option:text,text]
%TC:macro \citep [option:text,text]
%TC:macro \citet [option:text,text]
%TC:envir table 0 1
%TC:envir table* 0 1
%TC:envir tabular [ignore] word
%TC:envir displaymath 0 word
%TC:envir math 0 word
%TC:envir comment 0 0
%%
%%
%% The first command in your LaTeX source must be the \documentclass
%% command.
%%
%% For submission and review of your manuscript please change the
%% command to \documentclass[manuscript, screen, review]{acmart}.
%%
%% When submitting camera ready or to TAPS, please change the command
%% to \documentclass[sigconf]{acmart} or whichever template is required
%% for your publication.
%%
%%

\section{Introduction}
\label{sec:1}

\begin{figure*}[t]
% \vskip 0.2in
\begin{center}
\centerline{\includegraphics[width=\textwidth]{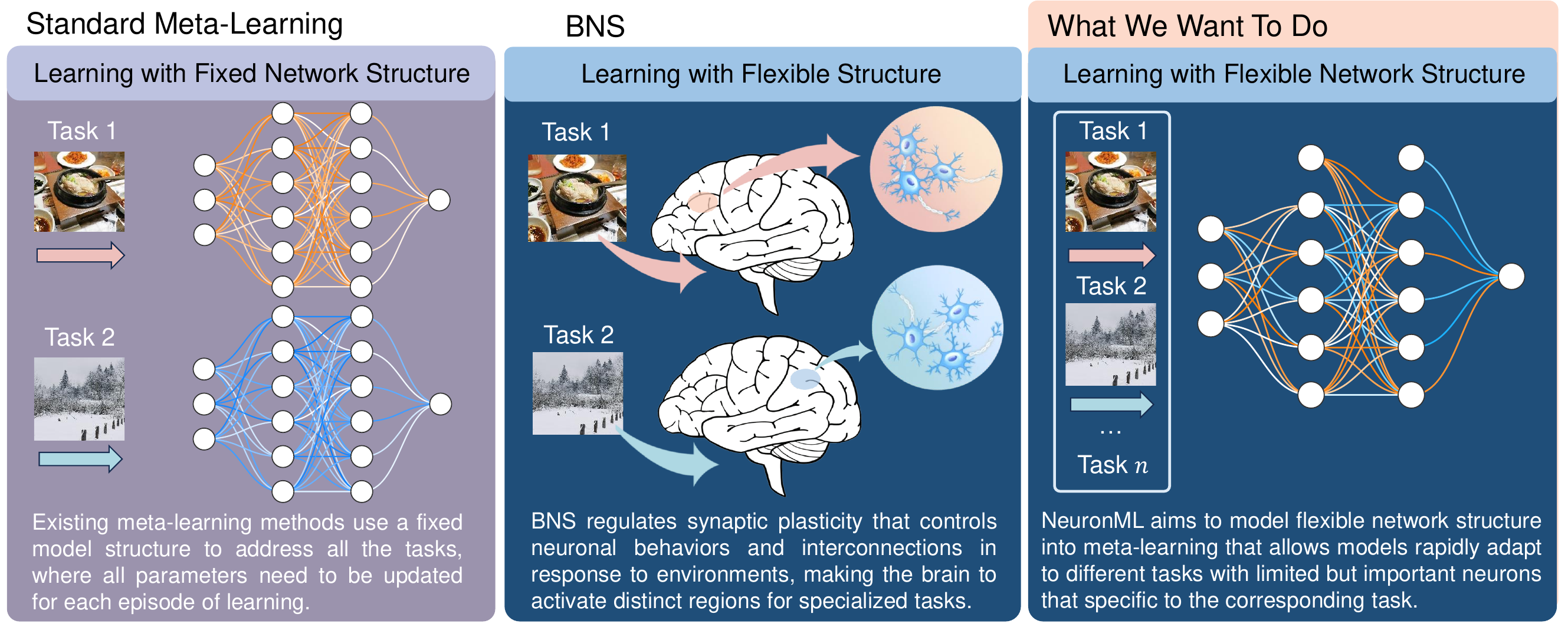}}
%%\vspace{-0.1in}
\caption{Existing meta-learning models perform learning through fixed model structures, where all parameters need to be updated for each episode of learning. In contrast, human cognition of the world is achieved by adjusting the behavior of neurons, which only needs part of the neurons while adjusting the activation area of the brain according to the downstream task. We ask if neuromodulated meta-learning can understand the world with flexible network structures.}
\label{fig:intro}
\end{center}
%%\vspace{-0.1in}
\end{figure*}

\IEEEPARstart{T}{he} adaptability of the human brain has long been associated with the biological nervous system (BNS) \cite{marder2012neuromodulation,bear2020neuroscience,de2021future}. The BNS regulates biological neural networks by modulating synaptic plasticity \cite{brines2005emerging,avery2017neuromodulatory}, which controls neuronal input/output behaviors and interconnections in response to environmental stimuli. This mechanism enables the brain to activate different regions for different tasks \cite{wang2023hacking,bateni2020improved,cesa2006prediction,miconi2023learning,wang2023amsa}. 
Thus, we conclude that BNS can be viewed as a global prior for human brain adaptability. 
In parallel, meta-learning \cite{finn2019online,hospedales2021meta,protonet,reptile}, often referred to as ``learning to learn'', involves (i) learning a global prior, i.e., a model initialization, from diverse tasks, and then (ii) fine-tuning the model on unseen tasks to achieve desired performance.
Comparing meta-learning with BNS, they both can be viewed as relying on a global prior to solve different tasks, i.e., for the human brain, the global prior is BNS, while for meta-learning, the global prior is the model initialization.

While meta-learning and BNS have the above connection, they still have differences when dealing with different tasks. 
Specifically, meta-learning employs a fixed model structure to solve all tasks, whereas BNS activates different brain regions for different tasks (\textbf{Figure \ref{fig:intro}}). Consequently, meta-learning lacks flexible network structure (FNS).
Compared to a fixed structure, FNS that is adjusted depending on the task may provide the following benefits: (i) alleviate overfitting caused by over-parameterization \cite{chen2022understanding,yao2021improving,wang2024rethinking}; (ii) reduce the computational overhead caused by redundant neurons \cite{lee2021meta,schwarz2022meta,chin2020towards,buciluǎ2006model}.
Thus, this prompts a key question: \textbf{is FNS crucial for meta-learning, e.g., further unleash the potential of meta-learning?}

\begin{figure*}
\begin{center}
     \subfigure[miniImagenet]{\includegraphics[width=0.246\textwidth]{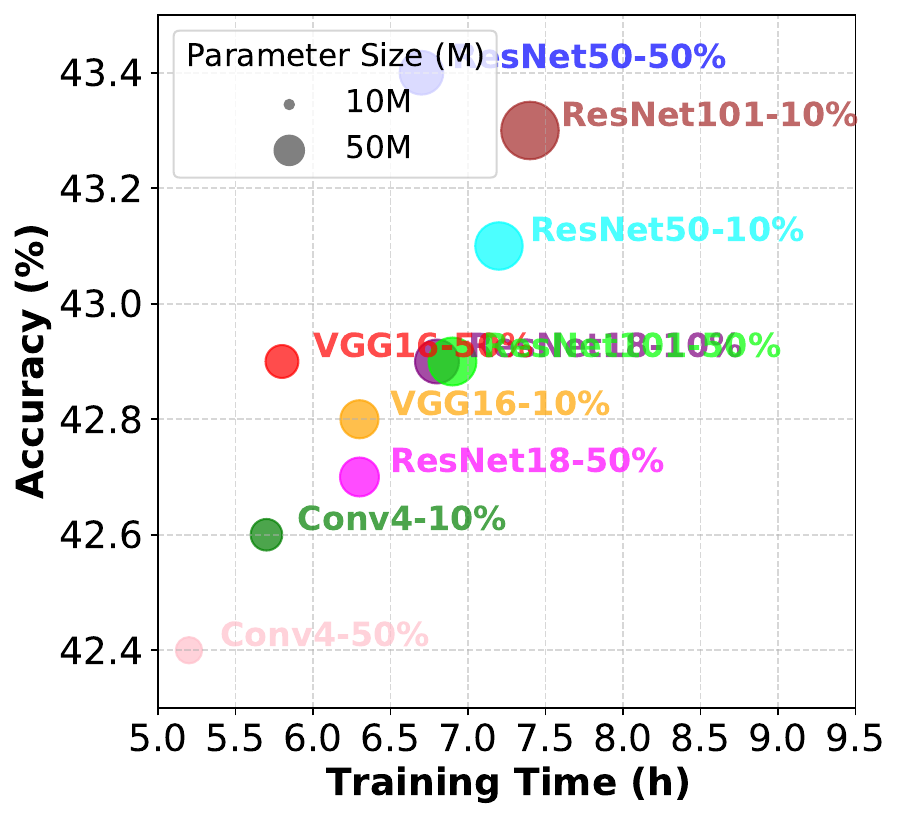}\label{fig:motivation_miniimagenet}}
     \subfigure[Omniglot]{\includegraphics[width=0.24\textwidth]{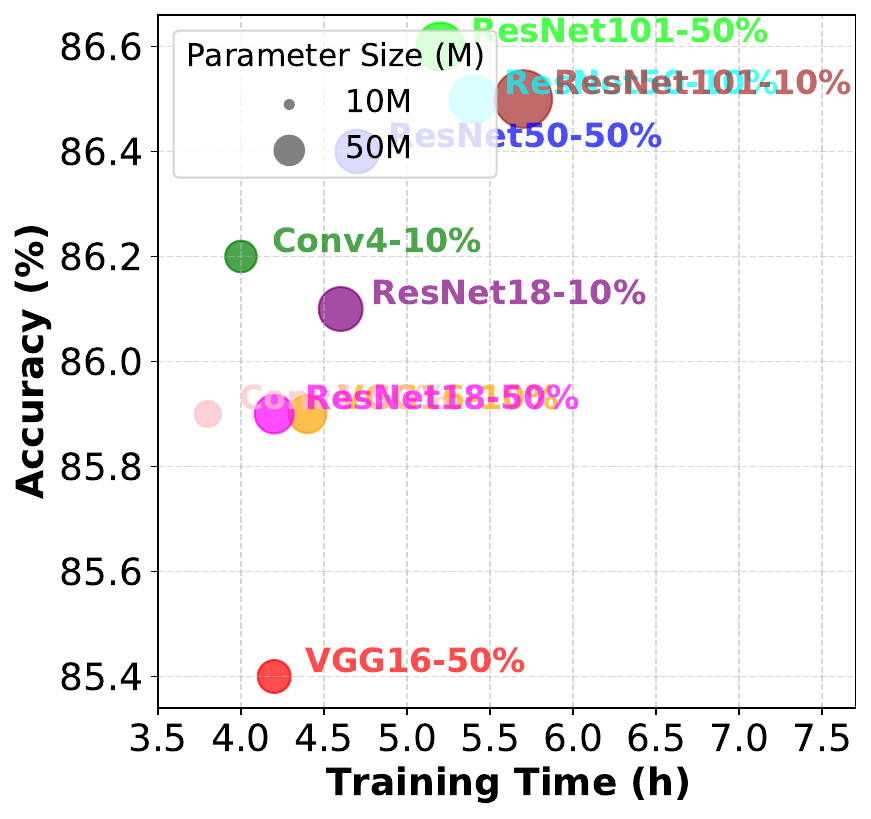}\label{fig:motivation_omniglot}}
     \subfigure[tieredImagenet]{\includegraphics[width=0.246\textwidth]{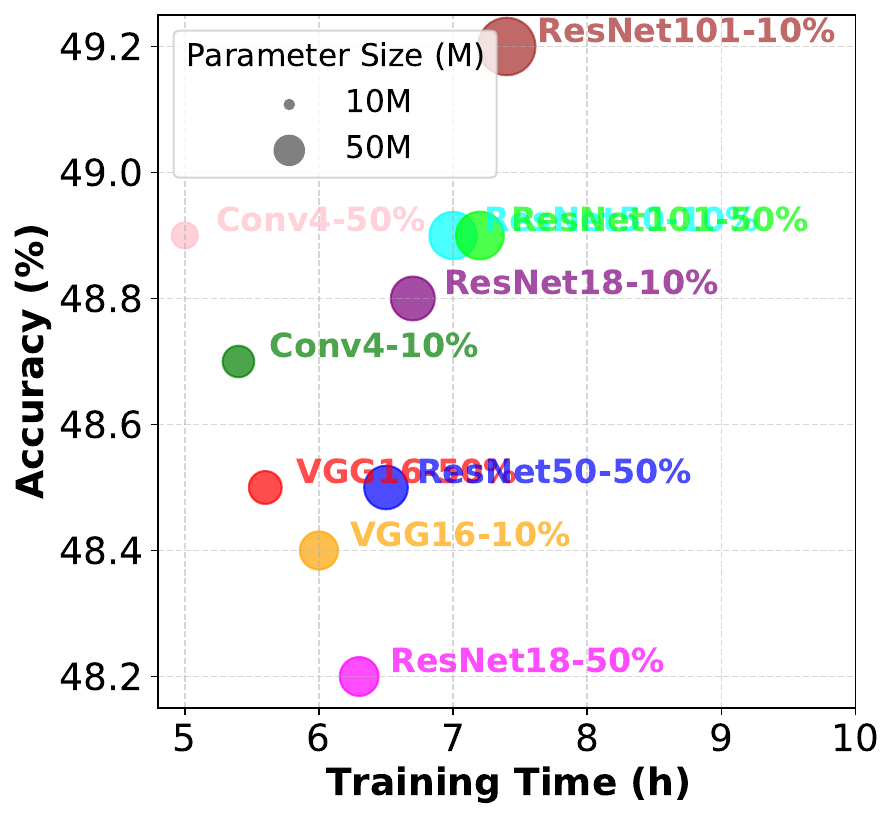}\label{fig:motivation_tieredimagenet}}
     \subfigure[CIFAR-FS]{\includegraphics[width=0.25\textwidth]{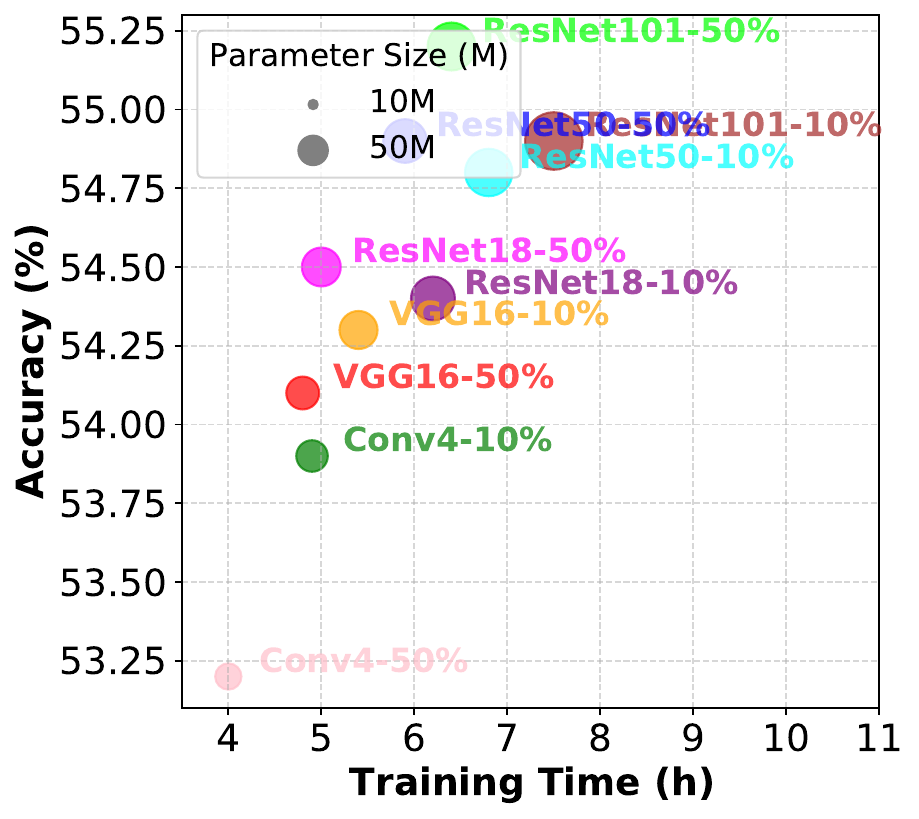}\label{fig:motivation_cifar-fs}}
\caption{Trade-off performance of MAML with different model structures on four benchmark datasets, i.e., miniImagenet (5-way 1-shot), Omniglot (20-way 1-shot), tieredImagenet (5-way 1-shot), and CIFAR-FS (5-way 1-shot). The horizontal axis represents the training time (hours), and the vertical axis represents the accuracy of each dataset. The area of the circle represents the model size of MAML with the corresponding model structure.}
\label{fig:motivation} 
\end{center}
\end{figure*}

To answer this question, we conduct both empirical and theoretical evaluations. 
Firstly, we conduct experiments with varying model structures (\textbf{Subsection \ref{sec:3.1}}). We select four benchmark datasets for evaluation: miniImagenet \cite{miniImagenet}, Omniglot \cite{Omniglot}, tieredImagenet \cite{tieredImagenet}, and CIFAR-FS \cite{CIFAR-FS}. Then, we construct ten models using five different backbones, i.e., Conv4 \cite{wang2023hacking}, VGG16 \cite{vgg16resnet50}, ResNet18 \cite{vyas2020learning}, ResNet50 \cite{vgg16resnet50}, and ResNet101 \cite{conv}, and two meta-learning strategies, i.e., MAML \cite{maml} and ProtoNet \cite{protonet}. All the backbones in different models are combined with the same classification head. Next, we adopt dropout in each model at varying levels (10\%, 30\%, and 50\%). Finally, we record the performance and training time of each model on different datasets.
The results in \textbf{Figure \ref{fig:motivation}} and \textbf{Table \ref{tab:motivation}} reveal that the optimal model with highest performance depends on the structure, without consistent pattern across datasets. 
Secondly, we conduct theoretical analyses to evaluate the relation between model structure and performance (\textbf{Subsection \ref{sec:3.2}}). 
We first show that finding the optimal model with a fixed network structure is difficult across different task distributions. Additionally, we demonstrate that the probability of selecting the optimal model from a set of meta-learning candidates depends on its structure.
Thus, from empirical and theoretical evidence, we can conclude that \textbf{FNS is crucial for meta-learning}. Based on this insight, this paper focuses on \textbf{how to define, measure, and model FNS in meta-learning.}

To address this issue, we first answer what an effective FNS is and provide its definition (\textbf{Subsection \ref{sec:Definition}}). We propose that a good FNS of the meta-learning model should possess: (i) frugality: activate a subset of neurons to solve the corresponding task; (ii) plasticity: activate different subsets of neurons to solve different tasks; (iii) sensitivity: activate the most sensitive (important) neurons to achieve comparable performance across tasks. 
Through frugality, the meta-learning model avoids excessive calculation and waste of resources, eliminating the overfitting problem. Through plasticity, the meta-learning model can adaptively adjust its network structure to fit diverse tasks without depending on a fixed structure. Through sensitivity, the meta-learning model can maintain comparable performance even with limited neurons, which serve as a proxy for the full set. Thus, together they determine what a good FNS is. Each of these properties is indispensable, also demonstrated by the experiments in \textbf{Subsection \ref{sec:7.4}}.

Next, we answer how to measure FNS in meta-learning  (\textbf{Subsection \ref{sec:Constraint}}). 
Specifically, based on the definition of FNS, we propose three measurements for the three properties:
(i) For frugality: we compute the $\ell_1$-norm of the parameters which reflects the ratio of activated neurons, incorporating a balance term between parameter dimension and data volume. It makes the model activate as few neurons as possible while maintaining the balance with the data volume.
(ii) For plasticity: we evaluate the overlap of activated neurons across tasks, using the Hebbian activation rule to assess their importance. It evaluates each neuron's activation levels across tasks, constraining the model to activate distinct neurons for each task, with higher importance resulting in less overlap.
(iii) For sensitivity: we calculate each neuron's contribution to loss using sensitivity scores, ensuring that only the most essential neurons are activated to sustain performance.
Theoretical analyses demonstrate that these measurements effectively measure the properties of a high-quality FNS. 
They together form the \emph{structure constraint}.

Finally, to model FNS in meta-learning, we propose Neuromodulated Meta-Learning (NeuronML) (\textbf{Subsection \ref{sec:Method}}). 
NeuronML employs bi-level optimization to separately optimize the neuron weights and network structure of the meta-learning model. The weight optimization uses gradient updates on the original meta-learning model, while structure optimization updates a learnable mask aligned with model parameters, where each mask element represents the activation probability of a corresponding neuron. In the first level, NeuronML fixes the mask and updates the model weights over several steps for each task, refining them based on performance across tasks. In the second level, with the model weights fixed, NeuronML optimizes the structure mask using the proposed structure constraint. This process enables joint optimization of weights and structure for enhanced meta-learning performance.
Besides, we develop several NeuronML variants for different scenarios, showcasing its adaptability (\textbf{Appendix \ref{sec:5}}).
Both theoretical and empirical analyses validate the effectiveness of NeuronML.

In summary, the main contributions are as follows:
\begin{itemize}
    \item Inspired by the flexibility in BNS, we conduct extensive empirical and theoretical analyses to explore the effect of model network structure in meta-learning, finding that FNS are crucial for meta-learning (\textbf{Section \ref{sec:3}}).
    \item We are the first to define, measure, and model FNS in meta-learning. We propose that a good FNS should possess frugality, plasticity, and sensitivity. Then, we conduct three measurements for these properties with theoretical support, which together constitute the structure constraint. 
    To model FNS in meta-learning, we propose NeuronML, a method that uses bi-level optimization to optimize weights and structure with the proposed constraint. 
    It can be applied to various scenarios (\textbf{Section \ref{sec:4} and Appendix \ref{sec:5}}).
    \item Extensive theoretical and empirical evaluations validate the effectiveness of NeuronML. The results show that NeuronML consistently achieves superior performance with FNS across various tasks (\textbf{Sections \ref{sec:6} and \ref{sec:7}}).
\end{itemize}

\section{Related Work}
\label{sec:2}

Meta-learning enables models to adapt quickly to new tasks with minimal training data, avoiding the need for complete retraining. This approach is particularly useful when data is scarce or difficult to obtain \cite{vanschoren2018meta, hospedales2021meta, wang2024comprehensive}. Meta-learning methods are generally categorized into three types: optimization-based, metric-based, and Bayesian-based methods.

Optimization-based methods aim to learn optimal model initialization that quickly converge when facing new tasks. Prominent techniques include MAML \cite{maml}, Reptile \cite{reptile}, and MetaOptNet \cite{lee2019meta}. MAML shares initialization parameters across tasks and refines them through multiple gradient updates \cite{abbas2022sharp, jeong2020ood,wang2024meta}. Reptile also shares parameters but employ an approximation strategy, fine-tuning the model through iterative updates to approach optimal values. MetaOptNet focuses on selecting the right optimizer and learning rate to adapt to new tasks without directly modifying the model parameters. More recently, optimization-based methods have been extended to tackle issues such as catastrophic forgetting \cite{chi2022metafscil,javed2019meta}.

Metric-based methods learn embedding functions that map instances from different tasks into a feature space, where instances can be classified using non-parametric methods. These methods differ in how they learn the embeddings and define similarity metrics. Key contributions include Siamese Networks \cite{koch2015siamese}, Matching Networks \cite{vinyals2016matching}, Prototypical Networks \cite{protonet}, Relation Networks \cite{relationnet}, and Graph Neural Networks (GNN) \cite{hospedales2021meta}. Siamese Networks \cite{koch2015siamese,chen2021exploring,wang2024image} maximize the similarity between two augmentations of a single instance for non-parametric learning, while GNN approach meta-learning through inference on partially observed graphical models. The other techniques generally aim to establish a metric space where classification is achieved by calculating distances to prototype representations and their variants \cite{hu2022unsupervised, bartler2022mt3}.

Bayesian-based methods utilize conditional probability to drive meta-learning processes. Notable examples include CNAPs \cite{zhang2021shallow} and SCNAP \cite{bateni2020improved}, which employ a feature extractor modulated by an adaptation network that processes task-specific data. Although less prevalent than optimization-based and metric-based methods, Bayesian approaches have seen significant progress. For example, BOOM \cite{grant2018recasting} uses Gaussian processes to model objective and acquisition functions, guiding the search for optimal meta-parameters. VMGP \cite{myers2021bayesian} applies Gaussian processes with learned deep kernel and mean functions to model predictive label distributions for each task.

Note that considering the over-parameterization problems of meta-learning, some works \cite{gordon2019meta,hospedales2021meta,tian2020meta,tseng2020regularizing} proposed using dropout and network pruning for model compression, helping to reduce model complexity. 
Dropout in meta-learning \cite{finn2018meta, raghurapid, gordon2019meta} works by randomly dropping neuron connections during training. Network pruning reduces model complexity by searching and removing the redundant neurons through a series of experiments \cite{tian2020meta,hospedales2021meta}. However, dropout is limited by the randomness in neuron removal, and network pruning is constrained by the computational overhead it requires. Additionally, given the varying requirements of different tasks, they struggle to establish a dynamic, universal scheme to reduce complexity. In contrast, in this paper, we explore FNS of meta-learning. It is a more fundamental concept which makes meta-learning adaptively adjust both model weights and structure across tasks for optimal performance.

\begin{figure}
\begin{center}
    \subfigure[miniImagenet]{\includegraphics[width=0.23\textwidth]{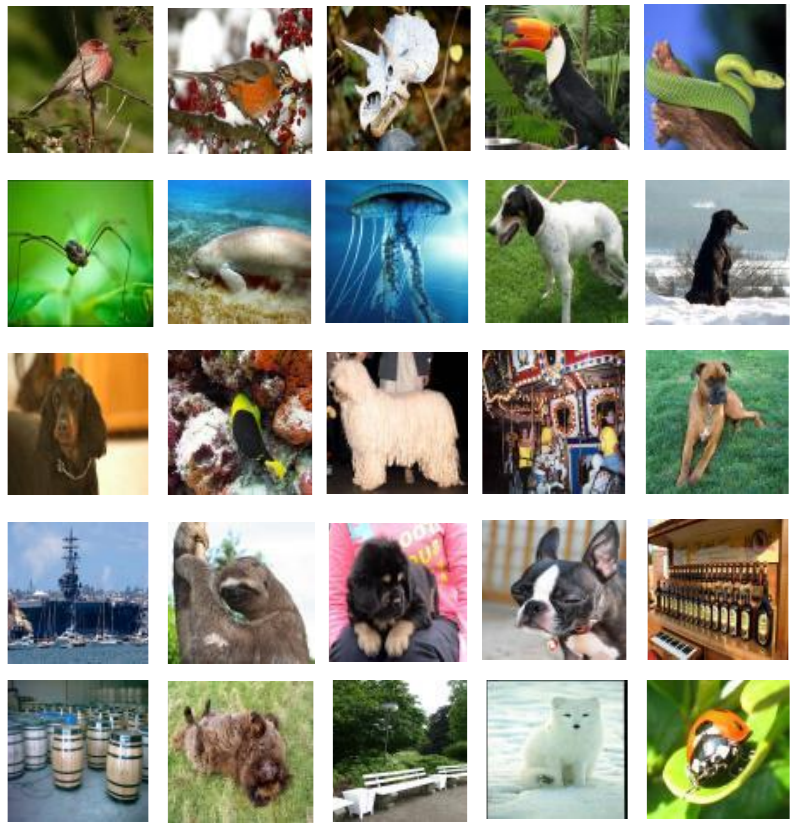}}
     \subfigure[Omniglot]{\includegraphics[width=0.23\textwidth]{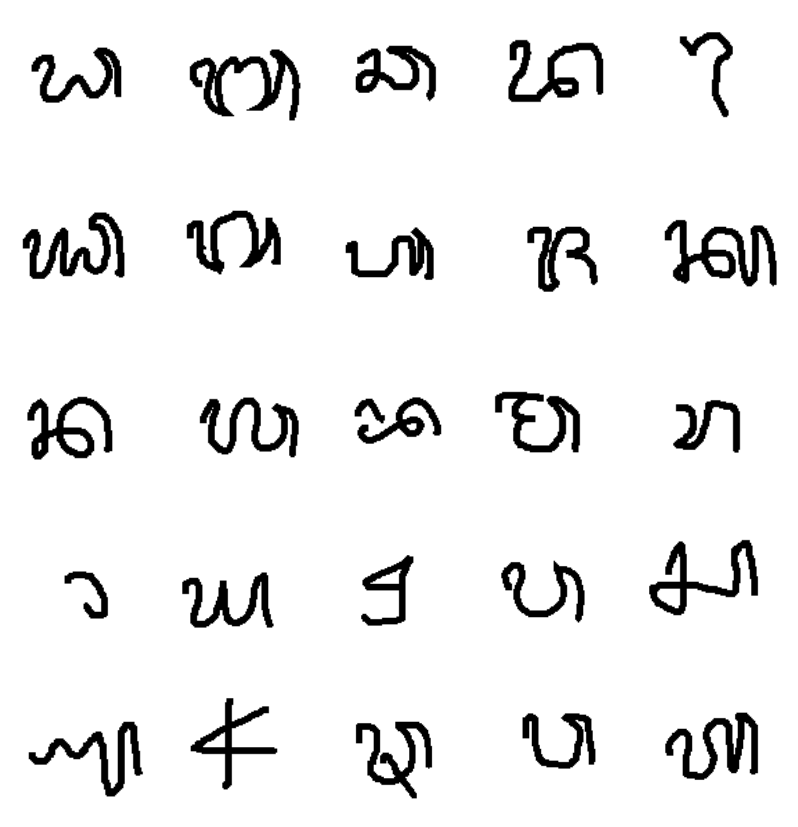}}
\caption{Examples of tasks sampled from different datasets. (a) shows the data sampled from the miniImagenet dataset \cite{miniImagenet} and (b) shows the data sampled from the Omniglot dataset \cite{Omniglot}. The distribution of tasks sampled from different datasets may vary greatly, where the former consists of various RGB images rich in information, while the latter is composed of binary characters.}
\label{fig:example_motivation}
\end{center}
\end{figure}

\begin{figure}
\begin{center}
\centerline{\includegraphics[width=0.9\columnwidth]{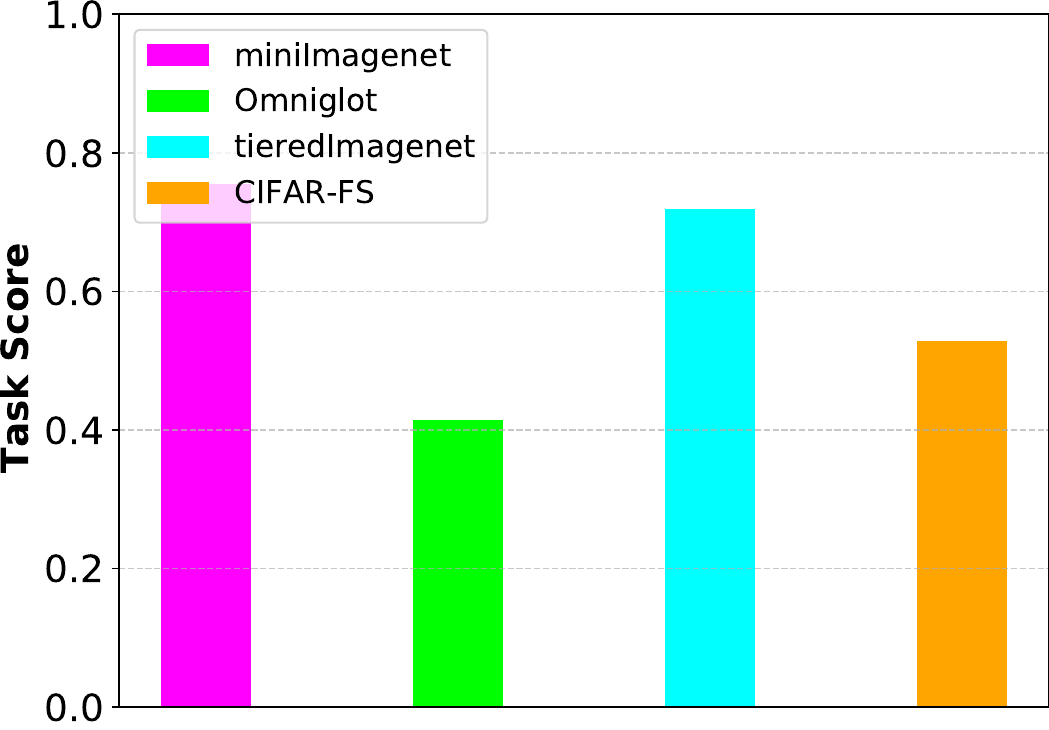}}
\caption{The scores of the task distribution in the four benchmark datasets, i.e., miniImagenet, Omniglot, tieredImagenet, and CIFAR-FS. A higher score means more knowledge is involved in the task.}
\label{fig:motivation_task_distribution}
\end{center}
\end{figure}

\begin{table*}[t]
\caption{Performance (accuracy $\pm$ 95\% confidence interval and training time) of meta-learning models with different structures. The experiments are conducted on four benchmark datasets, i.e., (5-way 1-shot) miniImagenet, (20-way 1-shot) Omniglot, (5-way 1-shot) tieredImagent, and (5-way 1-shot) CIFAR-FS. The $N\%$ indicates that the model is with $N\%$ dropout.}
\label{tab:motivation}
\begin{center}
\begin{small}
\begin{sc}
\resizebox{\linewidth}{!}{
\begin{tabular}{l|l|c|cc|cc|cc|cc}
\toprule[1.2pt]
\multirow{2}{*}{ } & \multirow{2}{*}{\textbf{Backbone}} & \multirow{2}{*}{\textbf{Dropout}}  & \multicolumn{2}{c|}{\textbf{miniImagenet}} & \multicolumn{2}{c|}{\textbf{Omniglot}} & \multicolumn{2}{c|}{\textbf{tieredImagenet}} & \multicolumn{2}{c}{\textbf{CIFAR-FS}} \\ 
    & & & \textbf{Accuracy (\%)} & \textbf{Training Time (h)} & \textbf{Accuracy (\%)} & \textbf{Training Time (h)} & \textbf{Accuracy (\%)} & \textbf{Training Time (h)} & \textbf{Accuracy (\%)} & \textbf{Training Time (h)}\\
\midrule
\multirow{15}{*}{\rotatebox{90}{\textbf{MAML}}}
& \multirow{3}{*}{\textbf{Conv4}} & 10\% & 42.63 $\pm$ 0.25 & 5.72 & 86.24 $\pm$ 0.37 & 4.02 & 48.75 $\pm$ 0.33 & 5.41 & 53.94 $\pm$ 0.37 & 4.98 \\
&  & 30\% & 42.54 $\pm$ 0.28 & 5.46 & 86.07 $\pm$ 0.26 & 3.94 & 48.82 $\pm$ 0.41 & 5.12 & 53.48 $\pm$ 0.31 & 4.72 \\
&  & 50\% & 42.47 $\pm$ 0.34 & 5.21 & 85.96 $\pm$ 0.31 & 3.88 & 48.98 $\pm$ 0.21 & 5.06 & 53.24 $\pm$ 0.31 & 4.02 \\
& \multirow{3}{*}{\textbf{VGG16}} & 10\% & 42.84 $\pm$ 0.36 & 6.34 & 84.72 $\pm$ 0.46 & 4.47 & 48.42 $\pm$ 0.32 & 6.08 & 54.32 $\pm$ 0.47 & 5.41 \\
&  & 30\% & 42.89 $\pm$ 0.27 & 5.97 & 85.09 $\pm$ 0.27 & 4.38 & 48.49 $\pm$ 0.31 & 5.97 & 54.22 $\pm$ 0.36 & 5.18 \\
&  & 50\% & 42.91 $\pm$ 0.47 & 5.83 & 85.47 $\pm$ 0.51 & 4.26 & 48.53 $\pm$ 0.27 & 5.63 & 54.19 $\pm$ 0.37 & 4.85 \\
& \multirow{3}{*}{\textbf{Resnet18}} & 10\% & 42.96 $\pm$ 0.53 & 6.81 & 85.12 $\pm$ 0.47 & 4.63 & 48.87 $\pm$ 0.24 & 6.74 & 54.46 $\pm$ 0.28 & 6.24 \\
&  & 30\% & 42.86 $\pm$ 0.51 & 6.53 & 85.07 $\pm$ 0.36 & 4.47 & 48.57 $\pm$ 0.29 & 6.51 & 54.49 $\pm$ 0.34 & 5.11\\
&  & 50\% & 42.78 $\pm$ 0.37 & 6.39 & 85.93 $\pm$ 0.26 & 4.21 & 48.29 $\pm$ 0.41 & 6.37 & 54.52 $\pm$ 0.46 & 5.08 \\
& \multirow{3}{*}{\textbf{Resnet50}} & 10\% & 43.15 $\pm$ 0.28 & 7.25 & 85.05 $\pm$ 0.58 & 5.42 & 48.93 $\pm$ 0.35 & 7.03 & 54.88 $\pm$ 0.36 & 6.82 \\
&  & 30\% & 43.23 $\pm$ 0.31 & 6.87 & 85.29 $\pm$ 0.34 & 4.98 & 48.72 $\pm$ 0.21 & 7.01 & 54.92 $\pm$ 0.37 & 6.39 \\
&  & 50\% & 43.44 $\pm$ 0.42 & 6.74 & 85.42 $\pm$ 0.27 & 4.74 & 48.57 $\pm$ 0.48 & 6.59 & 54.95 $\pm$ 0.41 & 5.94 \\
& \multirow{3}{*}{\textbf{Resnet101}} & 10\% & 43.32 $\pm$ 0.43 & 7.48 & 84.52 $\pm$ 0.39 & 5.78 & 49.28 $\pm$ 0.54 & 7.46 & 54.97 $\pm$ 0.23 & 7.54 \\
&  & 30\% & 42.98 $\pm$ 0.37 & 7.26 & 84.57 $\pm$ 0.42 & 5.42 & 49.16 $\pm$ 0.21 & 7.38 & 55.04 $\pm$ 0.24 & 7.28 \\
&  & 50\% & 42.93 $\pm$ 0.37 & 6.98 & 84.63 $\pm$ 0.59 & 5.24 & 48.91 $\pm$ 0.23 & 7.21 & 55.27 $\pm$ 0.31 & 6.45 \\
\midrule
\multirow{15}{*}{\rotatebox{90}{\textbf{ProtoNet}}}
& \multirow{3}{*}{\textbf{Conv4}} & 10\% & 43.41 $\pm$ 0.44 & 5.42 & 88.42 $\pm$ 0.37 & 4.86 & 48.61 $\pm$ 0.24 & 5.08 & 52.83 $\pm$ 0.41 & 4.29 \\
&  & 30\% & 43.58 $\pm$ 0.37 & 5.26 & 87.73 $\pm$ 0.23 & 5.18 & 48.79 $\pm$ 0.43 & 5.49 & 53.72 $\pm$ 0.46 & 4.83\\
&  & 50\% & 44.27 $\pm$ 0.32 & 6.27 & 87.69 $\pm$ 0.47 & 5.26 & 49.21 $\pm$ 0.33 & 5.84 & 53.86 $\pm$ 0.37 & 5.38\\
& \multirow{3}{*}{\textbf{VGG16}} & 10\% & 44.36 $\pm$ 0.27 & 5.87 & 87.64 $\pm$ 0.25 & 4.97 & 49.34 $\pm$ 0.57 & 4.69 & 53.69 $\pm$ 0.41 & 5.46 \\
&  & 30\% & 44.59 $\pm$ 0.35 & 5.94 & 87.81 $\pm$ 0.45 & 5.37 & 49.73 $\pm$ 0.29 & 4.92 & 53.84 $\pm$ 0.31 & 5.66\\
&  & 50\% & 44.62 $\pm$ 0.57 & 6.32 & 88.29 $\pm$ 0.26 & 5.44 & 49.87 $\pm$ 0.31 & 5.37 & 53.92 $\pm$ 0.48 & 5.72 \\
& \multirow{3}{*}{\textbf{Resnet18}} & 10\% & 43.56 $\pm$ 0.29 & 5.31 & 86.38 $\pm$ 0.26 & 5.69 & 48.21 $\pm$ 0.28 & 4.76 & 52.16 $\pm$ 0.37 & 4.69 \\
&  & 30\% & 43.61 $\pm$ 0.36 & 5.64 & 86.42 $\pm$ 0.39 & 5.78 & 48.68 $\pm$ 0.53 & 4.93 & 52.66 $\pm$ 0.31 & 4.79 \\
&  & 50\% & 43.87 $\pm$ 0.24 & 5.96 & 87.88 $\pm$ 0.31 & 5.84 & 48.76 $\pm$ 0.51 & 5.27 & 52.78 $\pm$ 0.33 & 4.96\\
& \multirow{3}{*}{\textbf{Resnet50}} & 10\% & 43.75 $\pm$ 0.27 & 5.78 & 86.30 $\pm$ 0.39 & 5.67 & 49.20 $\pm$ 0.47 & 4.86 & 52.78 $\pm$ 0.36 & 4.72 \\
&  & 30\% & 44.23 $\pm$ 0.26 & 5.89 & 86.32 $\pm$ 0.42 & 5.84 & 49.37 $\pm$ 0.28 & 4.92 & 53.87 $\pm$ 0.24 & 4.94\\
&  & 50\% & 44.67 $\pm$ 0.47 & 6.21 & 86.61 $\pm$ 0.43 & 5.92 & 49.58 $\pm$ 0.32 & 5.36 & 53.98 $\pm$ 0.33 & 5.27 \\
& \multirow{3}{*}{\textbf{Resnet101}} & 10\% & 43.27 $\pm$ 0.37 & 5.97 & 85.18 $\pm$ 0.51 & 4.23 & 48.76 $\pm$ 0.53 & 4.83 & 52.91 $\pm$ 0.29 & 5.34 \\
&  & 30\% & 43.64 $\pm$ 0.25 & 6.44 & 85.94 $\pm$ 0.28 & 4.86 & 48.92 $\pm$ 0.47 & 5.24 & 53.64 $\pm$ 0.27 & 4.96 \\
&  & 50\% & 43.78 $\pm$ 0.42 & 6.62 & 86.87 $\pm$ 0.35 & 5.21 & 49.33 $\pm$ 0.24 & 5.68 & 53.82 $\pm$ 0.48 & 4.63 \\
\bottomrule
\end{tabular}}
\end{sc}
\end{small}
\end{center}
\end{table*}

\section{Problem Analysis and Motivation}
\label{sec:3}

Based on previous analysis, despite meta-learning enhancing models' ability to tackle diverse tasks, it still faces one key limitation compared to the powerful BNS: it relies on a fixed structure to solve all tasks, lacking flexibility. Thus, we wonder whether FNS is crucial for meta-learning, e.g., enhancing performance. To answer this question, in this section, we first present the problem setting and notations of meta-learning in \textbf{Subsection \ref{sec:pre_meta-learning}}. Then, we provide both empirical and theoretical analyses to evaluate the effect of FNS in \textbf{Subsections \ref{sec:3.1} and \ref{sec:3.2}}, respectively.

\subsection{Problem Setting}
\label{sec:pre_meta-learning}
Meta-learning aims to learn from a set of tasks to quickly adapt to a new task. Generally, the tasks are assumed to be drawn from a fixed distribution $\mathbb{P} (\mathcal{T})$, denoted as $\mathcal{T}\sim \mathbb{P} (\mathcal{T})$. The tasks are divided into a training dataset $\mathcal{D}_{tr}$ and a test dataset $\mathcal{D}_{te}$ without class-level overlap. In the training phase, each batch contains $N_{tr}$ tasks, denoted as $\left \{ \tau_i \right \}_{i=1}^{N_{tr}} \in \mathcal{D}_{tr}$. Each task $\tau_{i}$ consists of a support set $\mathcal{D}_i^s=\{ (x_{i,j}^s,y_{i,j}^s)  \}_{j=1}^{N_i^s}$ and a query set $\mathcal{D}_i^q=  \{ (x_{i,j}^q,y_{i,j}^q)  \}_{j=1}^{N_i^q}$, where $(x_{i,j}^{\cdot},y_{i,j}^{\cdot})$ is the sample and the corresponding label, and $N_i^{\cdot}$ denotes the number of the samples. Meta-learning attempts to obtain a model $f_{\theta}$ using the training tasks, such that the model can be quickly fine-tuned to minimize the loss $\mathcal{L}(\mathcal{D}^{te},f_{\theta}) $ for deployment.
% The meta-learning model $f_{\theta}=h \circ g$ utilizes the feature encoder $g$ and the classifier $h$ to learn the above tasks.

The learning mechanism of meta-learning is regarded as an inner-outer loop process. In the inner loop, meta-learning model $f_{\theta}$ is fine-tuned on $\tau_i$ with the support set $\mathcal{D}_i^s$ and obtain the task-specific model $f_{\theta^i}$. The objective can be expressed as:
\begin{equation}
\label{eq:inner}
\begin{array}{l}
    \theta^i \gets \theta -\alpha \nabla_{\theta}\mathcal{L}(\mathcal{D} _i^s,\theta) \\[8pt]
    \text{where} \quad \mathcal{L}(\mathcal{D} _i^s,\theta )=\frac{1}{N_i^s} \sum_{j=1}^{N_i^s}y_{i,j}^s\log {f_\theta }(x_{i,j}^s)
\end{array}
\end{equation}
where $\alpha$ is the learning rate. In the outer loop, the meta-learning model $f_{\theta}$ is learned using the query sets from all training tasks and the corresponding expected task-specific models for each task. Thus, the objective can be expressed as:
\begin{equation}
\label{eq:outer}
\begin{array}{l}
    \theta \gets \theta-\beta \nabla_{\theta}\frac{1}{N_{tr}}\sum_{i=1}^{N_{tr}}\mathcal{L}(\mathcal{D} _i^q,\theta^i)  \\[8pt]
    \text{where} \quad \mathcal{L}(\mathcal{D} _i^q,\theta^i)= \frac{1}{N_i^q} \sum_{j=1}^{N_i^q}y_{i,j}^q\log {f_{\theta^i}}(x_{i,j}^q) 
\end{array}
\end{equation}
where $\beta$ is the learning rate. Note that $\theta^i$ is obtained by taking the derivative of $\theta$, so the update of $\theta^i$ can be regarded as a function of $\theta$. Therefore, the update of $\theta$ (Eq.\ref{eq:outer}) can be viewed as calculating the second derivative of $\theta$.

\subsection{Empirical Analysis}

\label{sec:3.1}

Meta-learning involves learning from a series of randomly sampled tasks. Due to sample differences and sampling randomness, tasks sampled from different datasets exhibit varying distributions. For instance, as shown in \textbf{Figure \ref{fig:example_motivation}}, Omniglot consists of a series of binary characters, while miniImagenet contains RGB samples across multiple classes. Recent works \cite{wang2020generalizing,wang2024towards,wang2023hacking} have shown that the difficulty of model learning varies under different task distributions. An ideal meta-learning model should perform well across all task distributions. To assess the impact of FNS on performance, we conduct experiments with different meta-learning models under various task distributions. If the model structure minimally influences performance, meta-learning models with different structures should demonstrate similar performance across all distributions. Conversely, if there are significant differences in performance under different model scales, it indicates that FNS greatly influences meta-learning.

Specifically, we select four meta-learning benchmark datasets for evaluation, including miniImagenet \cite{miniImagenet}, Omniglot \cite{Omniglot}, tieredImagenet \cite{tieredImagenet}, and CIFAR-FS \cite{CIFAR-FS}. 
Following \cite{wang2024towards}, we assess their task distributions via four steps: (i) randomly sample to create a candidate pool of training tasks; (ii) select 50 tasks from this pool; (iii) compute task scores based on three metrics: task diversity, task entropy, and task difficulty; and (iv) average these scores to derive a task distribution score for each dataset. Higher scores indicate more complex tasks, requiring more time for the model to learn. The task distribution scores are shown in \textbf{Figure \ref{fig:motivation_task_distribution}}, revealing that miniImagenet and tieredImagenet have the most complex tasks, followed by CIFAR-FS, with Omniglot featuring the simplest tasks.
We construct five meta-learning models with different backbone scales: Conv4, VGG16, ResNet18, ResNet50, and ResNet101, each followed by the same classification head. Conv4, with only four convolutional layers, has the fewest parameters, while ResNet101 has the most. We apply varying dropout rates (10\%, 30\%, and 50\%) across all datasets for each model. The dropout percentage indicates the proportion of neurons deactivated during training. Finally, we evaluate the performance and training time post-convergence of each model.

\textbf{Figure \ref{fig:motivation}} illustrates the trade-off performance of models using MAML as the meta-learning strategy. The results show that different datasets favor different model scales.
For instance, in Omniglot, the Conv4 model with a 10\% dropout achieves the highest training efficiency, yielding performance comparable to or better than deeper networks like ResNet18. However, in miniImagenet, although Conv4 results in the shortest training time, its performance is worse than that of ResNet18 under varying dropout rates. \textbf{Table \ref{tab:motivation}} presents quantitative results for different meta-learning strategies, revealing similar trends to those in \textbf{Figure \ref{fig:motivation}}: the optimal model scale varies across task distributions without a consistent pattern.

In summary, we obtain two interesting conclusions: (i) A universal model structure does not exist for all task distributions; it varies with changing task distributions, so a one-size-fits-all approach is not effective. (ii) Traditional meta-learning methods rely on preset structural priors, i.e., a fixed model to address all tasks, which limits their effectiveness. Therefore, FNS is crucial for meta-learning models.

\subsection{Theoretical Analysis}
\label{sec:3.2}

In this subsection, we theoretically show that the structure of meta-learning models plays a critical role in their performance.
Firstly, we prove that it is difficult to find an optimal model with a fixed structure across different task distributions (\textbf{Theorem \ref{theorem:motivation_1}}). Next, we demonstrate that the probability of obtaining an optimal model depends on its structure (\textbf{Theorem \ref{theorem:motivation_2}}). All the proofs are provided in \textbf{Appendix \ref{sec:app_2}}.

Firstly, we demonstrate that there is no universal and fixed meta-learning model for different task distributions. We achieve this by proving that no model can succeed in all learning tasks.

\begin{theorem}\label{theorem:motivation_1}

    Let $f$ be any meta-learning model for the task of binary classification with respect to the 0-1 loss over the data $\mathcal{X}$ of task $ \tau $. Then, there exists a distribution $ P_\mathcal{X} $ over $ \mathcal{X} \times \{0, 1\} $ such that:    \begin{itemize}
        \item There exists a function $\mathcal{F}:  \mathcal{X} \to \{0, 1\}$ with $\mathcal{L}_{P_\mathcal{X}}(\mathcal{F}) = 0$. 
        \item Let $ m  < \left | \mathcal{X} \right | /2 $ representing the size of the training set. over the choice of $\mathcal{X}^i\sim P_{\mathcal{X}}^m$, we have $P(\mathcal{L}_{P_{\mathcal{X}}}(f(\mathcal{X}^i)) \ge  1/8) \ge 1/7$.
    \end{itemize}
\end{theorem}
The theorem states that for any given model, it is impossible to achieve success in every task. Specifically, a trivial successful model would be the empirical risk minimization (ERM) model applied under a finite class of assumptions.

Next, we turn to demonstrate that model performance on different task distributions is related to structure. We achieve this by proving that given a candidate pool of models, the probability of selecting different models on a specific task distribution is related to the network structure and neuron weights of the model. The higher the selection probability, the better the model performs on the current task distribution.

\begin{theorem}\label{theorem:motivation_2}
    Suppose we have a set of candidate models $f^i, i=1,2,\dots, M$ and corresponding model parameters $\theta_i$, we aim to select the best model from them. Given a prior distribution $P(\theta_i|f^i)$ for $f^i$ and the dataset $\mathcal{D}_i$ for task $\tau_i$, to compare the model selection probabilities, we use Laplace approximation to estimate $P(\mathcal{D}_i|f^i)$:
    \begin{equation}
        \log P(\mathcal{D}_i|f^i)=\log P(\mathcal{D}_i|\hat{\theta}_i,f^i)-\frac{K_i}{2} \log N +\mathcal{O}(1)
    \end{equation}
    where $\hat{\theta}_i$ is a maximum likelihood estimate and $K_i$ denotes the number of free parameters in model $f^i$ that reflects structural frugality. Then, the posterior probability of each model $f^i$ for task $\tau_i$ can be estimated as:
        $\frac{e^{-\log P(\mathcal{D}_i|\hat{\theta}_i,f^i)}}{\sum_{j=1}^{M}e^{-\log P(\mathcal{D}_j|\hat{\theta}_j,f^j  )} } $,
    which is related to the network structure and neuron weight of $\theta_i$.
\end{theorem}

The model selection in the aforementioned theorem is asymptotically consistent. This means that given a series of models (including the ground-truth model), as the sample size $N \to \infty$, the probability of selecting the correct model approaches 1. This indicates that the probability of model selection for a specific task distribution is almost accurate. 
Therefore, we obtain two theoretically provable conclusions: (i) There is not an optimal model with a fixed structure across different tasks; (ii) The optimal model's selection probability is related to the model structure. 
In other words, FNS is essential for handling different tasks, yet current meta-learning methods are limited by fixed structure priors.
Combined with \textbf{Subsection \ref{sec:3.1}}, these empirical and theoretical results inspire us to seek FNS in meta-learning, enhancing meta-learning performance.

\begin{table}[t]
    \centering
    
    \caption{The definitions of all notations.}
    \resizebox{\linewidth}{!}{
    \begin{tabular}{c|c}
    \toprule
    Notations & Definition \\
    \midrule
    \multicolumn{2}{c}{Notations of Data} \\
    \midrule
       $\mathbb{P} (\mathcal{T})$  & The task distribution.  \\
       $\mathcal{T}\sim \mathbb{P} (\mathcal{T})$ & The tasks for training and testing \\
       $\mathcal{D}_{tr}$ & The training task set \\
       $\mathcal{D}_{te}$ & The testing task set\\
       $\left \{ \tau_i \right \}_{i=1}^{N_{tr}} \in \mathcal{D}_{tr}$ & The tasks for training\\
       $N_{tr}$ & The number of training tasks. \\
       $\mathcal{D}_i$ & The training samples of task $\tau_i$\\
       $N_i$ & The number of samples in $\mathcal{D}_i$\\
       $\mathcal{D}_i^s=\{ (x_{i,j}^s,y_{i,j}^s)  \}_{j=1}^{N_i^s}$ & The support set of training task $\tau_i$ \\
       $N_i^s$ & The number of samples in $\mathcal{D}_i^s$ \\
       $x_{i,j}^{s}$ & The $j$-th sample in $\mathcal{D}_i^s$ \\
       $y_{i,j}^{s}$ & The corresponding label of sample $x_{i,j}^{s}$ \\
       $\mathcal{D}_i^q=  \{ (x_{i,j}^q,y_{i,j}^q)  \}_{j=1}^{N_i^q}$ & The query set of task $\tau_i$ \\
       $N_i^q$ & The number of samples in $\mathcal{D}_i^q$ \\
       $x_{i,j}^{q}$ & The $j$-th sample in $\mathcal{D}_i^q$ \\
       $y_{i,j}^{q}$ & The corresponding label of sample $x_{i,j}^{q}$ \\
    \midrule
    \multicolumn{2}{c}{Notations of NeuronML} \\
    \midrule   
        $f_{\theta_{\mathcal{M}}}$ & The NeuronML model\\
        $\theta_{\mathcal{M}}$ & The parameter of $f_{\theta_{\mathcal{M}}}$, $\theta_{\mathcal{M}} \gets \mathcal{M} \odot \theta$ \\
        $\theta$ & The original parameters\\
        $\mathcal{M}$ & The embedded learnable structure mask \\
        $f_{\theta_{\mathcal{M}}^i}$ & The task-specific model for task $\tau_i$ \\
        $\theta_{\mathcal{M}}^i$ & The parameter of $f_{\theta_{\mathcal{M}}^i}$ \\
        $\mathcal{L}_{weight}(\cdot)$ & The loss of weight, e.g., cross-entropy\\
        $\mathcal{L}_{structure}(\cdot)$ & The proposed structure constraint (Eq.\ref{eq:structure_constraint}) \\
        $\mathcal{L}_{fr}(\cdot)$ & The constraint of frugality (Eq.\ref{eq:constraint_frugality})\\
        $\lambda_{fr}$ & The weight hyperparameter of $\mathcal{L}_{fr}(\cdot)$\\
        $\mathcal{L}_{pl}(\cdot)$ & The constraint of plasticity (Eq.\ref{eq:constraint_plasticity})\\
        $\lambda_{pl}$ & The weight hyperparameter of $\mathcal{L}_{pl}(\cdot)$\\
        $\mathcal{L}_{se}(\cdot)$ & The constraint of sensitivity (Eq.\ref{eq:constraint_sensitivity})\\
        $\lambda_{se}$ & The weight hyperparameter of $\mathcal{L}_{se}(\cdot)$\\
    \bottomrule
    \end{tabular}
    }
    \label{tab:notation1}
\end{table}

\begin{algorithm}[t]
   \caption{Neuromodulated Meta-Learning}
   \label{alg:example}
\begin{flushleft}
\textbf{Input:} Task distribution $\mathbb{P} (\mathcal{T})$, NeuronML model $f_{\theta_{\mathcal{M}}}$ with randomly initialized parameters $\theta_\mathcal{M}=\mathcal{M}\odot\theta$\\
\textbf{Output:} The trained NeuronML model $f_{\theta_{\mathcal{M}}}$\\
\end{flushleft}
\begin{algorithmic}[1]
   \WHILE{not done}
   \STATE Sample $N_{tr}$ tasks from $\mathbb{P} (\mathcal{T})$, $\left \{ \tau_i \right \}_{i=1}^{N_{tr}} \sim \mathbb{P} (\mathcal{T})$
   \FOR{$i=1$ {\bfseries to} $N_{tr}$}
   \STATE Divide the data $\mathcal{D}_i$ of $\tau_i$ into $\mathcal{D}_i^s$ and $\mathcal{D}_i^q$
   \STATE Compute the loss $\mathcal{L}_{weight}(\mathcal{D}^s _i,\theta_\mathcal{M})$ via Eq.\ref{eq:inner}
   \STATE Update $\theta_\mathcal{M}^i$ using $\mathcal{L}_{weight}(\mathcal{D}^s _i,\theta_\mathcal{M})$ via Eq.\ref{eq:neuron_weight}
   \STATE Compute $\mathcal{L}_{weight}(\mathcal{D}^q _i,\theta_\mathcal{M}^i)$ via Eq.\ref{eq:outer} 
   \STATE Compute $\mathcal{L}_{structure}(\mathcal{D}_i,\theta_\mathcal{M}^i)$ via Eq.\ref{eq:structure_constraint}
   \ENDFOR
   \STATE Update $\theta$ of $\theta_\mathcal{M}$ using $\mathcal{L}_{weight}(\mathcal{D}^q _i,\theta_\mathcal{M}^i)$ via Eq.\ref{eq:neuron_weight}
   \STATE Update $\mathcal{M}$ of $\theta_\mathcal{M}$ using $\mathcal{L}_{structure}(\mathcal{D}_i,\theta_\mathcal{M}^i)$ via Eq.\ref{eq:neuron_structure}
   \ENDWHILE
\end{algorithmic}
\end{algorithm}

\section{Methodology}
\label{sec:4}
In this section, we aim to define, measure, and model FNS in meta-learning. Specifically, based on the above analyses, we propose that a good FNS should possess frugality, plasticity, and sensitivity, and provide its definition in \textbf{Subsection \ref{sec:Definition}}. We then present three measurements for the three properties, backed by theoretical supports to ensure practical evaluation (\textbf{Subsection \ref{sec:Constraint}}). We further construct the structure constraint by combining them. Next, to model FNS in meta-learning, we propose NeuronML which updates both model weights and structure with the structure constraint (\textbf{Subsection \ref{sec:Method}}).
Note that NeuronML can be used in various problems, e.g., supervised, self-supervised, and reinforcement learning, with specific instantiations provided in \textbf{Appendix \ref{sec:5}}.

\subsection{Definition of FNS}
\label{sec:Definition}

In this subsection, we aim to answer what a good FNS is and provide its formal definition. Specifically, we propose that a good model structure should achieve a balance among model complexity, task distribution, and model performance. This balance allows it to adapt to known data variations while maintaining the predictive ability for unseen samples. 
Thus, we propose that a good FNS in meta-learning should possess the following three properties:
\begin{itemize}
    \item Frugality: activates only a small subset of neurons, controlling complexity and mitigating overfitting due to over-parameterization in traditional meta-learning models.
    \item Plasticity: the activated neurons vary across tasks and can be dynamically adjusted based on specific task demands, allowing for flexibility beyond a fixed structure.
    \item Sensitivity: the activated neurons should be the most effective and act as a proxy for the full set. This ensures excellent performance within a sparse structure, balancing training efficiency with performance.
\end{itemize}

Based on this insight, we propose a mathematical formal definition of a good FNS for meta-learning models.
\begin{definition}\label{Definition:structure}
A meta-learning model $f_{\theta}$ is defined as possessing a good flexible network structure (FNS) if it satisfies:
\begin{itemize}
    \item Frugality: For each task $\tau_i$, the set of activated neurons $\theta^i$ satisfies $\|\theta^i\|_0 \leq k$, where $k$ represents the maximum number of neurons allowed to be activated.
    \item Plasticity: For any two distinct tasks $\tau_i$ and $\tau_{j}$ ($i \neq j$), it holds that $\theta^i \neq \theta^{j}$ where $\theta^i$ and $\theta^j$ are the sets of activated neurons for the corresponding tasks.
    \item Sensitivity:
    For each task $\tau_i$, the activated neurons $\theta^i$ should maximize task performance: $ \theta^i = \arg\min_{\theta \in \{0,1\}^d, \|\theta\|_0 \leq k} \mathcal{L}(f_{\theta},\tau_i)  \nonumber$,

   where $\mathcal{L}(f_{\theta},\tau_i)$ denotes the training loss of $f_{\theta}$ on task $\tau_i$.
\end{itemize}
\end{definition}

Note that each property is indispensable for meta-learning: Considering the over-parameterization in existing meta-learning models, frugality ensures a balance between the parameter dimensions and the amount of data; Given the multi-task learning mechanism of meta-learning and the distribution shifts between tasks, plasticity ensures the optimal structure of the model for different tasks; Considering the performance of meta-learning model on various tasks, sensitivity ensures that the limited neurons can act as a proxy for the full set without significant difference of accuracy. The experiments in \textbf{Section \ref{sec:7}} also demonstrate that each property is indispensable. Therefore, this definition explains what a good FNS is, helping us measure and model it in practice.

\subsection{Measurements of FNS}
\label{sec:Constraint}
Based on the above definition, we propose three measurements for frugality, plasticity, and sensitivity to measure the flexibility of the meta-learning model structure in practice. 

First, we provide the measurement of frugality as:
\begin{theorem}[\textbf{Measurement of Frugality}]\label{theorem:constraint_sparse} Given the meta-learning model $f_\theta$ with parameter set $\theta$, an observation vector $\mathcal{Y} \in \mathbb{R}^n $ and a design matrix $\mathcal{X} \in \mathbb{R}^{n \times m}$, the goal of $f_\theta$ is to obtain $\mathcal{X}\theta=\mathcal{Y}$ with $\|\theta\|_0 \leq k$. Define a given subset $\mathcal{S}=\{1,\cdots,m\}$ with a convex cone $\mathbb{C}(\mathcal{S}^{\theta}):=\{\theta \in \mathbb{R}^{m}|\|\theta_{\mathcal{S}_{c}^{\theta}}\|_1\le \|\theta_{\mathcal{S}^{\theta}}\|_1\}$, if satisfies $Null(\mathcal{X})\cap \mathbb{C}(\mathcal{S}^{\theta})=\{0\}$, the basis pursuit relaxation of $\|\theta\|_0$ has a unique solution equal to $\|\theta\|_1$.
Then, for task $\tau_i$, we define the $\theta^i$ with frugality derived from the minimization of the subsequent objective:
    \begin{equation}\label{eq:constraint_frugality}
    \mathcal{L}_{fr}(\theta^i) = \|\theta^i\|_1,\text{s.t.} \|\theta^i\|_1 \leq \max\{C, \gamma \cdot d \cdot \log\left(\frac{N_i}{d}\right)\}
\end{equation}
where $C$ is a constant, $\gamma$ is a scaling factor, $N_i$ is the sample size for task $\tau_i$, and $d$ is the dimensionality of $\theta^i$.
\end{theorem}
\textbf{Theorem \ref{theorem:constraint_sparse}} states that when the restricted null space property is satisfied, i.e., the only elements in cone $\mathbb{C}(\mathcal{S}^{\theta})$ that lie in the null space of $\mathcal{X}$ are all-zero vectors, the $\ell_0$-norm and $\ell_1$-norm are equivalent. Meanwhile, the $\ell_0$ norm enforces frugality by counting the number of non-zero elements in a vector \cite{hurley2009comparing,donoho2006compressed}. Despite it allows control over the number of activated neurons as $\|\theta^i\|_0 \leq k$, the optimization problem is NP-hard \cite{hochba1997approximation,peharz2012sparse}.
Thus, we use $\ell_1$-norm, a relaxation form of $\ell_0$-norm, to control the model complexity, i.e., minimizing $\mathcal{L}_{{fr}}(\theta^i)$ leads to the model activating fewer neurons. 
It uses a provable balance between the parameter dimension and the data volume following \cite{hastie2015statistical}, i.e., $\max\{C, \gamma \cdot d \cdot \log\left(\frac{N_i}{d}\right)\}$, to constrain the bound of neuron frugality without relying on a fixed prior $k$. 
The proof is provided in \textbf{Appendix \ref{sec_app:constraint_spa}}.

Then, we provide the measurement of plasticity:

\begin{theorem}[\textbf{Measurement of Plasticity}]\label{theorem:constraint_plasticity}
Assume that for the matrix $ A(\theta) $, the minimal singular value $ \sigma_{\min}(A(\theta)) $ remains uniformly bounded below by a constant for all positive parameters. Denote subspace $ V_j $ for all $ \theta $, assume $\|A(\theta)\omega\| \leq \alpha_j \|x\|, \forall \omega \in V_j,$
   where $ \alpha_j $ is a positive parameter associated with the subspace $ V_j $. Then, we define the $\theta^i$ with plasticity derived from the minimization of the subsequent objective:
\begin{equation}\label{eq:constraint_plasticity}
    \mathcal{L}_{pl}(\theta^i, \theta^j) = \sum_{j \neq i} \sum_{\omega \in \theta^i} \mathbbm{1}(\theta^i[\omega], \theta^j[\omega]) \cdot p_\omega
\end{equation}
where $ p_\omega = \frac{e^{\beta \mathcal{L}_\omega}}{\sum_{k} e^{\beta \mathcal{L}_k}} $ is the activation probability of neuron $\omega$, based on its importance assessed in historical tasks following Hebbian rules \cite{kempter1999hebbian,lisman1989mechanism}, $\theta^i[\omega]$ represents the neuron $\omega$ in $\theta^i$, and $\mathbbm{1}(\theta^i[\omega], \theta^j[\omega])$ is the indicator function that returns 1 if both the neurons of $\tau_i$ and $\tau_j$, i.e., $\theta^i$ and $\theta^j$ activate $\omega$.
\end{theorem}
In \textbf{Theorem \ref{theorem:constraint_plasticity}}, the assumptions ensure that the minimal singular value of the matrix $A(\theta)$ is uniformly bounded below by a constant, ensuring stability. Then, $\mathcal{L}_{{pl}}(\cdot)$ uses $\mathbbm{1}(\cdot)$ to measure the overlap of activated neurons between different tasks and uses $p_\omega$ for their historical importance, i.e., the more $\omega$ is activated, the more important it is. Thus, the lower $\mathcal{L}_{{pl}}$ is, the fewer neurons are jointly activated across tasks, indicating higher plasticity. The proof is provided in \textbf{Appendix \ref{sec_app:constraint_dyn}}.

Next, we provide the measurement of sensitivity:

\begin{theorem}[\textbf{Measurement of Sensitivity}]\label{theorem:constraint_sensitivity}
Let $(\theta, \mathcal{X})$ be a positive definite tuple, and let $s\colon \theta\longrightarrow(0,\infty)$ be an upper sensitivity function with total score $S$. Then, we define the $\theta^i$ with sensitivity derived from the minimization of the objective:
\begin{equation}\label{eq:constraint_sensitivity}
    \mathcal{L}_{se}(\theta^i) = \sum_{\omega \in \theta^i} -\log\frac{s(\omega)}{S} \cdot \theta^i[\omega] 
\end{equation}
where $ s(\omega) = \frac{\partial \mathcal{L}(\theta^i, \tau_i)}{\partial \theta^i[\omega]} $ is the sensitivity score of neuron $\omega $ reflecting its impact on model performance of the corresponding task, and $\mathcal{L}(\theta^i, \tau_i)$ is the loss of model $f_{\theta^i}$ on task $\tau_i$.
There exists the sparse hyperparameter $C$ (\textbf{Theorem \ref{theorem:constraint_sparse}}) such that for any $0<\epsilon ,\delta<1$, if $C\geq \frac{S}{\epsilon ^2}\left(\mathsf{VC}\log S + \log\frac{1}{\delta}\right)$, with probability at least $1-\delta$, using the model $\hat{f}_{\theta}$ obtained via $\mathcal{L}_{se}(\theta^i)$ satisfies $\Big|\int f_\theta(x)\text{d} x - \int \hat{f}_\theta(x)\text{d} x \Big|\leq \epsilon$.
\end{theorem}
\textbf{Theorem \ref{theorem:constraint_sensitivity}} ensures that the selected neurons exhibit the highest sensitivity, indicating their high effectiveness for the given task. Specifically, the more critical the activated neurons are, the greater the reduction in task-specific loss, resulting in a smaller $ -\log\frac{s(\omega)}{S} \cdot \theta^i[\omega] $ and a lower $ \mathcal{L}_{se}(\theta^i) $. Furthermore, we prove that \( \mathcal{L}_{se}(\theta^i) \) guarantees that, under condition \( C \), the activated neurons can serve as a global proxy, ensuring the approximation \( \hat{f}_{\theta} \) closely matches the performance of the original model \( f_{\theta} \). The proof is provided in \textbf{Appendix \ref{sec_app:constraint_eff}}.

Combining the above three measurements, we propose the \textbf{structure constraint}, which can be expressed as:
\begin{equation}\label{eq:structure_constraint}
    \mathcal{L}_{structure}(\mathcal{D}_i,\theta^i) = \lambda_{{fr}}\mathcal{L}_{{fr}}(\theta^i) + \lambda_{{pl}}\mathcal{L}_{{pl}}(\theta^i, \theta^j) + \lambda_{{se}}\mathcal{L}_{{se}}(\theta^i)
\end{equation}
where $\mathcal{D}_i$ denotes the training samples of $\tau_i$. By adjusting the weight coefficients $\lambda_{{fr}}$, $\lambda_{{pl}}$, and $\lambda_{{se}}$, we can balance frugality, plasticity, and sensitivity, optimizing the meta-learning model structure to achieve the desired flexibility.

\subsection{Model FNS in Meta-Learning}
\label{sec:Method}
Based on the above analysis, we propose Neuromodulated Meta-Learning (NeuronML), a method that models FNS in meta-learning using the structure constraint. Its core idea is to decompose the parameter updates of meta-learning models into weight and structure components, enabling dynamic adjustment through bi-level optimization.

Specifically, we introduce a learnable structure mask $\mathcal{M}$ alongside the parameters $\theta$ of the meta-learning model $f_\theta$, with $\mathcal{M}$ sharing the same dimensionality as $\theta$, i.e., $\theta_{\mathcal{M}} \gets \mathcal{M} \odot \theta$. Each element of the structure mask $\mathcal{M}$ represents the activation probability of the corresponding neuron; for example, a score of 0 indicates that the neuron is deactivated for the given task. We leverage a bi-level optimization process to alternately update the parameters $\theta$ and the structure mask $\mathcal{M}$, enabling simultaneous constraints on both the weights (first level) and structure (second level) of the meta-learning model. 

\textbf{In the First-Level}, with the $\mathcal{M}$ held fixed, we optimize the parameter $\theta$ of meta-learning model $f_{\theta}$. It follows the inner-outer loop structure as mentioned in \textbf{Subsection \ref{sec:pre_meta-learning}}, but using the $\theta_{\mathcal{M}}$ instead of $\theta$. The overall objective becomes:
\begin{equation}\label{eq:neuron_weight}
\begin{array}{l}
    \arg \underset{\theta}{\min} \frac{1}{N_{tr}}\sum_{i=1}^{N_{tr}}\mathcal{L}_{weight}(\mathcal{D} _i^q,\theta_{\mathcal{\mathcal{M}}}) \\[5pt]
    s.t.\quad \theta_{\mathcal{M}}^i\gets \theta_{\mathcal{M}}-\alpha \nabla_\theta\mathcal{L}_{weight}(\mathcal{D} _i^s,\theta_{\mathcal{M}} )
\end{array}
\end{equation}
where $\mathcal{L}_{weight}$ is the same loss function as Eq.\ref{eq:inner} and Eq.\ref{eq:outer}. Through this equation, we aim to obtain a well-generalized parameter weight initialization $\theta$ as previous meta-learning does, but combined with a better structure constrained by $\mathcal{M}$. 

\textbf{In the Second-Level}, with the weights of $\theta$ held fixed, we optimize the learnable structure mask $\mathcal{M}$ using the structure constraint $\mathcal{L}_{structure}(\cdot)$ (Eq.\ref{eq:structure_constraint}), expressed as:
\begin{equation}\label{eq:neuron_structure}
    \arg \underset{\mathcal{M}}{\min} \frac{1}{N_{tr}}\sum_{i=1}^{N_{tr}}\mathcal{L}_{structure}(\mathcal{D} _i,\theta_{\mathcal{\mathcal{M}}})
\end{equation}
This equation makes the model optimize the structure mask $\mathcal{M}$ by minimizing $\mathcal{L}_{\text{structure}}(\cdot)$, approximating the optimal configuration for each task. In essence, because each task has a unique optimal model configuration, FNS is achieved when the model attains the best performance on each task by adapting its structure mask accordingly.

\textbf{In summary}, we optimize the parameter weights in the first level and the structure mask in the second level, then combine them to obtain a meta-learning model with optimal weights and structure. The pseudo-code is provided in \textbf{Algorithm \ref{alg:example}}.

\section{Theoretical Analysis}
\label{sec:6}
In this section, we provide theoretical guarantees of NeuronML. 
Specifically, we first demonstrate that the optimization objective of NeuronML inherits smoothness and strong convexity, showing it can converge to the optimal value (\textbf{Theorem \ref{the:objective}} and \textbf{Corollary \ref{cor:bound}}). Next, we provide a tighter generalization bound for NeuronML (\textbf{Theorem \ref{the:generalization}}), compared with previous meta-learning methods.
The proofs are shown in \textbf{Appendix \ref{sec:app_2}}.

First, we analyze the objective of NeuronML. The update procedure is shorted as $\mathcal{F}(\theta)$, with the population loss $\mathcal{L}^p(f_{\theta})=\sum_{i=1}^{N_{tr}}\mathcal{L}(f_{\theta^i}) $ and empirical loss $\mathcal{L}^e(f_{\theta})=\sum_{i=1}^{N_{tr}}\mathcal{L}(\mathcal{D}_i,f_{\theta^i}) $ where $\mathcal{D}_i$ is the data of task $\tau_i$.
Then, we have:
\begin{theorem}\label{the:objective}
    Suppose $\mathcal{F}$ has gradients bounded by $\mathrm{G}$, i.e. $|| \nabla \mathcal{F}(\theta) || \leq \mathrm {G} \ \forall \ \theta$, and $\varsigma -$smooth, i.e., $|| \nabla \mathcal{F}(\theta_1) - \nabla \mathcal{F}(\theta_1) || \leq \varsigma  ||\theta_1-\theta_2|| \ \forall (\theta_1,\theta_2 )$. Let $\tilde{f}_{\theta} $ be the function evaluated after a one-step gradient update procedure, i.e.,
    \begin{equation}
        \tilde{f}_{\theta} = f_{\theta}-\beta \nabla_{f_{\theta}}\mathcal{L}(\mathcal{D}^q _i,f_{\theta^i})
    \end{equation}
    If the step size $\beta \leq \min \{ \frac{1}{2\varsigma}, \frac{\sigma}{8 \rho \mathrm {G}  } \}$ where $\sigma-$strongly convex means $ || \nabla \mathcal{F}(\theta_1) - \nabla \mathcal{F}(\theta_1) || \geq \sigma||\theta_1-\theta_2||  $, then $\tilde{f}_{\theta}$ is convex. Also, if $f_{\theta}$ and $f_{\theta^i}$ also satisfy the above $\mathrm{G}$-Lipschitz and $\varsigma -$smooth assumptions of $\mathcal{F}$, then the learning objective $\arg \min _{f_\theta}\sum_{i=1}^{N_{tr}}\mathcal{L}(\mathcal{D}^q _i,f_{\theta^i})$
    with $\beta \leq \min \lbrace \frac{1}{2\varsigma}, \frac{\sigma}{8 \rho G} \rbrace$ is $\frac{9\varsigma}{8}-$ smooth and $\frac{\sigma}{8}-$strongly convex.
\end{theorem}
Since the objective function is convex, we next provide the first set of results under which NeuronML's objective function can be provably and efficiently optimized.
\begin{corollary}
\label{cor:bound}
    Suppose that for all iterations, $f_{\theta}$ and $f_{\theta^i}$ satisfy the $\mathrm{G}$-Lipschitz and $\varsigma -$smooth assumptions, and the update procedure in NeuronML is performed with $\beta \leq \min \{ \frac{1}{2\varsigma}, \frac{\sigma}{8 \rho \mathrm {G}  } \}$. Then, the model $f_{\theta}$ enjoys the guarantee $\sum_{i=1}^{N_{tr}}\mathcal{F}(\theta)= \min_{f_\theta }\sum_{i=1}^{N_{tr}}\mathcal{F}(\theta)+O(\frac{32\mathrm{G}^2}{\sigma}\log N_{tr} )$.
\end{corollary}

More generally, \textbf{Theorem \ref{the:objective}} and \textbf{Corollary \ref{cor:bound}} implies that the NeuronML enjoys sub-linear regret of meta-learning \cite{chen2021generalization,finn2019online} based on the inherited smoothness and strong convexity, and prove the reliability of NeuronML's objective. Further, we provide the theoretical performance guarantee of NeuronML.

\begin{theorem}\label{the:generalization}
    If $f_{\theta}$ is a possibly randomized symmetric and the assumption of \textbf{Corollary \ref{cor:bound}} hold, then
    $\mathbb{E}_{f_{\theta },\mathcal{D} }\left [ \mathcal{L}^p(f_{\theta})- \mathcal{L}^e(f_{\theta}) \right ]\le \xi $, where $\xi := O(1) \frac{G^2 (1 + \beta \varsigma K)}{N_{tr} N \sigma}$ is the bound of generalization error. It decays by a factor of $O(K/N_{tr} N)$, where $N_{tr}$ is the number of training tasks, $N$ is the number of available samples per task, and $K$ is the number of data points used in actual training batch.
\end{theorem}

\textbf{Theorem \ref{the:generalization}} proves that NeuronML has a tighter generalization bound than previous meta-learning \cite{maml,protonet}.
According to the above discussion, the reliability and superiority of NeuronML have been theoretically supported.

\section{Experiments}
\label{sec:7}

In this section, we conduct experiments to answer the following questions: (i) Can NeuronML be used well for meta-learning in various tasks, including regression, classification, and reinforcement learning? (ii) Does NeuronML provide empirical benefits over previously proposed methods?

To this end, we compare the following settings: (i) Training with fine-tuning (TFT), which involves training jointly until round $i\ne 1$ (task $\tau_i$), after which it fine-tunes specifically to round $i$ using only the data of task $\tau_i$, i.e., $\mathcal{D}_i$; (ii) Train from scratch (TFS), also the common used supervised learning manner, which initializes the parameter $\theta$ randomly and trains using $\mathcal{D}_i$; and (iii) Train on everything (TOE), also the multi-task learning manner, which jointly trains a single model on all available data so far following~\cite{finn2019online}. We choose various classic and outstanding methods in different scenarios as baselines for comparison, covering all types of meta-learning methods mentioned in \textbf{Subsection \ref{sec:pre_meta-learning}}. 
Meanwhile, we focus on multiple indicators, i.e., task performance, and model efficiency. All the meta-learning problems require some amount of adaptation to new tasks at test time. More details and full results are provided in \textbf{Appendices \ref{sec:app_3}-\ref{sec:app_5}}.

\begin{table}[t]
\caption{Performance comparison (MSE $\pm $ 95\% confidence interval) on sinusoid regression and pose prediction. The ``+TFS'' and ``+TOE'' indicate that the models are trained via the corresponding learning paradigm with the ResNet50 backbone.}
\label{tab:cap_1}
\begin{center}
\begin{small}
\begin{sc}
\resizebox{\linewidth}{!}{
\begin{tabular}{l|cc|cc}
\toprule
\multirow{2}{*}{\textbf{Model}}
    & \multicolumn{2}{c|}{\textbf{Sinusoid Reg.}} & \multicolumn{2}{c}{\textbf{Pose Pred.}} \\ 
    & \textbf{5-shot} & \textbf{10-shot} & \textbf{10-shot} & \textbf{15-shot}\\
\midrule
ResNet50+TFS & 2.37$\pm$0.30 & 1.98$\pm$0.11 & 5.81$\pm$0.29 & 3.97$\pm$0.19\\
ResNet50+TOE & 0.42$\pm$0.10 & 0.11$\pm$0.03 & 2.66$\pm$0.15 & 2.11$\pm$0.18\\
\midrule
MAML & 0.59$\pm$0.12 & 0.16$\pm$0.06 & 3.11$\pm$0.24  & 2.49$\pm$0.18\\
Reptile & 0.58$\pm$0.13 & 0.17$\pm$0.02 & 3.12$\pm$0.25 & 2.48$\pm$0.15\\
ProtoNet  & 0.57$\pm$0.14 & 0.18$\pm$0.03 & 3.14$\pm$0.21 & 2.46$\pm$0.16\\
Meta-Aug & 0.55$\pm$0.12 & 0.18$\pm$0.05 & 3.12$\pm$0.14 & 2.39$\pm$0.19\\
MR-MAML & 0.55$\pm$0.13 & 0.17$\pm$0.09 & 3.13$\pm$0.23 & 2.45$\pm$0.21\\
ANIL & 0.54$\pm$0.11 & 0.11$\pm$0.03  & 2.98$\pm$0.12 & 2.29$\pm$0.25\\
MetaSGD & 0.57$\pm$0.12 & 0.15$\pm$0.04 & 2.81$\pm$0.23 & 2.01$\pm$0.18\\
T-NET &0.56$\pm$0.12 & 0.11$\pm$0.04 & 2.84$\pm$0.17 & 2.71$\pm$0.22\\
TAML & 0.53$\pm$0.11 & 0.11$\pm$0.06 & 2.92$\pm$0.14 & 2.75$\pm$0.23\\
iMAML & 0.51$\pm$0.13 & 0.12$\pm$0.03 & 2.79$\pm$0.19 & 2.24$\pm$0.16\\
\rowcolor{orange!10}\textbf{NeuronML} & \textbf{0.49$\pm$0.13} & \textbf{0.10$\pm$0.04} & \textbf{2.76$\pm$0.20} & \textbf{1.99$\pm$0.15}\\
\bottomrule
\end{tabular}}
\end{sc}
\end{small}
\end{center}
\end{table}

\subsection{Regression}
\label{sec:7.1}
\paragraph{Experimental Setup}
We start with two regression problems, i.e., sinusoid regression and pose prediction. For sinusoid regression, we consider a more complex scenario following~\cite{jiang2022role}. Specifically, we conduct 680 tasks for training and testing, and each task involves regressing from the input to the output of a wave in the form of $A\sin w \cdot x + b + \epsilon $, where $A \in \left [ 0.1, 5.0 \right ] $, $w \in \left [ 0.5, 2.0 \right ]$, and $b \in \left [ 0,2\pi \right ] $. Thus, $\mathbb{P} (\mathcal{T})$ is continuous, and the amplitude and phase of the wave are varied between tasks. We add Gaussian noise with $\mu = 0$ and $\epsilon = 0.3$ to each data point sampled from the target task. For pose prediction, we conduct tasks for training and testing from the Pascal 3D dataset \cite{xiang2014beyond}. Specifically, we randomly select 50 objects for meta-training and 15 additional objects for meta-testing. In this experiment, we use the Mean Squared Error (MSE) as the evaluation metric, and use the ResNet-50 as the backbone. More details are provided in \textbf{Appendix \ref{sec:app_3}}.

\paragraph{Results and Analysis}
\textbf{Table \ref{tab:cap_1}} shows the comparison results between the models learned by NeuronML and the aforementioned three settings, i.e., TFT, TFS, and TOE. NeuronML achieves outstanding results on both datasets.
Results in \textbf{Figure \ref{fig:ex1_1}} and \textbf{Figure \ref{fig:app_ex1}} in the \textbf{Appendix \ref{sec:app_3}} show the adaptation results on the two regression tasks.
The results demonstrate that NeuronML adapts quickly and better with fewer neurons via 5 randomly sampled data points, while the other models hard to achieve this (affected by noise, with larger deviations of some points). 
For even fewer, adaptation results with 3 data points, our model is able to adequately fit such few data points without catastrophically overfitting. 
Crucially, when the data points are all located at half or even 1/4 of the input range, NeuronML can still infer amplitude and phase for a range of time into the future, while the inference ability of other models is much lower than that of NeuronML. This further demonstrates the superiority of NeuronML, especially for complex tasks requiring inference.

\begin{figure*}[t]
     \centering
     \subfigure[Quantitative regression results]{\includegraphics[width=0.48\linewidth]{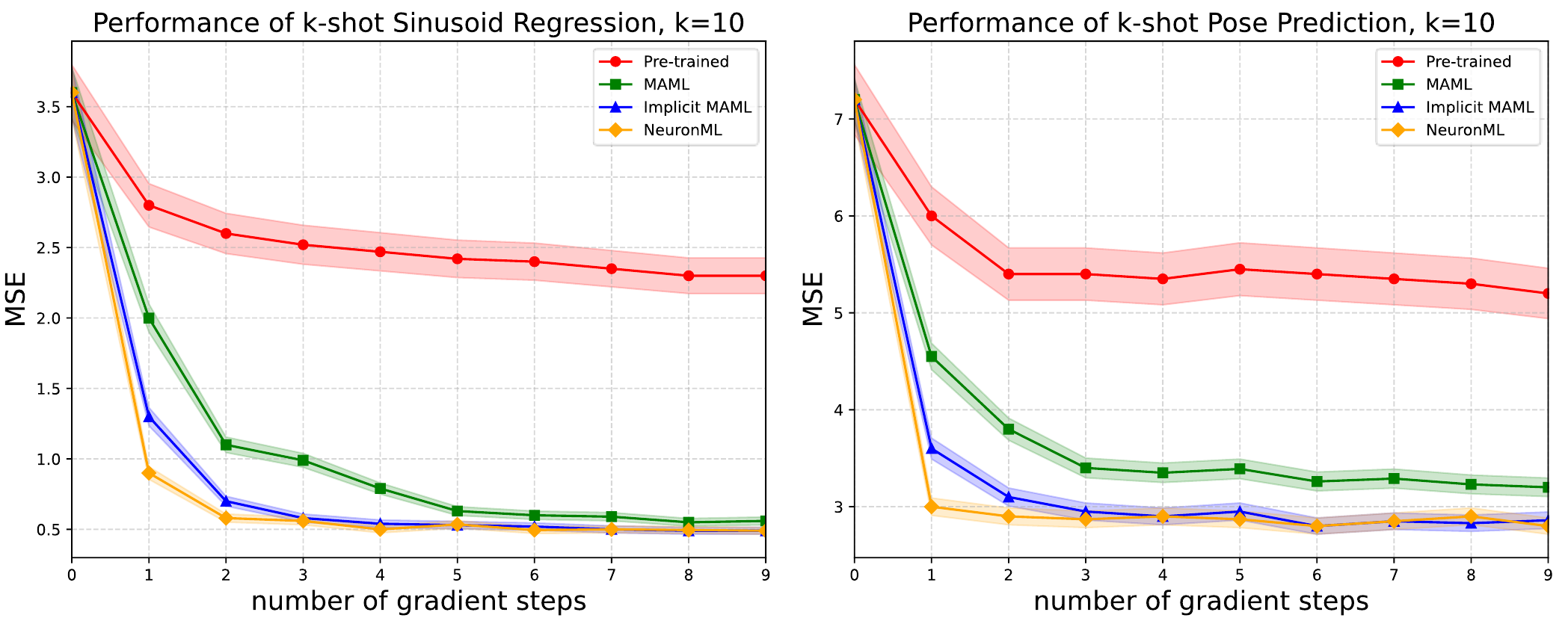}\label{fig:quantitative}}
    \subfigure[Adaptation curves and trade-off performance]{\includegraphics[width=0.48\linewidth]{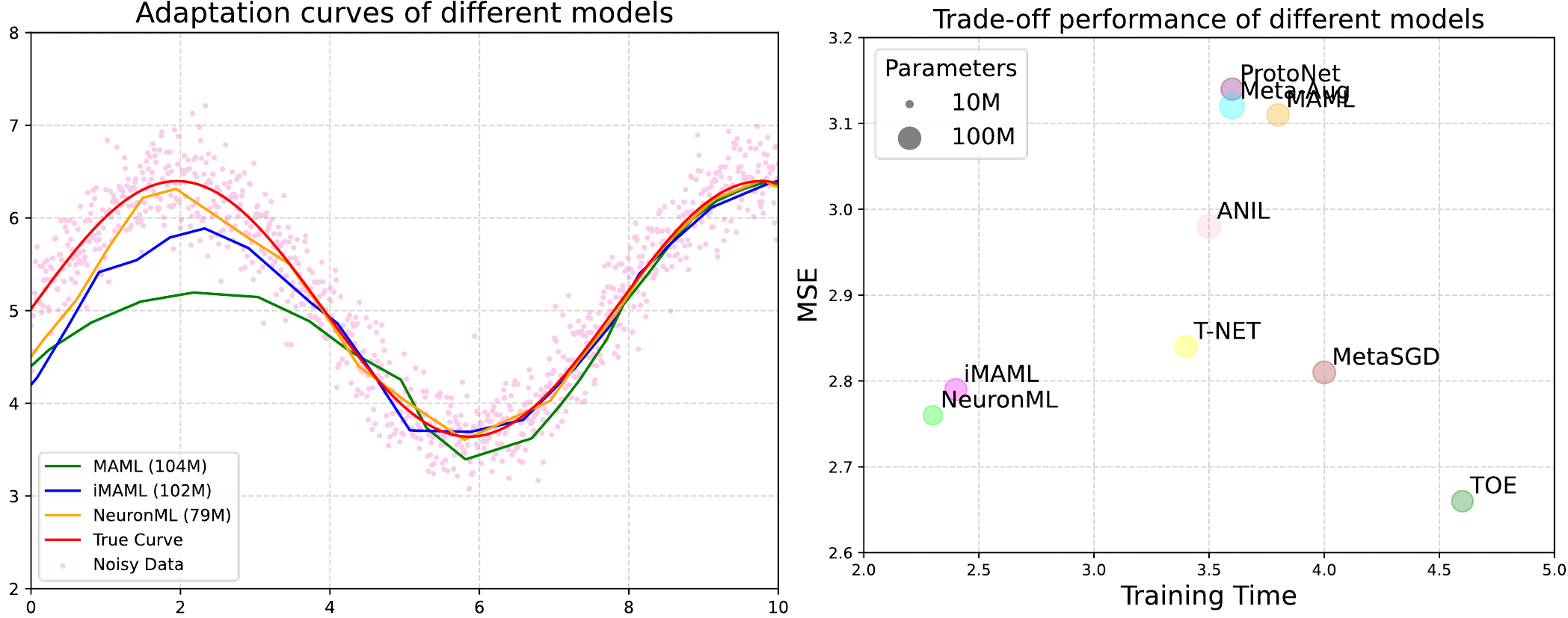}\label{fig:standard}}
    \caption{Adaptation for regression. (a) Quantitative results show the learning curves of different models at meta-test-time, where NeuronML achieves the best MSE score with fewer update steps. (b) Adaptation curves and trade-off performance. Left: adaptation curves of different models (steps = 3) on sinusoid regression. Right: the trade-off performance of different models on pose prediction.
    We randomly select tasks for learning, and NeuronML can better fit the waveform under limited update conditions with fewer parameters.}
    \label{fig:ex1_1}
\end{figure*}

\begin{table*}[t]
\caption{Performance (accuracy $\pm 95\%$ confidence interval) of SFSL classification on 20-way 1-shot (20,1) \& 20-way 5-shot (20,5) Omniglot, 5-way 1-shot (5,1) \& 5-way 5-shot (5,5) miniImagenet, 5-way 1-shot (5,1) \& 5-way 5-shot (5,5) tieredImagenet, and 5-way 1-shot (5,1) \& 5-way 5-shot (5,5) CIFAR-FS. The best results are highlighted in \textbf{bold}.}
\label{tab:ex2_1}
\begin{center}
\begin{small}
\begin{sc}
\resizebox{0.95\linewidth}{!}{
\begin{tabular}{l|c|cc|cc|cc|cc}
\toprule
\multirow{2}{*}{\textbf{Model}} & \multirow{2}{*}{\textbf{Backbone}}
    &  \multicolumn{2}{c|}{\textbf{Omniglot}} & \multicolumn{2}{c|}{\textbf{miniImagenet}} & \multicolumn{2}{c|}{\textbf{tieredImagenet}} & \multicolumn{2}{c}{\textbf{CIFAR-FS}}\\ 
    & & \textbf{(20,1)} & \textbf{(20,5)} & \textbf{(5,1)} & \textbf{(5,5)} & \textbf{(5,1)} & \textbf{(5,5)} & \textbf{(5,1)} & \textbf{(5,5)}\\
\midrule
Conv4+TFS & Conv4 & 43.2$\pm$0.9 & 52.1$\pm$1.1 & 25.2$\pm$0.9 & 33.4$\pm$0.8 & 26.5$\pm$0.9 & 34.6$\pm$0.5 & 29.4$\pm$1.2 & 38.7$\pm$1.3\\
ResNet50+TFS & ResNet50 & 43.4$\pm$0.8 & 52.2$\pm$0.9 & 25.4$\pm$0.7 & 33.3$\pm$0.7 & 26.8$\pm$1.3 & 35.5$\pm$1.4 & 29.3$\pm$1.6 & 39.2$\pm$1.2\\
DenseNet+TFS & DenseNet & 43.2$\pm$1.1 & 52.4$\pm$1.4 & 25.6$\pm$0.9 & 33.2$\pm$1.2 & 26.6$\pm$1.6 & 34.7$\pm$1.2 & 29.1$\pm$1.3 & 39.1$\pm$0.9\\
\midrule
Conv4+TOE & Conv4 & 89.2$\pm$0.7 & 94.1$\pm$0.3 & 49.2$\pm$1.3 & 65.9$\pm$1.1 &51.2$\pm$1.4  &67.3$\pm$1.2  &55.4$\pm$1.5  & 70.2$\pm$1.1\\
ResNet50+TOE & ResNet50 & 89.7$\pm$0.5 & 94.5$\pm$0.6 & 50.1$\pm$1.0 & 65.7$\pm$0.8 & 51.5$\pm$1.2 & 67.3$\pm$1.1 & 55.7$\pm$1.4 & 70.3$\pm$0.9\\
\midrule
MAML  & Conv4 & 86.2$\pm$0.5 & 92.3$\pm$0.2 & 42.6$\pm$1.6 & 61.4$\pm$0.8  & 48.7$\pm$0.5 & 64.3$\pm$0.7 & 53.6$\pm$0.9 & 67.9$\pm$1.1\\
ProtoNet  & Conv4  & 88.4$\pm$0.6 & 93.2$\pm$0.1 & 43.4$\pm$1.2 & 59.6$\pm$0.9 & 48.5$\pm$1.4 & 65.3$\pm$1.6 & 52.8$\pm$1.2 & 66.4$\pm$1.1\\
Meta-Aug & Conv4 & 89.6$\pm$0.4 & 94.4$\pm$0.2 & 44.2$\pm$1.7 & 64.3$\pm$0.9 & 49.3$\pm$0.8 & 66.2$\pm$0.9 & 51.6$\pm$0.9 & 67.1$\pm$0.7\\
MR-MAML & Conv4 & 88.3$\pm$0.5 & 95.2$\pm$0.2 & 44.4$\pm$0.9 & 64.7$\pm$0.8 & 48.5$\pm$0.4 & 66.4$\pm$0.8 & 53.2$\pm$0.9 & 69.2$\pm$0.8\\
ANIL & Conv4 & 88.7$\pm$0.6 & 94.8$\pm$0.2 & 47.7$\pm$1.5 & 61.3$\pm$0.7  & 50.4$\pm$0.9 & 69.5$\pm$1.3 & 51.2$\pm$0.8 & 71.6$\pm$1.1\\
MetaSGD  & Conv4 & 86.2$\pm$0.4 & 95.4$\pm$0.1 & 42.2$\pm$1.6 & 62.5$\pm$0.9 & 47.7$\pm$1.3 & 64.3$\pm$1.6 & 55.2$\pm$1.1 & 68.7$\pm$0.9\\
T-NET  & Conv4 & 86.6$\pm$0.3 & 94.5$\pm$0.1 & 52.8$\pm$1.5 & 66.2$\pm$0.9 & 54.7$\pm$1.1 & 67.8$\pm$1.3 & 55.3$\pm$1.4 & 70.2$\pm$1.2\\
\rowcolor{orange!10}NeuronML & Conv4 & \textbf{89.8$\pm$0.4} & \textbf{95.7$\pm$0.2} & \textbf{55.7$\pm$1.3} & \textbf{67.9$\pm$0.7} & \textbf{56.2$\pm$0.6} & \textbf{70.3$\pm$0.8} & \textbf{57.2$\pm$0.9} & \textbf{71.8$\pm$0.9}\\
\midrule
MAML & ResNet50 & 87.1$\pm$0.6 & 93.5$\pm$0.2 &  43.1$\pm$1.7 & 61.9$\pm$0.9 & 48.7$\pm$0.9 & 64.2$\pm$0.8 & 54.4$\pm$0.8 & 67.6$\pm$0.7\\
ProtoNet  & ResNet50 & 89.3$\pm$0.4 & 94.0$\pm$0.2 & 43.7$\pm$0.9 & 60.2$\pm$1.3 & 49.2$\pm$1.6 & 65.5$\pm$1.2 & 52.6$\pm$1.4 & 67.1$\pm$1.1\\
Meta-Aug & ResNet50  & 89.7$\pm$0.6 & 94.5$\pm$0.2 & 44.7$\pm$1.5 & 62.1$\pm$0.9 & 49.2$\pm$0.8 & 66.4$\pm$1.4 & 52.4$\pm$0.9 & 67.2$\pm$0.8\\
MR-MAML & ResNet50   & 89.2$\pm$0.5 & 95.0$\pm$0.2 & 45.2$\pm$1.6 & 65.4$\pm$0.9 & 48.7$\pm$1.2 & 67.1$\pm$1.3 & 53.5$\pm$1.4 & 69.4$\pm$0.8\\
ANIL & ResNet50 & 89.1$\pm$0.6 & \textbf{95.8$\pm$0.2} & 48.9$\pm$1.7 & 62.5$\pm$0.9 & 51.4$\pm$0.7 & 70.7$\pm$0.9 & 51.7$\pm$1.3 & 71.3$\pm$0.9\\
MetaSGD & ResNet50 & 87.8$\pm$0.6 & 95.5$\pm$0.2 & 51.9$\pm$1.3 & 62.2$\pm$0.8 & 47.2$\pm$0.9 & 64.1$\pm$1.1 & 55.4$\pm$0.9 & 68.6$\pm$1.2\\
T-NET & ResNet50 & 87.6$\pm$0.5 & 95.6$\pm$0.2 & 53.6$\pm$1.7 & 66.3$\pm$0.7 & 54.9$\pm$1.3 & 68.9$\pm$1.2 & 56.3$\pm$1.5 & 70.5$\pm$0.9\\
\rowcolor{orange!10}NeuronML & ResNet50 & \textbf{90.2$\pm$0.1} & \textbf{95.8$\pm$0.1} & \textbf{57.1$\pm$1.3} & \textbf{68.2$\pm$1.2} & \textbf{57.6$\pm$1.6} & \textbf{71.8$\pm$1.2} & \textbf{60.1$\pm$1.4} & \textbf{71.9$\pm$1.1}\\
\bottomrule
\end{tabular}
}
\end{sc}
\end{small}
\end{center}
\end{table*}

\begin{table}[t]
\caption{Performance (accuracy $\pm 95\%$ confidence interval) of CFSL classification. Here, we compare all the TFT models that are trained on miniImagenet and evaluated on CUB and Places. The best results are highlighted in \textbf{bold}.}
\label{tab:ex2_2}
\begin{center}
\begin{small}
\begin{sc}
\resizebox{\linewidth}{!}{
\begin{tabular}{l|c|cc|cc}
\toprule
\multirow{2}{*}{\textbf{Model}} & \multirow{2}{*}{\textbf{Backbone}}
    &  \multicolumn{2}{c|}{\textbf{miniImagenet$\to$CUB}} & \multicolumn{2}{c}{\textbf{miniImagenet$\to$Places}}\\ 
    & & \textbf{(5,5)} & \textbf{(5,20)} & \textbf{(5,5)} & \textbf{(5,20)}\\
\midrule
MAML & Conv4 & 33.6$\pm$0.1 & 42.5$\pm$0.4 & 29.8$\pm$0.5 & 36.1$\pm$0.8\\
ProtoNet  & Conv4 & 33.3$\pm$0.4 & 41.8$\pm$0.6 & 30.9$\pm$0.6 & 42.4$\pm$0.5\\
SCNAP & Conv4 & 41.9$\pm$0.2 & 53.6$\pm$0.3 & 31.2$\pm$0.5 & 56.2$\pm$0.8\\
MetaQDA & Conv4 & 48.7$\pm$0.6 & 57.4$\pm$0.4 & 47.2$\pm$0.5 & 56.2$\pm$0.8\\
Baseline++ & Conv4 & 39.2$\pm$0.7 & 49.7$\pm$0.6 & 42.2$\pm$0.4 & 50.1$\pm$0.5\\
S2M2 & Conv4 & 48.2$\pm$0.9 & 54.2$\pm$0.3 & 46.1$\pm$0.2 & 57.4$\pm$0.6\\
\rowcolor{orange!10}NeuronML & Conv4 & \textbf{49.3$\pm$0.8} & \textbf{58.1$\pm$0.5} & \textbf{47.3$\pm$0.4} & \textbf{57.9$\pm$0.4}\\
\midrule
MAML & ResNet50 & 33.8$\pm$0.2 & 43.8$\pm$0.4 & 29.6$\pm$0.7 & 36.2$\pm$0.4\\
ProtoNet  & ResNet50 & 34.7$\pm$0.4 & 43.6$\pm$0.3 & 31.4$\pm$0.5 & 43.8$\pm$0.7\\
SCNAP & ResNet50 & 42.2$\pm$0.3 & 54.2$\pm$0.4 & 32.8$\pm$0.3 & 40.2$\pm$0.4\\
ANIL & ResNet50 & 49.6$\pm$0.4 & 58.6$\pm$0.7 & 47.4$\pm$0.6 & 56.8$\pm$0.2\\
MetaQDA & ResNet50 & 39.1$\pm$0.5 & 49.7$\pm$0.3 & 43.8$\pm$0.4 & 49.5$\pm$0.3\\
Baseline++ & ResNet50 & 49.8$\pm$0.4 & 59.4$\pm$0.7 & 46.6$\pm$0.3 & 58.2$\pm$0.4\\
S2M2 & ResNet50 & 49.8$\pm$0.5 & 59.3$\pm$0.6 & 45.4$\pm$0.7 & 57.6$\pm$0.2\\
\rowcolor{orange!10}NeuronML & ResNet50 & \textbf{50.3$\pm$0.5} & \textbf{60.1$\pm$0.5} & \textbf{48.0$\pm$0.4} & \textbf{58.3$\pm$0.6}\\
\bottomrule
\end{tabular}}
\end{sc}
\end{small}
\end{center}
\end{table}

\begin{table}[t]
\caption{Performance (accuracy $\pm 95\%$ confidence interval) of MFSL classification on Meta-dataset. The ten datasets within Meta-dataset are divided into ID datasets and OOD datasets, and the overall results are obtained by training on all ten datasets. The best results are highlighted in \textbf{bold}.}
\label{tab:ex2_3}
\begin{center}
\begin{small}
\begin{sc}
\resizebox{\linewidth}{!}{
\begin{tabular}{l|c|ccc}
\toprule
\textbf{Model} & \textbf{Backbone} & \textbf{ID} & \textbf{OOD} & \textbf{Overall}\\
\midrule
Conv4+TOE & Conv4 & 60.5$\pm$0.3 & 21.3$\pm$0.3 & 40.1$\pm$0.4\\
ResNet50+TOE & ResNet50 & 62.3$\pm$0.1 & 20.9$\pm$0.2 &  44.4$\pm$0.4\\
DenseNet+TOE & DenseNet & 59.8$\pm$0.4 & 21.2$\pm$0.4 & 43.5$\pm$0.3\\
\midrule
MAML & Conv4 & 31.3$\pm$0.1 & 19.2$\pm$0.2 & 24.5$\pm$0.1\\
Reptile & Conv4 & 63.8$\pm$0.3 & 49.2$\pm$0.1 & 59.7$\pm$0.3\\
ProtoNet  & Conv4 & 42.2$\pm$0.2 & 30.1$\pm$0.1 & 37.8$\pm$0.2\\
MatchingNet & Conv4 & 65.7$\pm$0.2 & 55.6$\pm$0.2 & 60.2$\pm$0.3\\
CNAPs & Conv4 & 68.4$\pm$0.3 & 55.5$\pm$0.2 & 65.9$\pm$0.3\\
\rowcolor{orange!10}NeuronML & Conv4 & 68.6$\pm$0.4 & \textbf{55.9$\pm$0.1} & 57.3$\pm$0.5\\
\rowcolor{orange!10}NeuronML & ResNet & 69.3$\pm$0.2 & 55.6$\pm$0.2 & 58.0$\pm$0.4\\
\rowcolor{orange!10}NeuronML & DenseNet & \textbf{69.5$\pm$0.2} & 55.8$\pm$0.2 & \textbf{66.8$\pm$0.3}\\
\bottomrule
\end{tabular}}
\end{sc}
\end{small}
\end{center}
\end{table}

\subsection{Classification}
\label{sec:7.2}

\paragraph{Experimental Setup} For classification, we select various problems to evaluate the performance of NeuronML, i.e., standard few-shot learning (SFSL), cross-domain few-shot learning (CFSL), and multi-domain few-shot learning (MFSL). For SFSL, we select four benchmark datasets for evaluation, i.e., miniImagenet \cite{miniImagenet}, Omniglot \cite{Omniglot}, tieredImagenet \cite{tieredImagenet}, and CIFAR-FS \cite{CIFAR-FS}. For CFSL, we select two benchmark datasets for evaluation, i.e., CUB \cite{cub} and Places \cite{places}. The models in the CFSL experiment are trained on miniImagenet, and evaluated on CUB and Places. For MFSL, we choose the challenging Meta-dataset \cite{metadataset}, which contains both the in-domain (ID) and out-of-domain (OOD) datasets and is used to measure the model performance under different conditions \cite{wang2024towards}. We adhere to the most commonly used protocol mentioned in \cite{maml}, involving rapid learning of $N$-way classification with 1 and 5 shots. The setup is as follows: choose $N$ unseen classes, present the model with $K$ instances for each class, and assess the model's capacity to classify new instances of these $N$ classes. For the classifier, we select multiple networks, e.g., Conv4, ResNet, and DenseNet, as the backbones.
The loss function is the cross-entropy error between the predicted and true class. More details are provided in \textbf{Appendix \ref{sec:app_4}}.

\paragraph{Results and Analysis}

We list the results of SFSL, CFSL, and MFSL in \textbf{Table \ref{tab:ex2_1}}, \textbf{Table \ref{tab:ex2_2}}, and \textbf{Table \ref{tab:ex2_3}} respectively. From the results, we can observe that: (i) Compared with the latest results of the task also based on the TFT setting, the performance of the NeuronML model is stably higher than the previous method. This demonstrates the effectiveness of NeuronML. 
(ii) Compared with models trained under TFS and TOE settings, the NeuronML model achieves similar or even better results on each data set based on fewer update steps. In addition, for unknown tasks, TFS and TOE models need to be trained from scratch, and NeuronML gives the model powerful adjustment capabilities and can adapt well to new tasks. Next, from the results of CFSL, the NeuronML model still achieves comparable results, which shows that NeuronML can enable the model to learn a more accurate representation from the source domain and apply it to the target domain, proving its good applicability. Meanwhile, in MFSL, the model trained via NeuronML performs well on in-distribution (ID) and out-of-distribution (OOD) data, further proving its effectiveness and being able to well balance classes and tasks.

\begin{table}[t]
\caption{Mean return over 1000 test tasks for different models on MuJoCo. Standard deviation is presented over 30 seeds. The best results are highlighted in \textbf{bold}.}
\label{tab:ex3_1}
% \vskip 0.15in
% %\vspace{-0.1in}
\begin{center}
\begin{small}
\begin{sc}
\resizebox{\linewidth}{!}{
\begin{tabular}{l|ccc|ccc}
\toprule
\multicolumn{1}{c|}{\multirow{2}{*}{}} & \multicolumn{3}{c|}{\textbf{HalfCheetah}}  & \multicolumn{3}{c}{\textbf{HalfCheetah 10D-task}}    \\
\multicolumn{1}{c|}{}  & \textbf{Vel} & \textbf{Mass} & \textbf{Body}  & \textbf{(a)} & \textbf{(b)}  & \textbf{(c)} \\ \midrule
CeSoR  & $-1316 \pm 18$ & $1398 \pm 31$ & $1008 \pm 34$ & $1222 \pm 23$ & $1388 \pm 20$ & $1274 \pm 32$ \\
PAIRED & $-545 \pm 55$ & $662 \pm 30$ & $492 \pm 51$  & $551 \pm 53$  & $706 \pm 36$ & $561 \pm 65$  \\
CVaR-ML & $-574 \pm 22$  & $113 \pm 8$ & $193 \pm 6$        & $263 \pm 15$  & $250 \pm 11$ & $192 \pm 5$ \\
VariBAD  & $-82 \pm 2$ & $1558 \pm 32$   & $1616 \pm 28$   & $\pmb{1893 \pm 6}$ & $1984 \pm 67$ & $\pmb{1617 \pm 12}$ \\
PEARL & $-534 \pm 15$   & $1726 \pm 13$ & $1655 \pm 6$ & $1843 \pm 9$ & $1866 \pm 13$  & $1425 \pm 6$   \\
RoML (VariBAD)  & $-95 \pm 3$  & $1581 \pm 32$   & $1582 \pm 21$  & $1819 \pm 8$  & $1950 \pm 20$  & $1616 \pm 13$  \\ %\hline
RoML (PEARL)  & $-519 \pm 15$ & $1553 \pm 18$ & $1437 \pm 8$ & $1783 \pm 7$  & $1859 \pm 10$ & $1399 \pm 8$ \\ 
\rowcolor{orange!10}NeuronML (VariBAD)  & $\pmb{-78 \pm 7}$ & 1596 $\pm$ 28 & $1621 \pm 31$ & 1890 $\pm$ 8 & \textbf{1986 $\pm$ 55} & $\pmb{1617 \pm 10}$ \\
\rowcolor{orange!10}NeuronML (PEARL)  & $-528 \pm 9$ & $\pmb{1739 \pm 21}$ & $\pmb{1657 \pm 16}$ & 1830 $\pm$ 21 & 1859 $\pm$ 16 & 1401 $\pm$ 9 \\
\midrule
\multirow{2}{*}{} & \multicolumn{3}{c|}{\textbf{Humanoid}}   & \multicolumn{3}{c}{\textbf{Ant}}  \\
    & \textbf{Vel}   & \textbf{Mass}  & \textbf{Body}      & \textbf{Goal}   & \textbf{Mass}  & \textbf{Body}   \\ \midrule
VariBAD   & $880 \pm 4$  & $\pmb{1645 \pm 22}$ & $1678 \pm 17$ & $-229 \pm 3$  & $1473 \pm 3$  & $1476 \pm 1$  \\
RoML (VariBAD)  & $883 \pm 4$ & $1580 \pm 17$  & $1618 \pm 18$       & $-224 \pm 3$  & $1475 \pm 2$  & $1472 \pm 3$ \\ 
\rowcolor{orange!10}NeuronML (VariBAD)  & $\pmb{890 \pm 3}$ & $1630 \pm 19$ & $\pmb{1689 \pm 21}$ & $\pmb{-217 \pm 4}$ & $\pmb{1480 \pm 5}$ & $\pmb{1483 \pm 2}$ \\
\bottomrule
\end{tabular}}
\end{sc}
\end{small}
\end{center}
\vskip -0.1in
\end{table}

\begin{figure*}[t]
     \centering
     \subfigure[Performance on Khazad-Dum]{\includegraphics[width=0.48\linewidth]{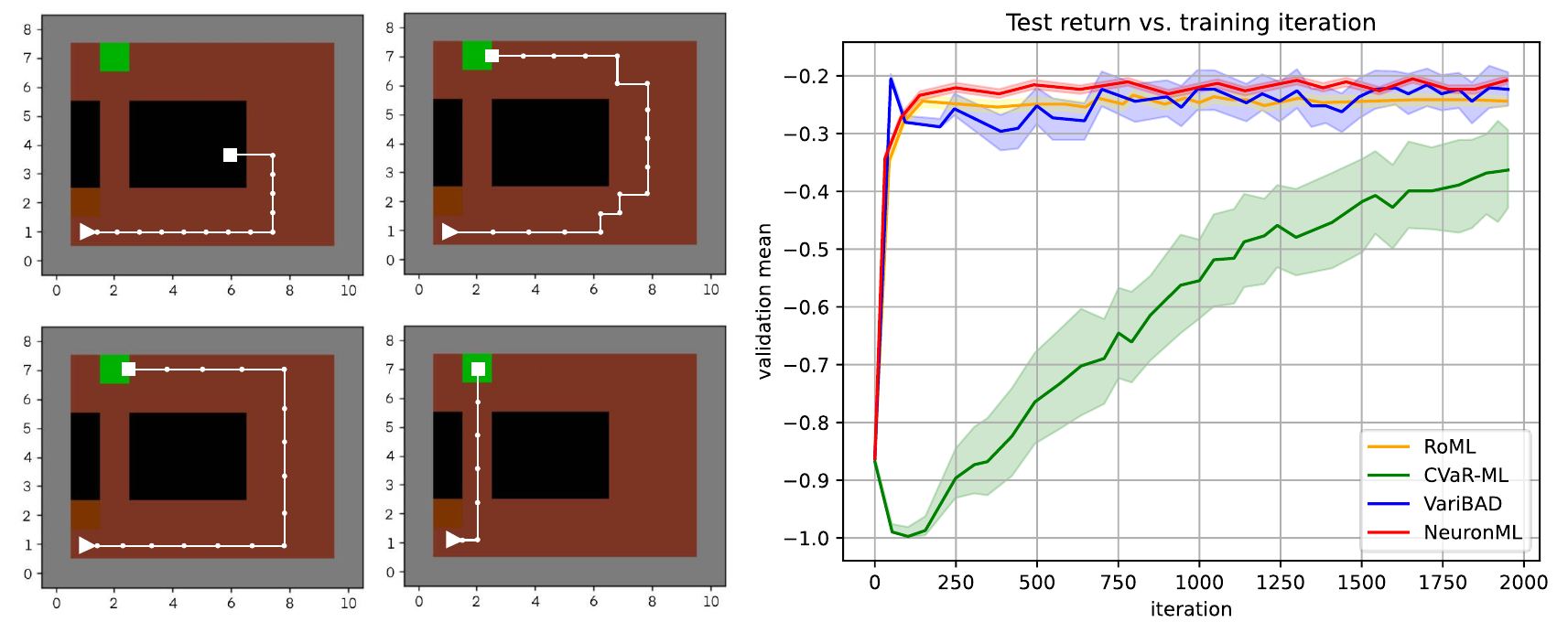}\label{fig:Khazad-Dum}}
    \subfigure[Performance on MoJoCo]{\includegraphics[width=0.48\linewidth]{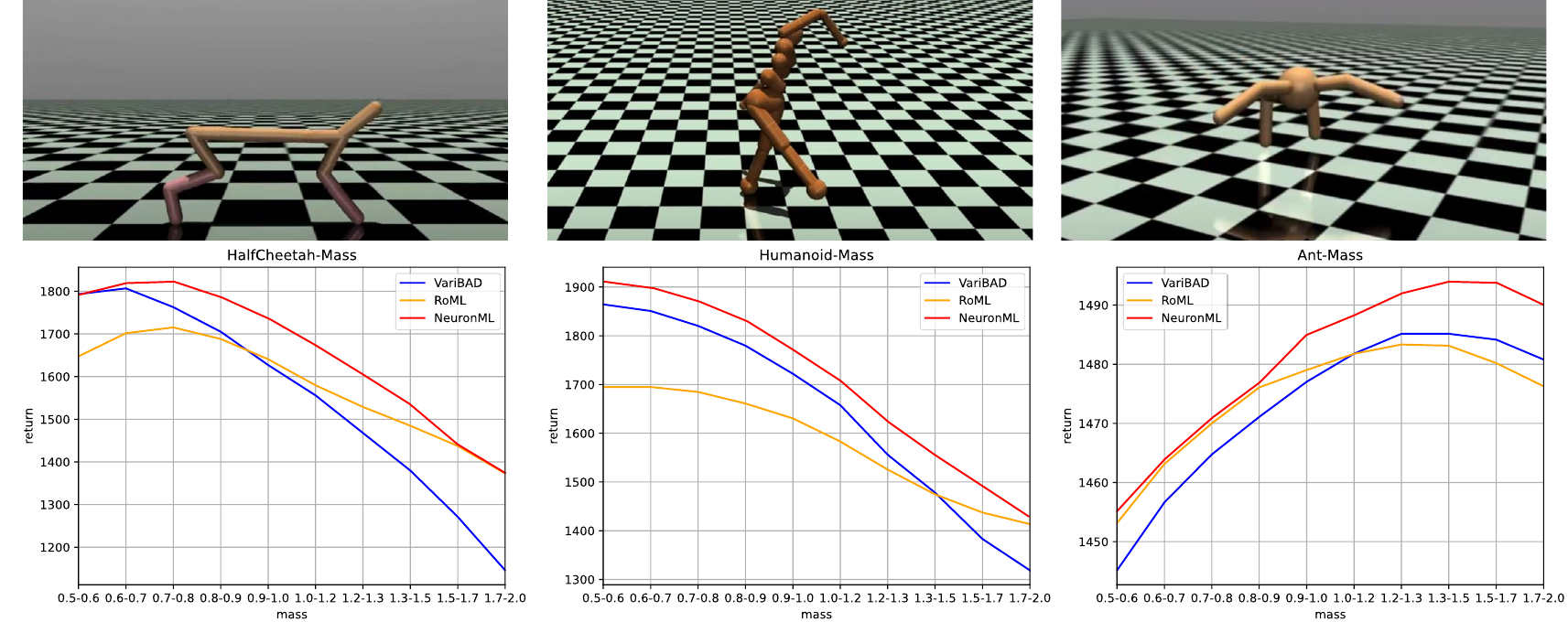}\label{fig:MoJoCo}}
    %\vspace{-0.1in}
    \caption{Reinforcement learning performance on Khazad-Dum and MoJoCo. (a) Performance on Khazad-Dum, where the Left shows sample episodes of different methods, and the Right shows the learning curves of test return vs. training iteration, with 95\% confidence intervals over 30 seeds. (b) Performance on MoJoCo, which shows the average returns per range of tasks in HalfCheetah, Hum, and Ant.}
    \label{fig:ex3_1}
    %\vspace{-0.1in}
\end{figure*}

\subsection{Reinforcement Learning}
\label{sec:7.3}
\paragraph{Experimental Setup}

To evaluate the performance of NeuronML on reinforcement learning problems, we adopt two benchmark environments, i.e., Khazad Dum \cite{greenberg2023train} and MuJoCo~\cite{mujoco}. Khazad Dum is used for evaluation in a discrete action space with randomness, where the agent starts from a random point in the left corner and reaches the green target without falling into the black abyss. The agent will encounter motion noise caused by rainfall during its movement, which is set to an exponential distribution with a risk level of 0.01. In the deployment, the agent action step size is fixed to 2 and trained based on $5\cdot10^6$ frames. On the other hand, we rely on MuJoCo ~\cite{mujoco} to evaluate the performance of NeuronML in high-dimensional, continuous action spaces, including three environments, i.e., HalfCheetah, Humanoid, and Ant. Each environment contains 3 tasks: (i) Velocity, corresponding to different position or speed targets; (ii) Mass, corresponding to different body weight; (3) Body, corresponding to different masses, head sizes, and physical damping levels. Additionally, we randomly sampled 10 numerical variables from \verb|env.model| in HalfCheetah and let them vary between tasks following \cite{greenberg2023train}, where the conducted datasets are denoted as HalfCheetah 10D-task \{a,b,c\}. More details are provided in \textbf{Appendix \ref{sec:app_5}}.

\paragraph{Results and Analysis}

\textbf{Figure \ref{fig:ex3_1}} shows the performance of NeuronML on Khazad Dum and MoJoCo, which compares the test return of the model with state-of-the-art (SOTA) and multiple outstanding baselines. From the results in Khazad Dum, we find that NeuronML achieves convergence much faster than baselines with acceptable feedback decreases, i.e., slightly lower than the optimal RoML but higher than or consistent with other baselines.
In the MuJoCo environment, we also report the rewards of NeuronML on four sets of tasks compared with multiple baselines in \textbf{Table \ref{tab:ex3_1}}. From the results, we can observe that: (i) NeuronML achieves comparable performance on all datasets; 
(ii) Combined with the results shown in \textbf{Figure \ref{fig:ex3_1}}, NeuronML’s return curve is as smooth as the curve of RoML (SOTA), which indicates that NeuronML acts robustly and performs better on high-risk tasks. More results and analysis are provided in \textbf{Appendix \ref{sec:app_5}}. These results further demonstrate the effectiveness of the proposed NeuronML.

\subsection{Ablation Studies}
\label{sec:7.4}

\begin{figure}
    \centering
    \includegraphics[width=1\linewidth]{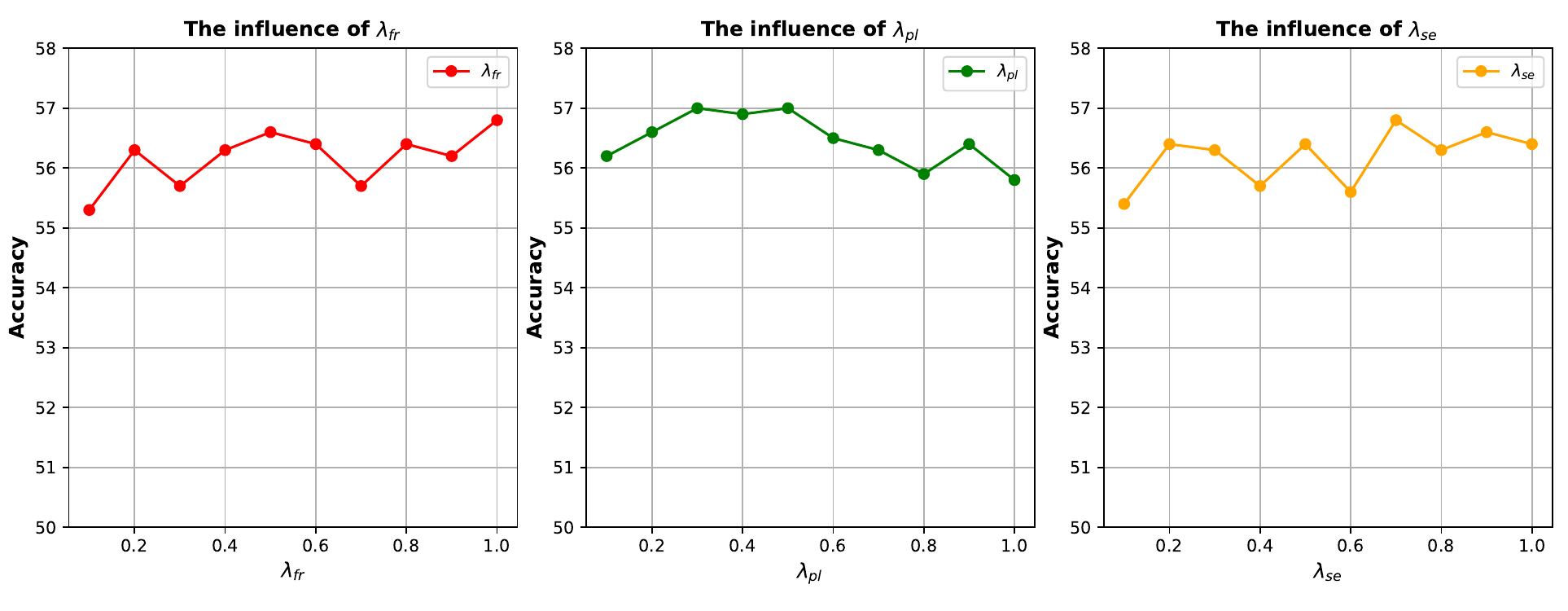}
    \caption{The influence of parameters $\lambda_{fr}$, $\lambda_{pl}$ and $\lambda_{se}$ on the results.}
    \label{fig:exp_para_sen}
\end{figure}

\paragraph{Parameter Sensitivity}
We conduct an experimental investigation of the NeuronML's hyperparameters. NeuronML uses three parameters $\lambda_{fr}$, $\lambda_{pl}$ and $\lambda_{se}$ to balance frugality, flexibility, and sensitivity respectively. Specifically, we vary the $\lambda_{fr}$, $\lambda_{pl}$ and $\lambda_{se}$ in the range of [0.1, 1.0] respectively, and record the classification accuracy of our method using a RESNET50 on miniImagenet dataset. Meanwhile, the weight decay is set at $10^{-4}$.
In the experiment, we fix two of the parameters to 0.5 and explore the effect of the other one on the performance, as shown in \textbf{Figure \ref{fig:exp_para_sen}}. The experimental results indicate that our method has minimal variation in accuracy, indicating that hyperparameter tuning is easy in practice. 

\paragraph{Constraint Analysis}
We analyze the impact of these three constraints. We fix the parameters of two of the constraints to 0.5, and set the parameter of the other one to 0. \textbf{Table \ref{tab:ConAna}} records the classification accuracy using a RESNET50 on miniImagenet dataset, ``w/o" stands for ``without".
The experimental results show the necessity of these three constraints.

\begin{table}[t]
\caption{The impact of different constraints on experimental results.}
\label{tab:ConAna}
\begin{center}
\begin{small}
\begin{sc}
    \resizebox{0.75\linewidth}{!}{
    \begin{tabular}{l|cc}
	\toprule
        \multirow{2}{*}{\textbf{Model}} & \multicolumn{2}{c}{\textbf{miniImagenet}} \\
        & \textbf{(5,1)} & \textbf{(5,5)}\\
	\midrule
        MAML & 43.1 $\pm$1.7 & 61.9$\pm$0.9\\
        \midrule
        Our w/o Frugality & 55.1$\pm$1.5 & 65.9 $\pm$1.0\\
        Our w/o Plasticity & 55.8$\pm$1.2 & 66.4 $\pm$0.9\\
        Our w/o Sensitivity & 52.7$\pm$1.5 & 62.3 $\pm$0.9\\
        \midrule
        NeuronML & \textbf{57.1$\pm$1.3} & \textbf{68.2 $\pm$1.2}\\
	\bottomrule
    \end{tabular}
    }
\end{sc}
\end{small}
\end{center}
\end{table}

\subsection{Discussion}
\label{sec:7.5}

\paragraph{Limitations} 

Although this work has been shown to achieve good results in various fields, it lacks validation on more complex cognitive scenarios, e.g., mathematical logic, due to the lack of a complete knowledge base. We will study more cases to further extend this work.

\paragraph{Broader Impact}

We believe that this work is a step towards general artificial intelligence. This technique introduces the biological BNS regulation mechanism into meta-learning, allowing the model to flexibly adjust its structure and be quickly applied to any problem in the real world. This technique provides an idea for the development of brain-like machine learning, which has many potential societal consequences. 

\paragraph{Amount of computing} 
All the results in this paper are produced by a machine with NVIDIA V100 GPUs or NVIDIA GeForce RTX 4090s. Meanwhile, all results are recorded as the average value after five experiments.

\section{Conclusion}
\label{sec:8}

Inspired by BNS, this paper explores the impact of FNS and proposes to define, measure, and model it in meta-learning. Firstly, through both empirical and theoretical analyses, we find that: (i) existing meta-learning methods rely on fixed structures for different tasks, affecting performance on different tasks; (ii) model performance is closely linked to network structure, with different task distributions requiring different structures. Thus, we conclude that FNS is crucial for meta-learning.
Based on the analysis, we propose that a good FNS in meta-learning should exhibit frugality, plasticity, and sensitivity. Further, to measure FNS in practice, we design three measurements corresponding to the three properties, which together form the structure constraint. Both theoretical and empirical evidence support the necessity of these properties and the reliability of the proposed measurements.
Building on this foundation, we propose NeuronML, a method that leverages bi-level optimization to optimize both the weights and structure with the structure constraint. NeuronML can be applied in various scenarios. Extensive theoretical and empirical evaluations confirm the superiority of NeuronML.

\section*{Data Availability}

The benchmark datasets can be downloaded from the literature cited in each subsection of \textbf{Section \ref{sec:7}}.

\section*{Conflict of interest}

The authors declare no conflict of interest.

% Can use something like this to put references on a page
% by themselves when using endfloat and the captionsoff option.
\ifCLASSOPTIONcaptionsoff
  \newpage
\fi

% trigger a \newpage just before the given reference
% number - used to balance the columns on the last page
% adjust value as needed - may need to be readjusted if
% the document is modified later
%\IEEEtriggeratref{8}
% The "triggered" command can be changed if desired:
%\IEEEtriggercmd{\enlargethispage{-5in}}

% references section

% can use a bibliography generated by BibTeX as a .bbl file
% BibTeX documentation can be easily obtained at:
% http://mirror.ctan.org/biblio/bibtex/contrib/doc/
% The IEEEtran BibTeX style support page is at:
% http://www.michaelshell.org/tex/ieeetran/bibtex/
%\bibliographystyle{IEEEtran}
% argument is your BibTeX string definitions and bibliography database(s)
%\bibliography{IEEEabrv,../bib/paper}
%
% <OR> manually copy in the resultant .bbl file
% set second argument of \begin to the number of references
% (used to reserve space for the reference number labels box)
% \begin{thebibliography}{1}

% \bibitem{IEEEhowto:kopka}
% H.~Kopka and P.~W. Daly, \emph{A Guide to \LaTeX}, 3rd~ed.\hskip 1em plus
%   0.5em minus 0.4em\relax Harlow, England: Addison-Wesley, 1999.

% \end{thebibliography}
% \input{references}
% \newpage
\bibliographystyle{IEEEtran}
\bibliography{reference}

% Generated by IEEEtran.bst, version: 1.14 (2015/08/26)
\begin{thebibliography}{10}
\providecommand{\url}[1]{#1}
\csname url@samestyle\endcsname
\providecommand{\newblock}{\relax}
\providecommand{\bibinfo}[2]{#2}
\providecommand{\BIBentrySTDinterwordspacing}{\spaceskip=0pt\relax}
\providecommand{\BIBentryALTinterwordstretchfactor}{4}
\providecommand{\BIBentryALTinterwordspacing}{\spaceskip=\fontdimen2\font plus
\BIBentryALTinterwordstretchfactor\fontdimen3\font minus \fontdimen4\font\relax}
\providecommand{\BIBforeignlanguage}[2]{{%
\expandafter\ifx\csname l@#1\endcsname\relax
\typeout{** WARNING: IEEEtran.bst: No hyphenation pattern has been}%
\typeout{** loaded for the language `#1'. Using the pattern for}%
\typeout{** the default language instead.}%
\else
\language=\csname l@#1\endcsname
\fi
#2}}
\providecommand{\BIBdecl}{\relax}
\BIBdecl

\bibitem{marder2012neuromodulation}
E.~Marder, ``Neuromodulation of neuronal circuits: back to the future,'' \emph{Neuron}, vol.~76, no.~1, pp. 1--11, 2012.

\bibitem{bear2020neuroscience}
M.~Bear, B.~Connors, and M.~A. Paradiso, \emph{Neuroscience: exploring the brain, enhanced edition: exploring the brain}.\hskip 1em plus 0.5em minus 0.4em\relax Jones \& Bartlett Learning, 2020.

\bibitem{de2021future}
D.~De~Ridder, J.~Maciaczyk, and S.~Vanneste, ``The future of neuromodulation: Smart neuromodulation,'' \emph{Expert Review of Medical Devices}, vol.~18, no.~4, pp. 307--317, 2021.

\bibitem{brines2005emerging}
M.~Brines and A.~Cerami, ``Emerging biological roles for erythropoietin in the nervous system,'' \emph{Nature Reviews Neuroscience}, vol.~6, no.~6, pp. 484--494, 2005.

\bibitem{avery2017neuromodulatory}
M.~C. Avery and J.~L. Krichmar, ``Neuromodulatory systems and their interactions: a review of models, theories, and experiments,'' \emph{Frontiers in neural circuits}, vol.~11, p. 108, 2017.

\bibitem{wang2023hacking}
J.~Wang, W.~Qiang, Y.~Ren, Z.~Song, J.~Zhang, and C.~Zheng, ``Hacking task confounder in meta-learning,'' \emph{arXiv preprint arXiv:2312.05771}, 2023.

\bibitem{bateni2020improved}
P.~Bateni, R.~Goyal, V.~Masrani, F.~Wood, and L.~Sigal, ``Improved few-shot visual classification,'' in \emph{Proceedings of the IEEE/CVF Conference on Computer Vision and Pattern Recognition}, 2020, pp. 14\,493--14\,502.

\bibitem{cesa2006prediction}
N.~Cesa-Bianchi and G.~Lugosi, \emph{Prediction, learning, and games}.\hskip 1em plus 0.5em minus 0.4em\relax Cambridge university press, 2006.

\bibitem{miconi2023learning}
T.~Miconi, ``Learning to acquire novel cognitive tasks with evolution, plasticity and meta-meta-learning,'' in \emph{International Conference on Machine Learning}.\hskip 1em plus 0.5em minus 0.4em\relax PMLR, 2023, pp. 24\,756--24\,774.

\bibitem{wang2023amsa}
J.~Wang, L.~Mou, L.~Ma, T.~Huang, and W.~Gao, ``Amsa: Adaptive multimodal learning for sentiment analysis,'' \emph{ACM Transactions on Multimedia Computing, Communications and Applications}, vol.~19, no.~3s, pp. 1--21, 2023.

\bibitem{finn2019online}
C.~Finn, A.~Rajeswaran, S.~Kakade, and S.~Levine, ``Online meta-learning,'' in \emph{International Conference on Machine Learning}.\hskip 1em plus 0.5em minus 0.4em\relax PMLR, 2019, pp. 1920--1930.

\bibitem{hospedales2021meta}
T.~Hospedales, A.~Antoniou, P.~Micaelli, and A.~Storkey, ``Meta-learning in neural networks: A survey,'' \emph{IEEE transactions on pattern analysis and machine intelligence}, vol.~44, no.~9, pp. 5149--5169, 2021.

\bibitem{protonet}
J.~Snell, K.~Swersky, and R.~Zemel, ``Prototypical networks for few-shot learning,'' \emph{Advances in neural information processing systems}, vol.~30, 2017.

\bibitem{reptile}
A.~Nichol and J.~Schulman, ``Reptile: a scalable metalearning algorithm,'' \emph{arXiv preprint arXiv:1803.02999}, vol.~2, no.~3, p.~4, 2018.

\bibitem{chen2022understanding}
L.~Chen, S.~Lu, and T.~Chen, ``Understanding benign overfitting in gradient-based meta learning,'' \emph{Advances in neural information processing systems}, vol.~35, pp. 19\,887--19\,899, 2022.

\bibitem{yao2021improving}
H.~Yao, L.-K. Huang, L.~Zhang, Y.~Wei, L.~Tian, J.~Zou, J.~Huang \emph{et~al.}, ``Improving generalization in meta-learning via task augmentation,'' in \emph{International conference on machine learning}.\hskip 1em plus 0.5em minus 0.4em\relax PMLR, 2021, pp. 11\,887--11\,897.

\bibitem{wang2024rethinking}
J.~Wang, W.~Qiang, J.~Li, L.~Si, and C.~Zheng, ``Rethinking meta-learning from a learning lens,'' \emph{arXiv preprint arXiv:2409.08474}, 2024.

\bibitem{lee2021meta}
J.~Lee, J.~Tack, N.~Lee, and J.~Shin, ``Meta-learning sparse implicit neural representations,'' \emph{Advances in Neural Information Processing Systems}, vol.~34, pp. 11\,769--11\,780, 2021.

\bibitem{schwarz2022meta}
J.~R. Schwarz and Y.~W. Teh, ``Meta-learning sparse compression networks,'' \emph{arXiv preprint arXiv:2205.08957}, 2022.

\bibitem{chin2020towards}
T.-W. Chin, R.~Ding, C.~Zhang, and D.~Marculescu, ``Towards efficient model compression via learned global ranking,'' in \emph{Proceedings of the IEEE/CVF conference on computer vision and pattern recognition}, 2020, pp. 1518--1528.

\bibitem{buciluǎ2006model}
C.~Buciluǎ, R.~Caruana, and A.~Niculescu-Mizil, ``Model compression,'' in \emph{Proceedings of the 12th ACM SIGKDD international conference on Knowledge discovery and data mining}, 2006, pp. 535--541.

\bibitem{miniImagenet}
O.~Vinyals, C.~Blundell, T.~Lillicrap, D.~Wierstra \emph{et~al.}, ``Matching networks for one shot learning,'' \emph{Advances in neural information processing systems}, vol.~29, 2016.

\bibitem{Omniglot}
B.~M. Lake, R.~Salakhutdinov, and J.~B. Tenenbaum, ``The omniglot challenge: a 3-year progress report,'' \emph{Current Opinion in Behavioral Sciences}, vol.~29, pp. 97--104, 2019.

\bibitem{tieredImagenet}
M.~Ren, E.~Triantafillou, S.~Ravi, J.~Snell, K.~Swersky, J.~B. Tenenbaum, H.~Larochelle, and R.~S. Zemel, ``Meta-learning for semi-supervised few-shot classification,'' \emph{arXiv preprint arXiv:1803.00676}, 2018.

\bibitem{CIFAR-FS}
L.~Bertinetto, J.~F. Henriques, P.~H. Torr, and A.~Vedaldi, ``Meta-learning with differentiable closed-form solvers,'' \emph{arXiv preprint arXiv:1805.08136}, 2018.

\bibitem{vgg16resnet50}
D.~Theckedath and R.~Sedamkar, ``Detecting affect states using vgg16, resnet50 and se-resnet50 networks,'' \emph{SN Computer Science}, vol.~1, no.~2, p.~79, 2020.

\bibitem{vyas2020learning}
N.~Vyas, S.~Saxena, and T.~Voice, ``Learning soft labels via meta learning,'' \emph{arXiv preprint arXiv:2009.09496}, 2020.

\bibitem{conv}
N.~S{\"u}nderhauf, S.~Shirazi, F.~Dayoub, B.~Upcroft, and M.~Milford, ``On the performance of convnet features for place recognition,'' in \emph{2015 IEEE/RSJ international conference on intelligent robots and systems (IROS)}.\hskip 1em plus 0.5em minus 0.4em\relax IEEE, 2015, pp. 4297--4304.

\bibitem{maml}
C.~Finn, P.~Abbeel, and S.~Levine, ``Model-agnostic meta-learning for fast adaptation of deep networks,'' in \emph{International conference on machine learning}.\hskip 1em plus 0.5em minus 0.4em\relax PMLR, 2017, pp. 1126--1135.

\bibitem{vanschoren2018meta}
J.~Vanschoren, ``Meta-learning: A survey,'' \emph{arXiv preprint arXiv:1810.03548}, 2018.

\bibitem{wang2024comprehensive}
J.~Wang, ``A comprehensive survey on meta-learning: Applications, advances, and challenges,'' \emph{Authorea Preprints}, 2024.

\bibitem{lee2019meta}
K.~Lee, S.~Maji, A.~Ravichandran, and S.~Soatto, ``Meta-learning with differentiable convex optimization,'' in \emph{Proceedings of the IEEE/CVF conference on computer vision and pattern recognition}, 2019, pp. 10\,657--10\,665.

\bibitem{abbas2022sharp}
M.~Abbas, Q.~Xiao, L.~Chen, P.-Y. Chen, and T.~Chen, ``Sharp-maml: Sharpness-aware model-agnostic meta learning,'' in \emph{International Conference on Machine Learning}.\hskip 1em plus 0.5em minus 0.4em\relax PMLR, 2022, pp. 10--32.

\bibitem{jeong2020ood}
T.~Jeong and H.~Kim, ``Ood-maml: Meta-learning for few-shot out-of-distribution detection and classification,'' \emph{Advances in Neural Information Processing Systems}, vol.~33, pp. 3907--3916, 2020.

\bibitem{wang2024meta}
J.~Wang, Y.~Tian, Y.~Yang, X.~Chen, C.~Zheng, and W.~Qiang, ``Meta-auxiliary learning for micro-expression recognition,'' \emph{arXiv preprint arXiv:2404.12024}, 2024.

\bibitem{chi2022metafscil}
Z.~Chi, L.~Gu, H.~Liu, Y.~Wang, Y.~Yu, and J.~Tang, ``Metafscil: a meta-learning approach for few-shot class incremental learning,'' in \emph{Proceedings of the IEEE/CVF Conference on Computer Vision and Pattern Recognition}, 2022, pp. 14\,166--14\,175.

\bibitem{javed2019meta}
K.~Javed and M.~White, ``Meta-learning representations for continual learning,'' \emph{Advances in neural information processing systems}, vol.~32, 2019.

\bibitem{koch2015siamese}
G.~Koch, R.~Zemel, R.~Salakhutdinov \emph{et~al.}, ``Siamese neural networks for one-shot image recognition,'' in \emph{ICML deep learning workshop}, vol.~2, no.~1.\hskip 1em plus 0.5em minus 0.4em\relax Lille, 2015.

\bibitem{vinyals2016matching}
O.~Vinyals, C.~Blundell, T.~Lillicrap, D.~Wierstra \emph{et~al.}, ``Matching networks for one shot learning,'' \emph{Advances in neural information processing systems}, vol.~29, 2016.

\bibitem{relationnet}
F.~Sung, Y.~Yang, L.~Zhang, T.~Xiang, P.~H. Torr, and T.~M. Hospedales, ``Learning to compare: Relation network for few-shot learning,'' in \emph{Proceedings of the IEEE conference on computer vision and pattern recognition}, 2018, pp. 1199--1208.

\bibitem{chen2021exploring}
X.~Chen and K.~He, ``Exploring simple siamese representation learning,'' in \emph{Proceedings of the IEEE/CVF conference on computer vision and pattern recognition}, 2021, pp. 15\,750--15\,758.

\bibitem{wang2024image}
J.~Wang, L.~Mou, C.~Zheng, and W.~Gao, ``Image-based freeform handwriting authentication with energy-oriented self-supervised learning,'' \emph{arXiv preprint arXiv:2408.09676}, 2024.

\bibitem{hu2022unsupervised}
Z.~Hu, Z.~Li, X.~Wang, and S.~Zheng, ``Unsupervised descriptor selection based meta-learning networks for few-shot classification,'' \emph{Pattern Recognition}, vol. 122, p. 108304, 2022.

\bibitem{bartler2022mt3}
A.~Bartler, A.~B{\"u}hler, F.~Wiewel, M.~D{\"o}bler, and B.~Yang, ``Mt3: Meta test-time training for self-supervised test-time adaption,'' in \emph{International Conference on Artificial Intelligence and Statistics}.\hskip 1em plus 0.5em minus 0.4em\relax PMLR, 2022, pp. 3080--3090.

\bibitem{zhang2021shallow}
X.~Zhang, D.~Meng, H.~Gouk, and T.~M. Hospedales, ``Shallow bayesian meta learning for real-world few-shot recognition,'' in \emph{Proceedings of the IEEE/CVF International Conference on Computer Vision}, 2021, pp. 651--660.

\bibitem{grant2018recasting}
E.~Grant, C.~Finn, S.~Levine, T.~Darrell, and T.~Griffiths, ``Recasting gradient-based meta-learning as hierarchical bayes,'' \emph{arXiv preprint arXiv:1801.08930}, 2018.

\bibitem{myers2021bayesian}
V.~Myers and N.~Sardana, ``Bayesian meta-learning through variational gaussian processes,'' \emph{arXiv preprint arXiv:2110.11044}, 2021.

\bibitem{gordon2019meta}
J.~Gordon, J.~Bronskill, M.~Bauer, S.~Nowozin, and R.~Turner, ``Meta-learning probabilistic inference for prediction,'' in \emph{International Conference on Learning Representations (ICLR 2019)}.\hskip 1em plus 0.5em minus 0.4em\relax OpenReview. net, 2019.

\bibitem{tian2020meta}
H.~Tian, B.~Liu, X.-T. Yuan, and Q.~Liu, ``Meta-learning with network pruning,'' in \emph{Computer Vision--ECCV 2020: 16th European Conference, Glasgow, UK, August 23--28, 2020, Proceedings, Part XIX 16}.\hskip 1em plus 0.5em minus 0.4em\relax Springer, 2020, pp. 675--700.

\bibitem{tseng2020regularizing}
H.-Y. Tseng, Y.-W. Chen, Y.-H. Tsai, S.~Liu, Y.-Y. Lin, and M.-H. Yang, ``Regularizing meta-learning via gradient dropout,'' in \emph{Proceedings of the Asian Conference on Computer Vision}, 2020.

\bibitem{finn2018meta}
C.~Finn and S.~Levine, ``Meta-learning and universality: Deep representations and gradient descent can approximate any learning algorithm,'' in \emph{International Conference on Learning Representations}, 2018.

\bibitem{raghurapid}
A.~Raghu, M.~Raghu, S.~Bengio, and O.~Vinyals, ``Rapid learning or feature reuse? towards understanding the effectiveness of maml,'' in \emph{International Conference on Learning Representations}, 2019.

\bibitem{wang2020generalizing}
Y.~Wang, Q.~Yao, J.~T. Kwok, and L.~M. Ni, ``Generalizing from a few examples: A survey on few-shot learning,'' \emph{ACM computing surveys (csur)}, vol.~53, no.~3, pp. 1--34, 2020.

\bibitem{wang2024towards}
J.~Wang, W.~Qiang, X.~Su, C.~Zheng, F.~Sun, and H.~Xiong, ``Towards task sampler learning for meta-learning,'' \emph{International Journal of Computer Vision}, pp. 1--31, 2024.

\bibitem{hurley2009comparing}
N.~Hurley and S.~Rickard, ``Comparing measures of sparsity,'' \emph{IEEE Transactions on Information Theory}, vol.~55, no.~10, pp. 4723--4741, 2009.

\bibitem{donoho2006compressed}
D.~L. Donoho, ``Compressed sensing,'' \emph{IEEE Transactions on information theory}, vol.~52, no.~4, pp. 1289--1306, 2006.

\bibitem{hochba1997approximation}
D.~S. Hochba, ``Approximation algorithms for np-hard problems,'' \emph{ACM Sigact News}, vol.~28, no.~2, pp. 40--52, 1997.

\bibitem{peharz2012sparse}
R.~Peharz and F.~Pernkopf, ``Sparse nonnegative matrix factorization with l0-constraints,'' \emph{Neurocomputing}, vol.~80, pp. 38--46, 2012.

\bibitem{hastie2015statistical}
T.~Hastie, R.~Tibshirani, and M.~Wainwright, ``Statistical learning with sparsity,'' \emph{Monographs on statistics and applied probability}, vol. 143, no. 143, p.~8, 2015.

\bibitem{kempter1999hebbian}
R.~Kempter, W.~Gerstner, and J.~L. Van~Hemmen, ``Hebbian learning and spiking neurons,'' \emph{Physical Review E}, vol.~59, no.~4, p. 4498, 1999.

\bibitem{lisman1989mechanism}
J.~Lisman, ``A mechanism for the hebb and the anti-hebb processes underlying learning and memory.'' \emph{Proceedings of the National Academy of Sciences}, vol.~86, no.~23, pp. 9574--9578, 1989.

\bibitem{chen2021generalization}
Q.~Chen, C.~Shui, and M.~Marchand, ``Generalization bounds for meta-learning: An information-theoretic analysis,'' \emph{Advances in Neural Information Processing Systems}, vol.~34, pp. 25\,878--25\,890, 2021.

\bibitem{jiang2022role}
Y.~Jiang, Z.~Chen, K.~Kuang, L.~Yuan, X.~Ye, Z.~Wang, F.~Wu, and Y.~Wei, ``The role of deconfounding in meta-learning,'' in \emph{International Conference on Machine Learning}.\hskip 1em plus 0.5em minus 0.4em\relax PMLR, 2022, pp. 10\,161--10\,176.

\bibitem{xiang2014beyond}
Y.~Xiang, R.~Mottaghi, and S.~Savarese, ``Beyond pascal: A benchmark for 3d object detection in the wild,'' in \emph{IEEE winter conference on applications of computer vision}.\hskip 1em plus 0.5em minus 0.4em\relax IEEE, 2014, pp. 75--82.

\bibitem{cub}
P.~Welinder, S.~Branson, T.~Mita, C.~Wah, F.~Schroff, S.~Belongie, and P.~Perona, ``Caltech-ucsd birds 200,'' 2010.

\bibitem{places}
B.~Zhou, A.~Lapedriza, A.~Khosla, A.~Oliva, and A.~Torralba, ``Places: A 10 million image database for scene recognition,'' \emph{IEEE transactions on pattern analysis and machine intelligence}, vol.~40, no.~6, pp. 1452--1464, 2017.

\bibitem{metadataset}
E.~Triantafillou, T.~Zhu, V.~Dumoulin, P.~Lamblin, U.~Evci, K.~Xu, R.~Goroshin, C.~Gelada, K.~Swersky, P.-A. Manzagol \emph{et~al.}, ``Meta-dataset: A dataset of datasets for learning to learn from few examples,'' \emph{arXiv preprint arXiv:1903.03096}, 2019.

\bibitem{greenberg2023train}
I.~Greenberg, S.~Mannor, G.~Chechik, and E.~Meirom, ``Train hard, fight easy: Robust meta reinforcement learning,'' \emph{arXiv preprint arXiv:2301.11147}, 2023.

\bibitem{mujoco}
E.~Todorov, T.~Erez, and Y.~Tassa, ``Mujoco: A physics engine for model-based control,'' in \emph{2012 IEEE/RSJ international conference on intelligent robots and systems}.\hskip 1em plus 0.5em minus 0.4em\relax IEEE, 2012, pp. 5026--5033.

\bibitem{ripley2007pattern}
B.~D. Ripley, \emph{Pattern recognition and neural networks}.\hskip 1em plus 0.5em minus 0.4em\relax Cambridge university press, 2007.

\bibitem{hoge2018primer}
M.~H{\"o}ge, T.~W{\"o}hling, and W.~Nowak, ``A primer for model selection: The decisive role of model complexity,'' \emph{Water Resources Research}, vol.~54, no.~3, pp. 1688--1715, 2018.

\bibitem{chwif2000simulation}
L.~Chwif, M.~R.~P. Barretto, and R.~J. Paul, ``On simulation model complexity,'' in \emph{2000 winter simulation conference proceedings (Cat. No. 00CH37165)}, vol.~1.\hskip 1em plus 0.5em minus 0.4em\relax IEEE, 2000, pp. 449--455.

\bibitem{har2011geometric}
S.~Har-Peled, \emph{Geometric approximation algorithms}.\hskip 1em plus 0.5em minus 0.4em\relax American Mathematical Soc., 2011, no. 173.

\bibitem{mohri2018foundations}
M.~Mohri, A.~Rostamizadeh, and A.~Talwalkar, \emph{Foundations of machine learning}.\hskip 1em plus 0.5em minus 0.4em\relax MIT press, 2018.

\bibitem{Meta-sgd}
Z.~Li, F.~Zhou, F.~Chen, and H.~Li, ``Meta-sgd: Learning to learn quickly for few-shot learning,'' \emph{arXiv preprint arXiv:1707.09835}, 2017.

\bibitem{wang2024gessl}
J.~Wang, W.~Qiang, and C.~Zheng, ``Explicitly modeling universality into self-supervised learning,'' \emph{arXiv preprint arXiv:2405.01053}, 2024.

\bibitem{byol}
J.-B. Grill, F.~Strub, F.~Altch{\'e}, C.~Tallec, P.~Richemond, E.~Buchatskaya, C.~Doersch, B.~Avila~Pires, Z.~Guo, M.~Gheshlaghi~Azar \emph{et~al.}, ``Bootstrap your own latent-a new approach to self-supervised learning,'' \emph{Advances in neural information processing systems}, vol.~33, pp. 21\,271--21\,284, 2020.

\bibitem{moco}
X.~Chen, H.~Fan, R.~Girshick, and K.~He, ``Improved baselines with momentum contrastive learning,'' \emph{arXiv preprint arXiv:2003.04297}, 2020.

\bibitem{wang2023unleash}
J.~Wang, Z.~Song, W.~Qiang, and C.~Zheng, ``Unleash model potential: Bootstrapped meta self-supervised learning,'' 2023.

\bibitem{ni2021close}
R.~Ni, M.~Shu, H.~Souri, M.~Goldblum, and T.~Goldstein, ``The close relationship between contrastive learning and meta-learning,'' in \emph{International Conference on Learning Representations}, 2021.

\bibitem{simclr}
T.~Chen, S.~Kornblith, M.~Norouzi, and G.~Hinton, ``A simple framework for contrastive learning of visual representations,'' in \emph{International conference on machine learning}.\hskip 1em plus 0.5em minus 0.4em\relax PMLR, 2020, pp. 1597--1607.

\bibitem{duan2016rl}
Y.~Duan, J.~Schulman, X.~Chen, P.~L. Bartlett, I.~Sutskever, and P.~Abbeel, ``{RL}$^2$: Fast reinforcement learning via slow reinforcement learning,'' \emph{arXiv preprint arXiv:1611.02779}, 2016.

\bibitem{liu2023simple}
E.~Z. Liu, S.~Suri, T.~Mu, A.~Zhou, and C.~Finn, ``Simple embodied language learning as a byproduct of meta-reinforcement learning,'' \emph{arXiv preprint arXiv:2306.08400}, 2023.

\bibitem{gupta2018meta}
A.~Gupta, R.~Mendonca, Y.~Liu, P.~Abbeel, and S.~Levine, ``Meta-reinforcement learning of structured exploration strategies,'' \emph{Advances in neural information processing systems}, vol.~31, 2018.

\bibitem{beck2023hypernetworks}
J.~Beck, M.~T. Jackson, R.~Vuorio, and S.~Whiteson, ``Hypernetworks in meta-reinforcement learning,'' in \emph{Conference on Robot Learning}.\hskip 1em plus 0.5em minus 0.4em\relax PMLR, 2023, pp. 1478--1487.

\bibitem{rajendran2020meta}
J.~Rajendran, A.~Irpan, and E.~Jang, ``Meta-learning requires meta-augmentation,'' \emph{Advances in Neural Information Processing Systems}, vol.~33, pp. 5705--5715, 2020.

\bibitem{yin2019meta}
M.~Yin, G.~Tucker, M.~Zhou, S.~Levine, and C.~Finn, ``Meta-learning without memorization,'' \emph{arXiv preprint arXiv:1912.03820}, 2019.

\bibitem{optimization_rapid}
A.~Raghu, M.~Raghu, S.~Bengio, and O.~Vinyals, ``Rapid learning or feature reuse? towards understanding the effectiveness of maml,'' \emph{arXiv preprint arXiv:1909.09157}, 2019.

\bibitem{tnet}
Y.~Lee and S.~Choi, ``Gradient-based meta-learning with learned layerwise metric and subspace,'' in \emph{International Conference on Machine Learning}.\hskip 1em plus 0.5em minus 0.4em\relax PMLR, 2018, pp. 2927--2936.

\bibitem{jamal2019task}
M.~A. Jamal and G.-J. Qi, ``Task agnostic meta-learning for few-shot learning,'' in \emph{Proceedings of the IEEE/CVF Conference on Computer Vision and Pattern Recognition}, 2019, pp. 11\,719--11\,727.

\bibitem{rajeswaran2019meta}
A.~Rajeswaran, C.~Finn, S.~M. Kakade, and S.~Levine, ``Meta-learning with implicit gradients,'' \emph{Advances in neural information processing systems}, vol.~32, 2019.

\bibitem{chen2019closer}
W.-Y. Chen, Y.-C. Liu, Z.~Kira, Y.-C.~F. Wang, and J.-B. Huang, ``A closer look at few-shot classification,'' \emph{arXiv preprint arXiv:1904.04232}, 2019.

\bibitem{mangla2020charting}
P.~Mangla, N.~Kumari, A.~Sinha, M.~Singh, B.~Krishnamurthy, and V.~N. Balasubramanian, ``Charting the right manifold: Manifold mixup for few-shot learning,'' in \emph{Proceedings of the IEEE/CVF winter conference on applications of computer vision}, 2020, pp. 2218--2227.

\bibitem{requeima2019fast}
J.~Requeima, J.~Gordon, J.~Bronskill, S.~Nowozin, and R.~E. Turner, ``Fast and flexible multi-task classification using conditional neural adaptive processes,'' \emph{Advances in Neural Information Processing Systems}, vol.~32, 2019.

\bibitem{cesor}
\BIBentryALTinterwordspacing
I.~Greenberg, Y.~Chow, M.~Ghavamzadeh, and S.~Mannor, ``Efficient risk-averse reinforcement learning,'' \emph{Advances in Neural Information Processing Systems}, 2022. [Online]. Available: \url{https://arxiv.org/abs/2205.05138}
\BIBentrySTDinterwordspacing

\bibitem{PAIRED}
M.~Dennis, N.~Jaques, E.~Vinitsky, A.~Bayen, S.~Russell, A.~Critch, and S.~Levine, ``Emergent complexity and zero-shot transfer via unsupervised environment design,'' \emph{Advances in neural information processing systems}, vol.~33, pp. 13\,049--13\,061, 2020.

\bibitem{varibad}
L.~Zintgraf, K.~Shiarlis, M.~Igl, S.~Schulze, Y.~Gal, K.~Hofmann, and S.~Whiteson, ``Varibad: A very good method for bayes-adaptive deep rl via meta-learning,'' in \emph{International Conference on Learning Representations}, 2019.

\bibitem{pearl}
K.~Rakelly, A.~Zhou, C.~Finn, S.~Levine, and D.~Quillen, ``Efficient off-policy meta-reinforcement learning via probabilistic context variables,'' in \emph{International conference on machine learning}.\hskip 1em plus 0.5em minus 0.4em\relax PMLR, 2019, pp. 5331--5340.

\end{thebibliography}

\newpage
% \appendix
\begin{appendices}
\section*{Appendix}
This supplementary material provides results for additional experiments and details to reproduce our results that could not be included in the paper submission due to space limitations.
\begin{itemize}
    \item Appendix \ref{sec:app_2} provides proofs and further analyses of the theorems in the text.
    \item Appendix \ref{sec:5} provides the specific instantiations of NeuronML, e.g., supervised learning, self-supervised learning, and reinforcement learning. The domains differ in the form of loss function and in how data is generated by the task, but the basic learning mechanism provided by NeuronML can be applied in all cases.
    \item Appendix \ref{sec:app_3} provides the additional details (datasets, baselines, and experimental setups), and full results for regression. 
    \item Appendix \ref{sec:app_4} provides the additional details (datasets, baselines, and experimental setups), and full results for classification.
    \item Appendix \ref{sec:app_5} provides the additional details (environments, baselines, and experimental setups), and full results for reinforcement learning.
\end{itemize}

% %\newpage
\section{Theoretical Analysis}
\label{sec:app_2}

\subsection{Proof of Theorem \ref{theorem:motivation_1}}

\renewcommand{\thetheorem}{III.1}
\begin{theorem}
    Let $f$ be any meta-learning model for the task of binary classification with respect to the 0-1 loss over the data $\mathcal{X}$ of task $ \tau $. Then, there exists a distribution $ P_\mathcal{X} $ over $ \mathcal{X} \times \{0, 1\} $ such that:
    \begin{itemize}
        \item There exists a function $\mathcal{F}:  \mathcal{X} \to \{0, 1\}$ with $\mathcal{L}_{P_\mathcal{X}}(\mathcal{F}) = 0$. 
        \item Let $ m  < \left | \mathcal{X} \right | /2 $ representing the size of the training set. over the choice of $\mathcal{X}^i\sim P_{\mathcal{X}}^m$, we have $P(\mathcal{L}_{P_{\mathcal{X}}}(f(\mathcal{X}^i)) \ge  1/8) \ge 1/7$.
    \end{itemize}
\end{theorem}

\proof
Consider a subset $S$ from $X$, where $| S | = n$. There are $T = 2^{n}$ possible functions for the binary classification task on $S$.
The core idea of the proof is that any meta-learning model that observes only half of the instances in $ \mathcal{S} $ lacks the necessary information to determine the labels of the remaining instances in $\mathcal{S} $.

Denote these possible functions by $ f_1, \ldots, f_T $. For any function $ f_i$, Let $P_{\mathcal{X}^i}$ be a distribution over $ \mathcal{S} \times \{0, 1\}$ defined as:
\begin{equation}\label{eq:proof_1_1}
    P_{\mathcal{X}^i}(x,y)=\left\{\begin{matrix}
 \frac{1}{|\mathcal{S} |}, \quad \text{if}\quad y=f_i(x) \\
0, \quad\quad otherwise
\end{matrix}\right.
\end{equation}
Here, The probability that the pair $(x,y)$ predicted by the function $f_i$ matches the true label is $1/|\mathcal{S}|$, and 0 otherwise. Thus, $\mathcal{L}_{P_{\mathcal{X}^i}}(f_i) = 0$. 

Let $S$ be a subset from $X$,  where $| S | = 2m$. There are $ k = (2m)^m $ possible sequences of $ m $ examples from $ \mathcal{S} $, denoted as $ \mathcal{S}_1, \cdots, \mathcal{S}_k $. For a sequence $ \mathcal{S}_j = (x_1, \cdots, x_m) $, we denote by $ \mathcal{S}_j^i $ the sequence where the instances in $ \mathcal{S}_j $ are labeled by the function $ f_i $, specifically, $ \mathcal{S}_j^i = ((x_1, f_i(x_1)), \ldots, (x_m, f_i(x_m))) $. If the distribution is $ P_{\mathcal{X}^i} $, then the possible training sets that $ A $ can receive are $ \mathcal{S}_1^i, \cdots, \mathcal{S}_k^i $, and all these training sets have the same probability of being sampled. Therefore, we get:
\begin{equation}
\begin{array}{l}
\underset{\mathcal{S}\sim  P_{\mathcal{X}^i}^m}{\max} \left [ \mathcal{L}_{ P_{\mathcal{X}^i}^m}(f(\mathcal{S} )) \right ] = \underset{i\in \left [ T \right ] }{\max} \frac{1}{k} \sum_{j=1}^{k} \mathcal{L}_{ P_{\mathcal{X}^i}^m}(f(\mathcal{S}_j^i ))\\[8pt] 
\ge \frac{1}{T} \sum_{i=1}^{T} \frac{1}{k} \sum_{j=1}^{k} \mathcal{L}_{ P_{\mathcal{X}^i}^m}(f(\mathcal{S}_j^i ))\\[8pt]
= \frac{1}{k} \sum_{j=1}^{k} \frac{1}{T} \sum_{i=1}^{T} \mathcal{L}_{ P_{\mathcal{X}^i}^m}(f(\mathcal{S}_j^i ))\\[8pt]
\ge \underset{j\in \left [ k \right ] }{\min}  \frac{1}{T} \sum_{i=1}^{T} \mathcal{L}_{ P_{\mathcal{X}^i}^m}(f(\mathcal{S}_j^i ))
\end{array}
\end{equation}
Next, fix part of $j \in \left [ k \right [$. Denote $\mathcal{S}_j = (x_1, \cdots, x_m)$ and let $(\hat{x}_1, . . . , \hat{x}_q)$ be the examples in $\mathcal{S}$ that do not appear in $\mathcal{S}_j^i$, where $q\ge m$. Thus, we get:
\begin{equation}
\begin{array}{l}
\frac{1}{T} \sum_{i=1}^{T} \mathcal{L}_{ P_{\mathcal{X}^i}^m}(f(\mathcal{S}_j^i )) \ge \frac{1}{T} \sum_{i=1}^{T} \frac{1}{2q} \sum_{p=1}^{q} \text{1}_{\left [ f(\mathcal{S}_j ^i)(\hat{x}_p)\ne f_i(\hat{x}_p )  \right ] } \\[8pt]
= \frac{1}{2q} \sum_{p=1}^{q} \frac{1}{T} \sum_{i=1}^{T}\text{1}_{\left [ f(\mathcal{S}_j ^i)(\hat{x}_p)\ne f_i(\hat{x}_p )  \right ] } \\[8pt]
\ge \underset{p \in \left [ q \right ] }{\min}  \frac{1}{T} \sum_{i=1}^{T}\text{1}_{\left [ f(\mathcal{S}_j ^i)(\hat{x}_p)\ne f_i(\hat{x}_p )  \right ] } \\[8pt]
\end{array}
\end{equation}
Next, fix some $ p \in [q] $, we can partition all the models (functions) $ f_1, \cdots, f_T $ into $ T / 2 $ disjoint pairs such that for each pair $ (f_i, f_i^{'}) $, we have $ f_i(s) \neq fi^{'}(s) $ for every $ s \in \mathcal{S} $ if and only if $ s = \hat{x}_p $. Since for such a pair we must have $ \mathcal{S}_j^i = \mathcal{S}_j^{i^{'}} $, it follows $\text{1}_{\left [ f(\mathcal{S}_j ^i)(\hat{x}_p)\ne f_i(\hat{x}_p )  \right ] }+\text{1}_{\left [ f(\mathcal{S}_j ^{i^{'}})(\hat{x}_p)\ne f_{i^{'}}(\hat{x}_p )  \right ] }=1$, which yields $\frac{1}{T}\sum_{i=1}^{T}  \text{1}_{\left [ f(\mathcal{S}_j ^i)(\hat{x}_p)\ne f_i(\hat{x}_p )  \right ] }=\frac{1}{2} $. Combining the above equations, we can demonstrate that for any algorithm $A$ which receives a training set of $M$ examples from $\mathcal{S} \times \{0, 1\}$ and returns a function $f(S) : \mathcal{S} \rightarrow \{0, 1\}$, the following holds:
\begin{equation}
    \underset{i\in \left [ T \right ] }{\max}\mathbb{E}_{\mathcal{S}\sim P_{\mathcal{X}^i}^m}\left [ \mathcal{L}_{P_{\mathcal{X}^i}}f(\mathcal{S} ) \right ] \ge\frac{1}{4}   
\end{equation}
This equation clearly implies that for any algorithm $ f^{'}$ which receives a training set of $ M $ examples from $ \mathcal{X} \times \{0, 1\} $, there exists a function $ f:  \mathcal{X} \rightarrow \{0, 1\} $ and a distribution $ P $ over $ \mathcal{X} \times \{0, 1\} $ such that $ \mathcal{L}_{P_{\mathcal{X}}}(f) = 0 $ and $\mathbb{E}_{\mathcal{S}\sim P_{\mathcal{X}}^m}\left [ \mathcal{L}_{P_{\mathcal{X}}}f^{'}(\mathcal{S} ) \right ] \ge\frac{1}{4} $. According to Markov's inequality, we can obtain:
\begin{equation}
P(\mathcal{L}_{P_{\mathcal{X}}}(f^{'}(\mathcal{X}^i)) \ge  1/8) \ge 1- \frac{1-1/4}{1-1/8} = 1/7
\end{equation}

\subsection{Proof of Theorem \ref{theorem:motivation_2}}

\renewcommand{\thetheorem}{III.2}
\begin{theorem}
    Suppose we have a set of candidate models $f^i, i=1,2,\dots, M$ and corresponding model parameters $\theta_i$, we aim to select the best model from them. Given a prior distribution $P(\theta_i|f^i)$ for $f^i$ and the dataset $\mathcal{D}_i$ for task $\tau_i$, to compare the model selection probabilities, we use Laplace approximation to estimate $P(\mathcal{D}_i|f^i)$:
    \begin{equation}
        \log P(\mathcal{D}_i|f^i)=\log P(\mathcal{D}_i|\hat{\theta}_i,f^i)-\frac{K_i}{2} \log N +\mathcal{O}(1)
    \end{equation}
    where $\hat{\theta}_i$ is a maximum likelihood estimate and $K_i$ denotes the number of free parameters in model $f^i$ that reflects structural frugality. Then, the posterior probability of each model $f^i$ for task $\tau_i$ can be estimated as:
        $\frac{e^{-\log P(\mathcal{D}_i|\hat{\theta}_i,f^i)}}{\sum_{j=1}^{M}e^{-\log P(\mathcal{D}_j|\hat{\theta}_j,f^j  )} } $,
    which is related to the network structure and neuron weight of $\theta_i$.
\end{theorem}

\proof
Before the detailed proof, we would like to introduce the Bayesian information criterion which is used for proving the model selection mentioned in this theorem. The Bayesian information criterion is applicable in settings where the fitting is carried out by maximization of a log-likelihood, i.e., $\mathcal{B}=-2\cdot \log \text{lik} + (\log N) \cdot d $. Further, under the Gaussian model, assuming the variance $\sigma^2_{\epsilon }$ is known, $-2\cdot \log \text{lik}$ equals $ {\textstyle \sum_{i}} (y_i-\hat{f}(x_i))^2/\sigma^2_{\epsilon } $, which is $N\cdot \text{err}/\sigma^2_{\epsilon }$ for squared error loss. Thus, we get:
\begin{equation}
    \mathcal{B}=\frac{N}{\sigma^2_{\epsilon }} \left [ \text{err}+(\log N)\cdot \frac{d}{N} \sigma^2_{\epsilon } \right ]  
\end{equation}
Thus, Bayesian information criterion tends to penalize complex models more heavily, giving preference to simpler models in selection, where $\sigma^2_{\epsilon }$ is typically estimated by the mean squared error of a low-bias model. 

Next, obtaining the preliminary, we turn to prove the theorem mentioned above. Reviewing the posterior probability of a given model:
    \begin{equation}
        \scalebox{0.85}{$P(f^i|\mathcal{D}_i) \propto P(f^i)\cdot P(\mathcal{D}_i|f^i ) \propto P(f^i)\cdot\int P(\mathcal{D}_i|\theta_i,f^i )P(\theta_i|f^i)d\theta_i$}
    \end{equation}
To compare two models $f^i$ and $f^j$, we form the posterior odds:
\begin{equation}
    \frac{P(f^i|\mathcal{D} )}{P(f^j|\mathcal{D})} = \frac{P(f^i)}{P(f^j)} \cdot \frac{P(\mathcal{D}|f^i)}{P(\mathcal{D}|f^j)} 
\end{equation}
If the odds are greater than one we choose model $f^i$, otherwise we choose model $f^j$. Here, the rightmost quantity is $\mathcal{B}_f(\mathcal{D})=\frac{P(\mathcal{D}|f^i)}{P(\mathcal{D}|f^j)} $. The standard practice is to presume an even distribution of prior probabilities across models, which implies that the probability of model $ f^i $, denoted as $P(f^i)$, remains constant. To estimate $P(\mathcal{D}|f^i)$, a method is required. Utilizing a Laplace approximation for the integral, along with additional simplifications as described by \cite{ripley2007pattern}, leads to the formulation presented in the standard form of Bayesian information criterion $\mathcal{B}=-2\cdot \log \text{lik} + (\log N) \cdot d $. Thus, we have:
\begin{equation}
     \log P(\mathcal{D}_i|f^i)=\log P(\mathcal{D}_i|\hat{\theta}_i,f^i)-\frac{K_i}{2} \log N +\mathcal{O}(1)
\end{equation}
where $\hat{\theta}_i$ is a maximum likelihood estimate and $K_i$ denotes the number of free parameters in model $f^i$ that reflects structural frugality. If we define our loss function to be $-2\log P(\mathcal{D}_i|\hat{\theta}_i,f^i)$ is equivalent to the equation of Bayesian information criterion. 

Consequently, opting for the model that yields the lowest Bayesian information criterion is akin to selecting the model with the highest (approximate) posterior probability. However, this framework extends beyond that. By calculating the Bayesian information criterion for a collection of $M$ models, resulting in $f^i$ for each $i=(1, 2,\cdots,M)$, we can then approximate the posterior probability of each individual model $f^i$ as $\frac{e^{-\log P(\mathcal{D}_i|\hat{\theta}_i,f^i)}}{\sum_{j=1}^{M}e^{-\log P(\mathcal{D}_j|\hat{\theta}_j,f^j  )} } $,
    which is related to the size and weight of the model parameters $\theta_i$. To this end, we conclude the results in Theorem \ref{theorem:motivation_2}.

\subsection{Proof of Theorem \ref{theorem:constraint_sparse}}
\label{sec_app:constraint_spa}
\renewcommand{\thetheorem}{IV.1}
\begin{theorem}[\textbf{Constraint of Frugality}]
Given the meta-learning model $f_\theta$ with parameter set $\theta$, an observation vector $\mathcal{Y} \in \mathbb{R}^n $ and a design matrix $\mathcal{X} \in \mathbb{R}^{n \times m}$, the goal of $f_\theta$ is to obtain $\mathcal{X}\theta=\mathcal{Y}$ with $\|\theta\|_0 \leq k$. Define a given subset $\mathcal{S}=\{1,\cdots,m\}$ with a convex cone $\mathbb{C}(\mathcal{S}^{\theta}):=\{\theta \in \mathbb{R}^{m}|\|\theta_{\mathcal{S}_{c}^{\theta}}\|_1\le \|\theta_{\mathcal{S}^{\theta}}\|_1\}$, if satisfies $Null(\mathcal{X})\cap \mathbb{C}(\mathcal{S}^{\theta})=\{0\}$, the basis pursuit relaxation of $\|\theta\|_0$ has a unique solution equal to $\|\theta\|_1$.
Then, for task $\tau_i$, we define the $\theta^i$ with frugality derived from the minimization of the subsequent objective:
    \begin{equation}
    \mathcal{L}_{fr}(\theta^i) = \|\theta^i\|_1,\text{s.t.} \|\theta^i\|_1 \leq \max\{C, \gamma \cdot d \cdot \log\left(\frac{N_i}{d}\right)\}
\end{equation}
where $C$ is a constant, $\gamma$ is a scaling factor, $N_i$ is the sample size for task $\tau_i$, and $d$ is the dimensionality of $\theta^i$.
\end{theorem}

\proof
We begin by considering the meta-learning model $f_\theta$ with parameter set $\theta$. Our objective is to solve the optimization problem defined by minimizing the $\ell_1$ norm of $\theta^i$ under the constraint that $\|\theta^i\|_0 \leq k$. Since the $\ell_0$ norm enforces frugality by counting the number of non-zero elements in a vector \cite{hurley2009comparing,donoho2006compressed}, we prove that in the mentioned restricted null space property, i.e., the only elements in cone $\mathbb{C}(\mathcal{S}^{\theta})$ that lie in the null space of $\mathcal{X}$ are all-zero vectors, the $\ell_0$-norm and $\ell_1$-norm are equivalent.

Recall the assumption, assuming that the design matrix $\mathcal{X}$ satisfies the restricted null space property, let $\hat{\theta} \in \mathbb{R}^p$ be an optimal solution to the basis pursuit linear program. We define the error vector $\Delta := \hat{\theta} - \theta^*$, where $\theta^*$ is the true parameter vector we aim to estimate.

To establish that $\Delta = 0$, we need to show that $\Delta$ lies in both the null space of $X$ and the convex cone defined by the support of $\theta^*$. First, since both $\theta^*$ and $\hat{\theta}$ are optimal solutions to the $\ell_0$ and $\ell_1$ problems, respectively, we have:
\begin{equation}
\mathcal{X}\theta^* = \mathcal{Y} = \mathcal{X}\hat{\theta}
\end{equation}
This implies:
\begin{equation}
\mathcal{X}\Delta = \mathcal{X}(\hat{\theta} - \theta^*) = 0
\end{equation}
Thus, we conclude that $\Delta \in \text{null}(\mathcal{X})$.

Next, we demonstrate that $\Delta$ also satisfies the conditions of the cone $\mathbb{C}(\mathcal{S}^{\theta})$. Since $\theta^*$ is feasible for the $\ell_1$ problem, the optimality of $\hat{\theta}$ guarantees:
\begin{equation}
\|\hat{\theta}\|_1 \leq \|\theta^*\|_1 = \|\theta^*_{\mathcal{S}^{\theta}}\|_1
\end{equation}
Expressing $\hat{\theta}$ as $\hat{\theta} = \theta^* + \Delta$, we write:
\begin{equation}
\|\theta^*_{\mathcal{S}^{\theta}}\|_1 \geq \|\hat{\theta}\|_1 = \|\theta^*_{\mathcal{S}^{\theta}} + \Delta_{\mathcal{S}^{\theta}}\|_1 + \|\Delta_{{\mathcal{S}^{\theta}_c}}\|_1
\end{equation}
Applying the triangle inequality, we obtain:
\begin{equation}
\|\theta^*_{\mathcal{S}^{\theta}}\|_1 \geq \|\theta^*_{\mathcal{S}^{\theta}}\|_1 - \|\Delta_{\mathcal{S}^{\theta}}\|_1 + \|\Delta_{\mathcal{S}^{\theta}_c}\|_1
\end{equation}
Rearranging these terms yields:
\begin{equation}
\|\Delta_{\mathcal{S}^{\theta}_c}\|_1 \leq \|\Delta_{\mathcal{S}^{\theta}}\|_1
\end{equation}
This indicates that $\Delta$ lies within the convex cone $\mathbb{C}(\mathcal{S}^{\theta})$.

Given that $\mathcal{X}$ satisfies the restricted null space property, the intersection of the null space and the cone condition implies that:
\begin{equation}
\Delta = 0
\end{equation}
This result shows that the unique solution to the basis pursuit problem confirms the necessity of the upper bound. 

Further, using results from high-dimensional statistics \cite{hastie2015statistical}, it is known that the number of non-zero parameters in the solution can be bounded as $k \leq \min\left\{ d, \frac{C}{\lambda} \right\}$,
where \( C \) is a constant that depends on the data and the loss function. In particular, for sufficiently large \( \lambda \), the solution becomes increasingly sparse.
To further refine this bound, we consider the sample size \( N_i \) and the dimensionality \( d \). Under certain assumptions about the data distribution and the design matrix, the sparsity level can be bounded by:
\[
 k \leq \gamma \cdot d \cdot \log\left(\frac{N_i}{d}\right),
\]
where \( \gamma \) is a scaling factor, \( N_i \) is the sample size for task \( \tau_i \), and \( d \) is the dimensionality of \( \theta^i \). This is also the balance term chose in our settiings.

Thus, our proof establishes that maintaining this bound ensures a valid and optimal sparse representation in the context of meta-learning models.

\subsection{Proof of Theorem \ref{theorem:constraint_plasticity}}
\label{sec_app:constraint_dyn}
\renewcommand{\thetheorem}{IV.2}
\begin{theorem}[\textbf{Constraint of Plasticity}]
Assume that for the matrix $ A(\theta) $, the minimal singular value $ \sigma_{\min}(A(\theta)) $ remains uniformly bounded below by a constant for all positive parameters. Denote subspace $ V_j $ for all $ \theta $, assume $\|A(\theta)\omega\| \leq \alpha_j \|x\|, \forall \omega \in V_j,$
   where $ \alpha_j $ is a positive parameter associated with the subspace $ V_j $. Then, we define the $\theta^i$ with plasticity derived from the minimization of the subsequent objective:
\begin{equation}
    \mathcal{L}_{pl}(\theta^i, \theta^j) = \sum_{j \neq i} \sum_{\omega \in \theta^i} \mathbbm{1}(\theta^i[\omega], \theta^j[\omega]) \cdot p_\omega
\end{equation}
where $ p_\omega = \frac{e^{\beta \mathcal{L}_\omega}}{\sum_{k} e^{\beta \mathcal{L}_k}} $ is the activation probability of neuron $\omega$, based on its importance assessed in historical tasks following Hebbian rules \cite{kempter1999hebbian,lisman1989mechanism}, $\theta^i[\omega]$ represents the neuron $\omega$ in $\theta^i$, and $\mathbbm{1}(\theta^i[\omega], \theta^j[\omega])$ is the indicator function that returns 1 if both the neurons of $\tau_i$ and $\tau_j$, i.e., $\theta^i$ and $\theta^j$ activate $\omega$.
\end{theorem}

\proof
To derive the mathematical formulas associated with the theorem, we start by considering the properties of the matrix $A(\theta)$ and the implications of the assumptions regarding its singular values and the behavior of the subspaces $V_j$.

Given that the minimal singular value $\sigma_{\min}(A(\theta))$ is bounded below by a constant $c > 0$, we have:
\begin{equation}
\sigma_{\min}(A(\theta)) \geq c > 0
\end{equation}
This condition ensures that the matrix $A(\theta)$ is stable and that it does not lose rank for any positive parameter $\theta$. Thus, $A(\theta)$ maintains full column rank for all applicable configurations.

Now, we focus on the assumption regarding the norm of $A(\theta)$ with respect to vectors $\omega$ in the subspace $V_j$:
\begin{equation}
\|A(\theta)\omega\| \leq \alpha_j \|x\|, \quad \forall \omega \in V_j
\end{equation}
Here, $\alpha_j$ is a positive parameter that quantifies how the mapping induced by $A(\theta)$ behaves on the subspace $V_j$.
These two assumptions provide the premise for the assessment of neuronal differences between tasks.

Next, we define the dynamic objective function $\mathcal{L}_{{pl}}(\theta^i, \theta^j)$:
\begin{equation}
\mathcal{L}_{pl}(\theta^i, \theta^j) = \sum_{j \neq i} \sum_{\omega=1}^{N_{\omega}} \mathbbm{1}(\theta^i[\omega], \theta^j[\omega]) \cdot p_\omega
\end{equation}
In this equation, the term $\mathbbm{1}(\theta^i[\omega], \theta^j[\omega])$ serves as an indicator function, contributing 1 to the sum when both tasks $\tau_i$ and $\tau_j$ activate the same neuron $\omega$. This quantifies the overlap of activated neurons between the tasks.

The activation probability $p_\omega$ for neuron $\omega$ is given by:
\begin{equation}
p_\omega = \frac{e^{\beta \mathcal{L}_\omega}}{\sum_{k} e^{\beta \mathcal{L}_k}}
\end{equation}
where $\mathcal{L}_\omega$ is the loss change caused by introducing or removing the neuron $\omega$. This formulation indicates that the more frequently a neuron is activated across historical tasks, the higher its probability of being considered important, which directly influences the dynamic behavior of the model.

Therefore, the dynamic objective $\mathcal{L}_{{pl}}(\cdot)$ balances the competing demands of activating fewer neurons across tasks while retaining important features from previous tasks.

Finally, as $\mathcal{L}_{{pl}}$ decreases, it indicates a reduction in the number of neurons that are jointly activated across tasks, leading to a higher dynamic state, which enhances the model's adaptability. So far, we have achieved the proof of the theorem.

Furthermore, we intuitively describe the insight and rationality of the constraints of the theorem to help a more intuitive understanding. The constraint described in the theorem implies that neurons with higher importance (i.e., those repeatedly activated across tasks) contribute more to the loss, \(\mathcal{L}_{pl}(\theta^i, \theta^j)\). 
Through this constraint, we may think that while this setup can improve model adaptability and task differentiation by reducing neuron redundancy, can it align with the goal of reinforcing neurons deemed important across multiple tasks? If the intention is to preserve such neurons for their generalization potential, why not directly prioritize selective reactivation of high-importance neurons without increasing the loss penalty? It is worth noting that directly constrain the model to repeat activations based on importance, may cause the model to fall into a trivial solution, i.e., the model's activated neurons will tend to be consistent, violating our requirement for plasticity. Thus, the current setup aligns with enhancing task specificity. Meanwhile, we introduce sensitivity to ensure that activated neurons share the model performance to make up for the gap of this constraint.

\subsection{Proof of Theorem \ref{theorem:constraint_sensitivity}}
\label{sec_app:constraint_eff}
\renewcommand{\thetheorem}{IV.3}
\begin{theorem}[\textbf{Constraint of Sensitivity}]
Let $(\theta, \mathcal{X})$ be a positive definite tuple, and let $s\colon \theta\longrightarrow(0,\infty)$ be an upper sensitivity function with total score $S$. Then, we define the $\theta^i$ with sensitivity derived from the minimization of the objective:
\begin{equation}
    \mathcal{L}_{se}(\theta^i) = \sum_{\omega=1}^{N_{\omega}} -\log\frac{s(\omega)}{S} \cdot \theta^i[\omega] 
\end{equation}
where $ s(\omega) = \frac{\partial \mathcal{L}(\theta^i, \tau_i)}{\partial \theta^i[\omega]} $ is the sensitivity score of neuron $\omega $ reflecting its impact on model performance of the corresponding task, and $\mathcal{L}(\theta^i, \tau_i)$ is the loss of model $f_{\theta^i}$ on task $\tau_i$.
There exists the sparse hyperparameter $C$ (\textbf{Theorem \ref{theorem:constraint_sparse}}) such that for any $0<\epsilon ,\delta<1$, if $C\geq \frac{S}{\epsilon ^2}\left(\mathsf{VC}\log S + \log\frac{1}{\delta}\right)$, with probability at least $1-\delta$, using the model $\hat{f}_{\theta}$ obtained via $\mathcal{L}_{se}(\theta^i)$ satisfies $\Big|\int f_\theta(x)\text{d} x - \int \hat{f}_\theta(x)\text{d} x \Big|\leq \epsilon$.
\end{theorem}

\proof

In this proof, we aim to show that by minimizing the objective \(\mathcal{L}_{{se}}(\theta^i)\), the model \(\hat{f}_{\theta}\) satisfies the inequality:
\[
\Big| \int f_\theta(x) \, \text{d} x - \int \hat{f}_\theta(x) \, \text{d} x \Big| \leq \epsilon.
\]
To this end, we leverage the properties of space combinations and define a triplet \((\theta, \hat{P}, \mathcal{R})\), where \((\theta, \hat{P})\) represents a probabilistic measure space, and \(\mathcal{R}\) is a subset of \(2^\theta\), consisting of \(P\)-measurable elements. Given a \(P\)-range space, the optimal relative error depends on a small additive parameter \(\varsigma > 0\).

For \(\epsilon, \varsigma \in (0, 1)\), a measure \(\mu\) is said to be a relative \((\epsilon, \varsigma)\)-approximation if for each \(R \in \mathcal{R}\), it holds that \(\mu\)-measurability is satisfied and the following inequality holds:
\[
\left| \hat{P}(R) - \mu(R) \right| \leq \epsilon \max(\varsigma, \hat{P}(R)).
\]

Combining the VC dimension of the parameter space, \cite{hoge2018primer, chwif2000simulation} showed that if the VC dimension is limited to \(\mathsf{VC}\), then sampling a sufficiently large sample from \(P\) produces a sample triplet that approximates the upper bound of relative error. When extending this result to the parameter space and the task of selecting neurons via sensitivity scores, constraining the contribution of neurons leads to a relative \((\epsilon, \varsigma)\)-approximation for \((\theta, \hat{P}, \mathcal{R})\).

Specifically, there exists a universal constant \(C\) such that for any \(\varsigma > 0\) and \(\epsilon, \delta \in (0, 1)\), with at least \(1 - \delta\) probability, the independent and identically distributed sample \(\Theta = \{\theta_1, \ldots, \theta_m\}\) drawn from \(P\) satisfies:
\[
\left| \hat{P}(R) - \frac{|\{r \in R \cap \Theta\}|}{m} \right| \leq \epsilon \max\left(\varsigma, \hat{P}(R)\right) \quad \forall R \in \mathcal{R}.
\]

To derive an unconditional relative error, we apply \(S\)-sensitive neurons, for which the upper bound sensitivity function \(s\) is defined, and the total sensitivity is \(S\). By setting \(\varsigma = \frac{1}{S}\), we obtain the unconditional relative error. Specifically, if the measure \(\mu\) is a relative \((\epsilon, \varsigma)\)-approximation, for \((\theta, \hat{P}, \iota(\mathcal{F}_{\mathcal{R}}, \succ))\), where \(\text{d} \hat{P}(\theta) = \frac{s(\theta)}{S} \, \text{d} P(\theta)\), then for each \(f \in \mathcal{R}\), we have:
\[
\Big| \int_{\theta} f(\theta) \, \text{d} P(\theta) - \int_{\theta} \frac{Sf(\theta)}{s(\theta)} \, \text{d} \mu(\theta) \Big| \leq (\epsilon + S\varsigma \epsilon) \int_{\theta} f(\theta) \, \text{d} P(\theta).
\]
Since \((\theta, \hat{P}, \iota(\mathcal{F}_{\mathcal{R}}, \succ))\) forms a \(\hat{P}\)-range, for \(\epsilon \in (0, 1)\) and \(\varsigma = \frac{1}{S}\), expanding \cite{har2011geometric}’s sampling result to this task, we can conclude that there exists a universal constant \(C\) such that for any sample \(\Theta = \{\theta_1, \ldots, \theta_m\}\) drawn independently from \(\hat{P}\), if $C\geq \frac{S}{\epsilon ^2}\left(\mathsf{VC}\log S + \log\frac{1}{\delta}\right)$, then with at least \(1 - \delta\) probability, we have:
\[
\left| \hat{P}(R) - \frac{|\{r \in R \cap \Theta\}|}{m} \right| \leq \frac{\epsilon}{2} \max\left(\frac{1}{S}, \hat{P}(R)\right) \quad \forall R \in \mathcal{R},
\]
which holds with probability at least \(1 - \delta\). This shows that the uniform probabilistic measure on \(\Theta\) serves as a relative \(\left(\frac{\epsilon}{2}, \frac{1}{S}\right)\)-approximation for \((\theta, \hat{P}, \iota(\mathcal{F}_{\mathcal{R}}, \succ))\). Applying the aforementioned lemma, we obtain that the error coefficient on the right-hand side is:
\[
\left(\frac{\epsilon}{2} + S \cdot \frac{1}{S} \cdot \frac{\epsilon}{2}\right) = \epsilon.
\]

Next, we prove the following integral relation:
\[
\int_{\theta} \min\{r, g(\theta)\} \, \text{d} \hat{P}(\theta) = \int_{0}^{r} \hat{P}_\succ(g, t) \, \text{d} t.
\]
By applying Fubini’s theorem, we can swap the order of integration over \(\theta\) and the range of the function. First, we introduce a useful shorthand notation:
\[
\hat{P}_\succ(g, t) = \hat{P}(\iota(g, \succ, t)) = \int_{\theta} \mathbbm{1}_{g(\theta) > t} \, \text{d} \hat{P}(\theta).
\]
We then perform the following transformation:
\[
\begin{aligned}
\int_{\theta} &\min\{r, g(\theta)\} \, \text{d} \hat{P}(\theta) \\
&= \int_{\theta} \left( \int_{0}^{\infty} \mathbbm{1}_{t < \min\{r, g(\theta)\}} \, \text{d} t \right) \, \text{d} \hat{P}(\theta) \\
&= \int_{\theta} \left( \int_{0}^{r} \mathbbm{1}_{t < g(\theta)} \, \text{d} t \right) \, \text{d} \hat{P}(\theta) \\
&= \int_{0}^{r} \left( \int_{\theta} \mathbbm{1}_{t < g(\theta)} \, \text{d} \hat{P}(\theta) \right) \, \text{d} t \quad \text{(*)} \\
&= \int_{0}^{r} \left( \int_{\{\theta : g(\theta) > t\}} \, \text{d} \hat{P}(\theta) \right) \, \text{d} t \\
&= \int_{0}^{r} \hat{P}_\succ(g, t) \, \text{d} t.
\end{aligned}
\]
Here, step (*) applies Fubini’s theorem to interchange the order of integration, and by integrating over the interval \(t \in [0, r]\), we obtain the final result.

Through the above derivation and the application of sensitivity functions, we have completed the proof of relative error approximation and established the integral transformation relation using Fubini’s theorem. Finally, we conclude that:
\[
\frac{1}{P(f)} \left| P(f) - \int_{\theta} \frac{Sf(\theta)}{s(\theta)} \, \text{d} \mu(\theta) \right| \leq \epsilon (1 + S\varsigma),
\]
which verifies that the model \(\hat{f}_\theta\) satisfies the desired inequality:
\[
\Big| \int f_\theta(x) \, \text{d} x - \int \hat{f}_\theta(x) \, \text{d} x \Big| \leq \epsilon.
\]
Thus, we have shown that \(\hat{f}_\theta\), obtained through the optimization of \(\mathcal{L}_{{se}}(\theta^i)\), satisfies the desired bound on the error in the integral.

%%%%%%%%%%%%%%%%%%%%%%%%%%%%%%%%%%%%%%%%%%%%%%%%%%%%%%%%%%%%%%%%%

%%%%%%%%%%%%%%%%%%%%%%%%%%%%%%%%%%%%%%%%%%%%%%%%%%%%%%%%%%%%%%%%%

\subsection{Proof of Theorem \ref{the:objective}}

\renewcommand{\thetheorem}{V.1}
\begin{theorem}
    Suppose $\mathcal{F}$ has gradients bounded by $\mathrm{G}$, i.e. $|| \nabla \mathcal{F}(\theta) || \leq \mathrm {G} \ \forall \ \theta$, and $\varsigma -$smooth, i.e., $|| \nabla \mathcal{F}(\theta_1) - \nabla \mathcal{F}(\theta_1) || \leq \varsigma  ||\theta_1-\theta_2|| \ \forall (\theta_1,\theta_2 )$. Let $\tilde{f}_{\theta} $ be the function evaluated after a one-step gradient update procedure, i.e.,
    \begin{equation}
        \tilde{f}_{\theta} = f_{\theta}-\beta \nabla_{f_{\theta}}\mathcal{L}(\mathcal{D}^q _i,f_{\theta^i})
    \end{equation}
    If the step size $\beta \leq \min \{ \frac{1}{2\varsigma}, \frac{\sigma}{8 \rho \mathrm {G}  } \}$ where $\sigma-$strongly convex means $ || \nabla \mathcal{F}(\theta_1) - \nabla \mathcal{F}(\theta_1) || \geq \sigma||\theta_1-\theta_2||  $, then $\tilde{f}_{\theta}$ is convex. Also, if $f_{\theta}$ and $f_{\theta^i}$ also satisfy the above $\mathrm{G}$-Lipschitz and $\varsigma -$smooth assumptions of $\mathcal{F}$, then the learning objective $\arg \min _{f_\theta}\sum_{i=1}^{N_{tr}}\mathcal{L}(\mathcal{D}^q _i,f_{\theta^i})$
    with $\beta \leq \min \lbrace \frac{1}{2\varsigma}, \frac{\sigma}{8 \rho G} \rbrace$ is $\frac{9\varsigma}{8}-$ smooth and $\frac{\sigma}{8}-$strongly convex.
\end{theorem}

\proof
    We first begin by reviewing some definitions and assumptions before proving the main theorem. Specificlly, we use $\mathcal{F}(\theta)$ as the shorthand of the learning objective of the main text, including the weight update and structure update), with the population loss $\mathcal{L}^p(f_{\theta})=\sum_{i=1}^{N_{tr}}\mathcal{L}(f_{\theta^i}) $ and empirical loss $\mathcal{L}^e(f_{\theta})=\sum_{i=1}^{N_{tr}}\mathcal{L}(\mathcal{D}_i,f_{\theta^i}) $ where $\mathcal{D}_i$ is the specific data distribution of task $\tau_i$. We make certain assumptions about every loss function within the learning problem for all tasks. Let $\theta_1$ and $\theta_2$ denote two arbitrary selections of model $f_\theta$'s parameters.
    \begin{assumption}\label{ass:theo2}
    \textbf{(Smoothness and Convexity)} Suppose that $\mathcal{F}$ is twice differentiable and:
    \begin{itemize}
        \item (Lipschitz in function value) $\mathcal{F}$ has gradients bounded by $\text{G}$, i.e., $|| \nabla \mathcal{F}(\theta) || \leq \text{G} \ \forall \ \theta$, which is equivalent to $f_{\theta}$ being $\text{G}-$Lipschitz.
        \item (Lipschitz gradient) $\mathcal{F}$ is $\beta-$smooth, i.e., $|| \nabla \mathcal{F}(\theta_1) - \nabla \mathcal{F}(\theta_1) || \leq \varsigma  ||\theta_1-\theta_2|| \ \forall (\theta_1,\theta_2 )$.
        \item (Lipschitz Hessian) $\mathcal{F}$ has $\rho-$Lipschitz Hessians, i.e. $ || \nabla^2 \mathcal{F}(\theta_1) - \nabla^2 \mathcal{F}(\theta_2) || \leq \rho ||\theta_1-\theta_2|| \ \forall (\theta_1,\theta_2) $.
        \item (Strong convexity) Suppose that $\mathcal{F}$ is convex. Furthermore, suppose $\mathcal{F}$ is $\sigma-$strongly convex, i.e., $ || \nabla \mathcal{F}(\theta_1) - \nabla \mathcal{F}(\theta_2) || \geq \sigma||\theta_1-\theta_2|| $.
    \end{itemize}
    \end{assumption}
    Consider two arbitrary points $\theta_1, \theta_2 \in \mathbb{R}^d$. Using the chain rule and our shorthand of $\tilde{\theta}_1  \equiv \mathcal{Q} (\theta_1), \tilde{\theta}_1  \equiv \mathcal{Q} (\theta_2)$. Then, we show the smoothness of the function. Taking the norm on both sides, for the specified $\beta$, we have:
    \begin{equation}
    \begin{aligned}
            &\left \|\nabla \mathcal{F} (\theta_1) - \nabla \mathcal{F} (\theta_2)\right \|  \\[8pt]
            &= \left \|\nabla \mathcal{Q}(\theta_1) \nabla \mathcal{F}(\theta_1) - \nabla \mathcal{Q}(\theta_2) \nabla \mathcal{F}(\theta_2)\right \|  \\[8pt]
            &= \|\left( \nabla \mathcal{Q}(\theta_1) - \nabla \mathcal{Q}(\theta_2) \right) \nabla \mathcal{F}(\theta_1) \\[8pt]
            &+ \nabla \mathcal{Q}(\theta_2) \left( \nabla \mathcal{F}(\theta_1) - \nabla \mathcal{F}(\theta_2) \right) \| \\[8pt]
            &\leq \| \left( \nabla \mathcal{Q}(\theta_1) - \nabla \mathcal{Q}(\theta_2) \right) \nabla \mathcal{F}(\theta_1) \| \\[8pt]
            &+ \| \nabla \mathcal{Q}(\theta_2) \left( \nabla \mathcal{F}(\theta_1) - \nabla \mathcal{F}(\theta_2) \right) \|
    \end{aligned}
    \end{equation}
    Next, we now bound both terms on the right-hand side, and consider the update procedure as $f_{\theta} \gets f_{\theta}-\beta \nabla_{f_{\theta}}\mathcal{F}_i(\theta )$ where $\mathcal{F}_i(\theta ):= f_{\theta}(\mathcal{D}^s _i,f_\theta)$ as the shorthand of the bi-level objective mentioned in the learning objective. Then, we have:
    \begin{equation}
    \begin{aligned}
        & \| \left( \nabla \mathcal{Q}(\theta_1) - \nabla \mathcal{Q}(\theta_2) \right) \nabla \mathcal{F}(\theta_1) \| \\[8pt]
        & \leq \| \nabla \mathcal{Q}(\theta_1) - \nabla \mathcal{Q}(\theta_2) \| \| \nabla \mathcal{F}(\theta_1) \| \\[8pt]
        & = \| \left( \mathbb{I}  - \beta \nabla^2 \mathcal{F}_i(\theta_1) \right) - \left( \mathbb{I}  - \beta \nabla^2 \mathcal{F}_i(\theta_2) \right) \| \| \nabla \mathcal{F}(\theta_1) \| \\[8pt]
        & = \beta \| \nabla^2 \mathcal{F}_i(\theta_1) - \nabla^2 \mathcal{F}_i(\theta_2) \| \| \nabla \mathcal{F}(\theta_1) \| \\[8pt]
        & \leq \beta \rho \| \theta_1 - \theta_2 \| \| \nabla \mathcal{F}(\theta_1) \| \\[8pt]
        & \leq \beta \rho \text{G} \| \theta_1 - \theta_2 \|
    \end{aligned}
    \end{equation}
    where the three $\leq$ equations are due to the submultiplicative property of the norm (Cauchy-Schwartz inequality), the Hessian Lipschitz property, and the Lipschitz property of the function value, respectively. Next, consider the Jacobian of $\mathcal{F}$ is given by $ \nabla\mathcal{F} (\theta_1) = \mathbb{I}  - \beta \nabla^2 \mathcal{F}_i(\theta_1) $. Since $\sigma \mathbb{I}  \preceq \nabla^2 \mathcal{F}_i(\theta_1) \preceq \varsigma \mathbb{I}  \ \forall \theta_1 \in \mathbb{R} ^d$, we can bound the Jacobian as:
    \begin{equation}
    \begin{aligned}
        (1-\beta \varsigma) \mathbb{I}  \preceq \nabla \mathcal{F}(\theta_1) \preceq (1-\beta \sigma) \mathbb{I}  \hspace*{10pt} \forall \ \theta_1 \in \mathbb{R}^d
    \end{aligned}
    \end{equation}

    The upper bound also implies that $\| \nabla \mathcal{F}(\theta_1) \| \leq (1-\beta \sigma) \ \forall \theta_1$. Then, we get that
    \begin{equation}
    \begin{aligned}
        \| \mathcal{F}(\theta_1) - \mathcal{F}(\theta_2) \| \leq (1-\beta \sigma) \| \theta_1-\theta_2 \|
    \end{aligned}
    \end{equation}
    Thus, for the second term of $\left \|\nabla \mathcal{F} (\theta_1) - \nabla \mathcal{F} (\theta_2)\right \|$, we get:
    \begin{equation}
    \begin{aligned}
    & \| \nabla \mathcal{Q}(\theta_2) \left( \nabla \mathcal{F}(\theta_1) - \nabla \mathcal{F}(\theta_2) \right) \| \\[8pt]
    & = \left\| \left( \mathbb{I}  - \beta \nabla^2 \mathcal{F}_i(\theta_2) \right) \Big( \nabla \mathcal{F}(\theta_1) - \nabla \mathcal{F}(\theta_2) \Big) \right\|
    \\[8pt]
    & \leq (1-\beta \sigma) \| \nabla \mathcal{F}(\theta_1) - \nabla \mathcal{F}(\theta_2) \| \\[8pt]
    & \leq (1-\beta \sigma) \beta \| \theta_1 - \theta_2 \| \\[8pt]
    & = (1-\beta \sigma) \beta \| \mathcal{Q}(\theta_1) - \mathcal{Q}(\theta_2) \| \\[8pt]
    & \leq (1-\beta \sigma) \beta (1-\beta \sigma) \| \theta_1 - \theta_2 \| \\[8pt]
    & = (1-\beta \sigma)^2 \beta \|\theta_1-\theta_2\|
    \end{aligned}
    \end{equation}
    Next, when $\beta \leq \min \{ \frac{1}{2\varsigma}, \frac{\sigma}{8 \rho \mathrm {G}  } \}$, putting the previous pieces together, we have:

    \begin{equation}
    \begin{aligned}
    \| \nabla \mathcal{F} (\theta_1) - \nabla \mathcal{F} (\theta_2) \| 
    & \leq \| \left( \nabla \mathcal{Q}(\theta_1) - \nabla \mathcal{Q}(\theta_2) \right) \nabla \mathcal{F}(\theta_1) \| \\[8pt]
    &+ \| \nabla \mathcal{Q}(\theta_2) \left( \nabla \mathcal{F}(\theta_1) - \nabla \mathcal{F}(\theta_2) \right) \| \\[8pt]
    & \scalebox{0.95}{$\leq \beta \rho \text{G} \| \theta_1 - \theta_2 \| + (1-\beta \sigma)^2 \beta \| \theta_1 - \theta_2 \|$} \\[8pt]
    & \leq ( \frac{\sigma}{8} + \varsigma  ) \| \theta_1 - \theta_2 \| \\[8pt]
    & \leq \frac{9 \varsigma}{8} \| \theta_1-\theta_2 \|
    \end{aligned}
    \end{equation}
    thus, $\mathcal{F}$ is $\frac{9\varsigma}{8}$ smooth. Similarly, for the lower bound, we have: 
    \begin{equation}
    \begin{aligned}
    \| \nabla \mathcal{F} (\theta_1) - \nabla \mathcal{F} (\theta_2) \| & = \| \left( \nabla \mathcal{Q}(\theta_1) - \nabla \mathcal{Q}(\theta_2) \right) \nabla \mathcal{F}(\theta_1) \\[8pt]
    &+ \nabla \mathcal{Q}(\theta_2) \left( \nabla \mathcal{F}(\theta_1) - \nabla \mathcal{F}(\theta_2) \right) \| \\[8pt]
    & \geq  \| \nabla \mathcal{Q}(\theta_2) \left( \nabla \mathcal{F}(\theta_1) - \nabla \mathcal{F}(\theta_2) \right) \| \\[8pt]
    &- \| \left( \nabla \mathcal{Q}(\theta_1) - \nabla \mathcal{Q}(\theta_2) \right) \nabla \mathcal{F}(\theta_1) \|
    \end{aligned}
    \end{equation}
    Then, for the first term, consider $\lambda_{\min} ( \mathbb{I}  - \beta \nabla^2 \mathcal{F}_i(\theta_2) ) \geq 1-\beta \varsigma$, we have:
    \begin{equation}
    \begin{aligned}
    & \| \nabla \mathcal{Q}(\theta_2) \left( \nabla \mathcal{F}(\theta_1) - \nabla \mathcal{F}(\theta_2) \right) \|\\[8pt]
    & = \left\| \left( \mathbb{I}  - \beta \nabla^2 \mathcal{F}_i(\theta_2) \right) \Big( \nabla \mathcal{F}(\theta_1) - \nabla \mathcal{F}(\theta_2) \Big) \right\| \\[8pt]
    &- \| \left( \nabla \mathcal{Q}(\theta_1) - \nabla \mathcal{Q}(\theta_2) \right) \nabla \mathcal{F}(\theta_1) \|\\[8pt]
     & \geq (1-\beta\beta) \| \nabla \mathcal{F}(\theta_1) - \nabla \mathcal{F}(\theta_2) \| \\[8pt]
     &- \| \left( \nabla \mathcal{Q}(\theta_1) - \nabla \mathcal{Q}(\theta_2) \right) \nabla \mathcal{F}(\theta_1) \|
    \end{aligned}
    \end{equation}
    Next, consider $\mathcal{F}_i$ which achieves strong convexity and smoothness, the equation turns to:
    \begin{equation}
    \begin{aligned}
     &\| \nabla \mathcal{Q}(\theta_2) \left( \nabla \mathcal{F}(\theta_1) - \nabla \mathcal{F}(\theta_2) \right) \|\\[8pt]
     & \geq (1-\beta \beta) \sigma \| \theta_1 - \theta_2 \| - \| \left( \nabla \mathcal{Q}(\theta_1) - \nabla \mathcal{Q}(\theta_2) \right) \nabla \mathcal{F}(\theta_1) \|\\[8pt]
     & = (1-\beta \beta) \sigma \| \theta_1 - \beta \nabla \mathcal{F}_i(\theta_1) - \theta_2 + \beta \nabla \mathcal{F}_i(\theta_2) \| \\[8pt]
     & \geq \sigma (1-\beta \beta) \Big( \| \theta_1-\theta_2 \| - \beta \| \nabla \mathcal{F}_i(\theta_1) - \nabla \mathcal{F}_i(\theta_2) \| \Big) \\[8pt]
     & \geq \sigma (1-\beta \beta) \Big( \| \theta_1-\theta_2 \| - \beta \beta \| \theta_1-\theta_2 \| \Big) \\[8pt]
     & \geq \sigma (1-\beta \beta)^2 \| \theta_1 - \theta_2 \| \\[8pt]
     & \geq \frac{\sigma}{4} \| \theta_1-\theta_2 \|
    \end{aligned}
    \end{equation}
    From all the above terms, we have:
    \begin{equation}
    \begin{aligned}
     \| \nabla \mathcal{F} (\theta_1) - \nabla \mathcal{F} (\theta_2) \|& \geq  \| \nabla \mathcal{Q}(\theta_2) \left( \nabla \mathcal{F}(\theta_1) - \nabla \mathcal{F}(\theta_2) \right) \| \\[8pt]
     &- \| \left( \nabla \mathcal{Q}(\theta_1) - \nabla \mathcal{Q}(\theta_2) \right) \nabla \mathcal{F}(\theta_1) \| \\[8pt]
     & \geq \left( \frac{\sigma}{4} - \frac{\sigma}{8} \right) \|\theta_1 - \theta_2 \| \geq \frac{\sigma}{8} \|\theta_1-\theta_2 \|
    \end{aligned}
    \end{equation}  
    Thus the function $\mathcal{F} (\cdot)$ is $\tilde{\sigma}=\frac{\sigma}{8}$ strongly convex.

%%%%%%%%%%%%%%%%%%%%%%%%%%%%%%%%%%%%%%%%%%%%%%%%%%%%%%%%%%%%%%%%%

%%%%%%%%%%%%%%%%%%%%%%%%%%%%%%%%%%%%%%%%%%%%%%%%%%%%%%%%%%%%%%%%%

\subsection{Proof of Corollary \ref{cor:bound}}

\renewcommand{\thecorollary}{V.1}
\begin{corollary}
    Suppose that for all iterations, $f_{\theta}$ and $f_{\theta^i}$ satisfy the $\mathrm{G}$-Lipschitz and $\varsigma -$smooth assumptions, and the update procedure in NeuronML is performed with $\beta \leq \min \{ \frac{1}{2\varsigma}, \frac{\sigma}{8 \rho \mathrm {G}  } \}$. Then, the model $f_{\theta}$ enjoys the guarantee $\sum_{t=1}^{T}\mathcal{F}(\theta)= \min_{\theta }\sum_{t=1}^{T}\mathcal{F}(\theta)+O(\frac{32\mathrm{G}^2}{\sigma}\log T )$.
\end{corollary}

\proof
From \textbf{Theorem~\ref{the:objective}}, we conclude that each function $\mathcal{F}$ is $\frac{\sigma}{8}-$strongly convex. Thus, when applied to the sequence of loss functions $\lbrace \mathcal{F} _i \rbrace_{i=1}^{N_{tr}}$, and it guarantees a regret of $O(\frac{32\mathrm{G}^2}{\sigma}\log N_{tr} )$, following \cite{cesa2006prediction,mohri2018foundations}.

%%%%%%%%%%%%%%%%%%%%%%%%%%%%%%%%%%%%%%%%%%%%%%%%%%%%%%%%%%%%%%%%%

%%%%%%%%%%%%%%%%%%%%%%%%%%%%%%%%%%%%%%%%%%%%%%%%%%%%%%%%%%%%%%%%%

\subsection{Proof of Theorem \ref{the:generalization}}

\renewcommand{\thetheorem}{V.2}
\begin{theorem}
    If $f_{\theta}$ is a possibly randomized symmetric and the assumption of \textbf{Corollary \ref{cor:bound}} hold, then
    $\mathbb{E}_{f_{\theta },\mathcal{D} }\left [ \mathcal{L}^p(f_{\theta})- \mathcal{L}^e(f_{\theta}) \right ]\le \xi $, where $\xi := O(1) \frac{G^2 (1 + \beta \varsigma K)}{N_{tr} N \sigma}$ is the bound of generalization error. It decays by a factor of $O(K/N_{tr} N)$, where $N_{tr}$ is the number of training tasks, $N$ is the number of available samples per task, and $K$ is the number of data points used in actual training batch.
\end{theorem}

\proof Before proving it, we first restate the assumptions on which the theorem is based:
    \begin{assumption}\label{ass:theo3}
        We assume $\mathcal{Z}=(\mathcal{X}, \mathcal{Y})$ is a Polish space (i.e., complete, separable, and metric) and $\text{F}_\mathcal{Z}$ is the Borel algebra over $\mathcal{Z}$. Moreover, for any $i$, $p_i$ is a non-atomic probability distribution over $(\mathcal{Z}, \text{F}_\mathcal{Z})$, i.e., $p_i(z)=0$ for every $z \in \mathcal{Z}$.	
    \end{assumption}  
    To show the claim, it just suffices to show	that for any $i$, we have:
    \begin{equation}
    \begin{aligned}
        \mathbb{E}_{f_{\theta },\mathcal{D} }\left [ \mathcal{L}^p_i(f_{\theta})- \mathcal{L}^e_i(f_{\theta}) \right ]\le \xi \\[8pt]
    \end{aligned} 
    \end{equation}
    where $\mathcal{L}^p_i(f_{\theta})$ and $\mathcal{L}^e_i(f_{\theta})$ are the population loss and the empirical loss respectively. Consider $\mathcal{D}_i^{in} = \{z^{in}_1,...,z^{in}_n\}$, $\mathcal{D}_i^{out}=\{z^{out}_1,...,z^{out}_n\}$. To see this, first note that:
    \begin{equation}
    \begin{aligned}
    \mathcal{L}^p_i(f_{\theta})
    = \mathbb{E}_{\{z_{i,j}\}_{j=1}^K , \tilde{z}} \left[ \ell \left (f_{\theta}(\{z_{i,j}\}_{j=1}^K,f_{\theta^i}), \tilde{z} \right ) \right].
    \end{aligned} 
    \end{equation}
    Here, $\{z_{i,j}\}_{j=1}^K$ represents a set of $K$ distinct points randomly drawn from $p_i$, and $\tilde{z}$ is independently sampled from the same distribution $p_i$. By \textbf{Assumption \ref{ass:theo2}}, we could assume $\tilde{z}$ is different from $K$ other points. Note that we have:
    \begin{equation}
    \begin{aligned}
    \mathbb{E}_{\mathcal{D}_i}[\mathcal{L}^p_i(f_{\theta})]	
    = \mathbb{E}_{\mathcal{D}, \{\{z_{i,j}\}_{j=1}^K , \tilde{z}\}} \left[ \ell \left (f_{\theta}(\{z_{i,j}\}_{j=1}^K,f_{\theta^i}), \tilde{z} \right ) \right].
    \end{aligned} 
    \end{equation}
     Next, note that, for $\mathcal{L}^p_i(f_{\theta})$, we have:
    \begin{equation}
    \begin{aligned}
    &\mathbb{E}_{f_{\theta },\mathcal{D}}[\mathcal{L}^p_i(f_{\theta})] \\[8pt]
    &= 
    \scalebox{0.9}{$\frac{1}{\binom{N}{K} |\mathcal{D}_i^{out}|} \sum_{\substack{\{z^{sym}_{i,j}\}_{j=1}^K \subset [N] \\ \tilde{z}^{sym} \in [N] }} 
    \mathbb{E}_{f_{\theta }, \mathcal{D}} \left[ \ell \left (f_{\theta}(\{z^{sym}_{i,j}\}_{j=1}^K,f_{\theta^i}), \tilde{z}^{sym} \right ) \right]$}
    \end{aligned} 
    \end{equation}
    where $\{z^{sym}_{i,j}\}_{j=1}^K$ are all different, and hence, due to the symmetry. Next, for fixed $\{z^{sym}_{i,j}\}_{j=1}^K \subset [N]$ and $\tilde{z}^{sym} \in [N]$, and define the dataset $\tilde{\mathcal{D}}$ by substituting $z^{in}_{sym}$ with $z_j$, for all $j$, and $z^{out}_{sym}$ with $\tilde{z}$, we can obtain that:
    \begin{equation}
    \begin{aligned}
    \mathbb{E}_{f_{\theta}, \mathcal{D}}[\mathcal{L}^e_i(f_{\theta})] &= 
    \mathbb{E}_{f_{\theta}, \mathcal{D}}\left [\ell \left (f_{\theta}(\{z^{in}_{sym}\}_{j=1}^K,f_{\theta^i}), \tilde{z}^{out}_{sym} \right) \right ] \\[8pt]
    &= \scalebox{0.85}{$\mathbb{E}_{f_{\theta}, \mathcal{D}, \{\{z_{i,j}\}_{j=1}^K , \tilde{z}\}} \left [\ell \left (f_{\theta}(\{z^{in}_{sym}\}_{j=1}^K,f_{\theta^i}), \tilde{z}^{out}_{sym} \right) \right ]$} \\[8pt]
    &= \scalebox{0.9}{$\mathbb{E}_{f_{\theta}, \mathcal{D}, \{\{z_{i,j}\}_{j=1}^K , \tilde{z}\}}\left [\ell \left (f_{\theta}(\{z_{i,j}\}_{j=1}^K,f_{\theta^i}), \tilde{z}_{i,j} \right) \right ]$}
    \end{aligned} 
    \end{equation}
    Thus, by putting the above equations together with Tonelli' theorem, we can obtain:
    \begin{equation}
    \begin{aligned}
    & \mathbb{E}_{f_{\theta}, \mathcal{D},}[\mathcal{L}^p_i(f_{\theta})- \mathcal{L}^e_i(f_{\theta})] \\[8pt]
    & \le  \scalebox{0.75}{$\mathbb{E}_{f_{\theta}, \mathcal{D},\{\{z_{i,j}\}_{j=1}^K , \tilde{z}\}}\left [\left | \ell \left (f_{\theta}(\{z_{i,j}\}_{j=1}^K,f_{\theta^i}), \tilde{z} \right)-\ell \left (\tilde{f} _{\theta}(\{z_{i,j}\}_{j=1}^K,f_{\theta^i}), \tilde{z} \right)   \right | \right ] $} \\[8pt]
    &= \scalebox{0.73}{$\mathbb{E}_{\mathcal{D},\{\{z_{i,j}\}_{j=1}^K , \tilde{z}\}}\left [\mathbb{E}_{f_{\theta}}\left [ \left | \ell \left (f_{\theta}(\{z_{i,j}\}_{j=1}^K,f_{\theta^i}), \tilde{z} \right)-\ell \left (\tilde{f} _{\theta}(\{z_{i,j}\}_{j=1}^K,f_{\theta^i}), \tilde{z} \right)   \right | \right ]\right ] $}
    \end{aligned} 
    \end{equation}
    Finally, note that since $f_{\theta}$ is $(\xi, K)$-uniformly stable, we could bound the inner integral by $\xi$, i.e., 
    \begin{equation}
    \begin{aligned}
    \mathbb{E}_{f_{\theta },\mathcal{D} }\left [ \mathcal{L}^p(f_{\theta})- \mathcal{L}^e(f_{\theta}) \right ]\le \xi 
    \end{aligned} 
    \end{equation}
    Next, recall the definition of $N_{tr}$ and $N$, where $N_{tr}$ is the number of training tasks, and $N$ is the number of available samples per task. Consider task $\tau_i$ is in the training batch $\mathcal{B}_t$ with probability $p_{i,t}/N_{tr}$, and if that happens, then $u_t$ would have a binomial distribution with mean $b/N$. Under \textbf{Assumption \ref{ass:theo2}} and \textbf{Assumption \ref{ass:theo3}}, we have:
    \begin{equation}
    \begin{aligned}
    \scalebox{0.98}{$\mathbb{E}_{f_{\theta },\mathcal{D} }\left [ \mathcal{L}^p(f_{\theta})- \mathcal{L}^e(f_{\theta}) \right ]
    \leq 	\frac{4\text{G}}{N_{tr}N} (1 + \beta \varsigma K) \frac{16(2\varsigma+\rho \beta \text{G}) + \sigma}{\sigma (2\varsigma+\rho \beta \text{G}}$}
    \end{aligned} 
    \end{equation}
    Before showing its proof, note that since $\varsigma \geq \sigma$, this could be simplified as
    \begin{equation}
    \begin{aligned}
    \mathbb{E}_{f_{\theta },\mathcal{D} }\left [ \mathcal{L}^p(f_{\theta})- \mathcal{L}^e(f_{\theta}) \right ]\le O(1) \frac{G^2 (1 + \beta \varsigma K)}{N_{tr} N \sigma}
    \end{aligned} 
    \end{equation}
    Thus, $\xi := O(1) \frac{G^2 (1 + \beta \varsigma K)}{N_{tr} N \sigma}$, which is the bound of generalization error.

\section{Species of Neuromodulated Meta-Learning}
\label{sec:5}
In this section, we explore specific implementations of NeuronML for supervised, self-supervised, and reinforcement learning. While these domains differ in loss function formulation and data generation methods, the core adaptation mechanism remains consistent across all scenarios.

\subsection{Supervised Learning}

Few-shot learning has been extensively explored within the realm of supervised tasks, where the objective is to acquire knowledge based on limited data for that specific task \cite{protonet,Meta-sgd}. This is achieved by constructing diverse tasks from limited data to acquire task-related knowledge. For instance, the few-shot classification task could involve classifying images of a British Shorthair with the aid of a model that has been exposed to just one or a few instances of this particular breed, having previously encountered numerous other object types. Similarly, in the context of few-shot regression, the goal is to make predictions for the outputs of a function with continuous values using only a small set of data points obtained from that function \cite{wang2023hacking,relationnet}. This occurs after the model has undergone training on a multitude of functions sharing similar statistical characteristics. Note that the implementation of NeuronML in this scenario is similar to Algorithm \ref{alg:example}, but discuss classification and regression tasks separately in few-shot settings.

In the context of meta-learning, the supervised regression and classification problems can be characterized by setting the training horizon to 1, which means the model is designed to handle a single input and generate a corresponding single output, rather than dealing with sequences of inputs and outputs. For a given task $\tau_i$, the task generates $N_i^s+N_i^q$ independent and identically distributed (i.i.d.) observations $x$ from the dataset $\mathcal{D}_i $ of task $\tau_i$. The task loss is quantified by the discrepancy between the model's output for $x$ and the corresponding target values $y$ associated with that observation and task. Two widely used loss functions for supervised classification and regression are cross-entropy and mean-squared error (MSE), which will be elaborated upon below. It's worth noting that other supervised loss functions may also be employed. In the case of regression tasks using mean-squared error, the loss is expressed as follows:
% %\vspace{-0.15cm}
\begin{equation}\label{eq:sl_r}
    \begin{array}{l}
         \mathcal{L}(\mathcal{D}_i,\theta^i )=\sum_{(x_{i,j}, y_{i,j}) \sim \tau_i}  \lVert f_{\theta^i}(x_{i,j}) - y_{i,j}  \rVert_2^2,
    \end{array}
\end{equation}
where $(x_{i,j}, y_{i,j})$ are an input/output pair sampled from task $\tau_i$. In $K$-shot regression tasks, $K$ input/output pairs are provided for learning for each task.

Similarly, for discrete classification tasks with a cross-entropy loss, the loss takes the form:
\begin{equation}\label{eq:sl_c}
    \begin{aligned}
         \mathcal{L}(\mathcal{D}_i,\theta^i )=&\sum_{(x_{i,j}, y_{i,j}) \sim \tau_i}  y_{i,j}\log f_{\theta^i}(x_{i,j})\\
         &+  (1-y_{i,j})\log (1-f_{\theta^i}(x_{i,j})),
    \end{aligned}
\end{equation}
Following traditional terminology, $K$-shot classification tasks involve employing $K$ input/output pairs from each class, resulting in a total of $N_i^s$ data points for $N_i^s/K$-way classification. By utilizing a task distribution $\mathbb{P} (\mathcal{T})$, these loss functions can be seamlessly incorporated into the equations outlined in \textbf{Section \ref{sec:4}} for learning via NeuronML, as elucidated in \textbf{Algorithm \ref{alg:SL}}.

\begin{algorithm}[t]
   \caption{NeuronML for Supervised Learning}
   \label{alg:SL}
\begin{flushleft}
\textbf{Input:} Task distribution $\mathbb{P} (\mathcal{T})$, NeuronML model $f_{\theta_{\mathcal{M}}}$ with randomly initialized parameters $\theta_\mathcal{M}=\mathcal{M}\odot\theta$\\
\textbf{Output:} The trained NeuronML model $f_{\theta}$\\
\end{flushleft}
\begin{algorithmic}[1]
   \WHILE{not done}
   \STATE Sample $N_{tr}$ tasks from $\mathbb{P} (\mathcal{T})$, $\left \{ \tau_i \right \}_{i=1}^{N_{tr}} \sim \mathbb{P} (\mathcal{T})$
   \FOR{$i=1$ {\bfseries to} $N_{tr}$}
   \STATE Divide the data of $\tau_i$ into $\mathcal{D}_i^s$ and $\mathcal{D}_i^q$
   \STATE Compute the loss $\mathcal{L}_{weight}(\mathcal{D}^s _i,\theta_\mathcal{M})$ via Eq.\ref{eq:sl_r} and Eq.\ref{eq:sl_c}
   \STATE Update $\theta_\mathcal{M}^i$ using $\mathcal{L}_{weight}(\mathcal{D}^s _i,\theta_\mathcal{M})$
   \STATE Compute $\mathcal{L}_{weight}(\mathcal{D}^q _i,\theta_\mathcal{M}^i)$ using query set $\mathcal{D}^q_i$
   \STATE Compute $\mathcal{L}_{structure}(\mathcal{D}_i,\theta_\mathcal{M}^i)$ using all the data $\mathcal{D}_i$
   \ENDFOR
   \STATE Update $\theta$ of $\theta_\mathcal{M}$ using $\mathcal{L}_{weight}(\mathcal{D}^q _i,\theta_\mathcal{M}^i)$ of all tasks
   \STATE Update $\mathcal{M}$ of $\theta_\mathcal{M}$ using $\mathcal{L}_{structure}(\mathcal{D}_i,\theta_\mathcal{M}^i)$ of all tasks
   \ENDWHILE
\end{algorithmic}
\end{algorithm}

\subsection{Self-supervised Learning}
Self-supervised learning is a machine learning paradigm characterized by the use of self-generated labels within the dataset during the training process \cite{wang2024gessl}, eliminating the need for external annotated target labels. In self-supervised learning, models learn representations by transforming tasks into their own supervisory signals, bypassing the necessity for manually labeled targets. The central idea of this paradigm \cite{byol,moco} is to automatically create targets from input data, thereby driving the model to learn meaningful feature representations, effectively solving the problem of lack of supervisory information faced by the model. Our NeuronML, as a general global learning mechanism, can solve self-supervised learning problems well. For better application, we also propose a simple meta-learning-based method to reconstruct self-supervised tasks.

We begin by reviewing the learning paradigms of both frameworks and highlighting their similarities following \cite{wang2023unleash,ni2021close}. 
For self-supervised learning: i) sample $ \{ x_i  \}_{i=1}^{N}$ from the distribution of the raw data space $\mathcal{X} $; ii) apply multiple data augmentations to $ \{ x_i  \}_{i=1}^{N}$, e.g., random scaling, rotation, and cropping, obtaining $\mathcal{A} = \{ a^j(x_i)  \}_{i=1,j=1}^{i=N,j=M}$; iii) learn a general visual representation based on minimizing self-supervised learning loss, e.g., contrastive loss \cite{simclr}; iv) use pre-trained models to extract features and verify them in downstream tasks. For meta-learning: i) sample $\{ x_i\}_{i=1}^{N}$ from the distribution of the raw data space $X$; ii) construct tasks $ \tau_{i}^{K}$ by partitioning $\{x_i\}_{i=1}^{N}$; iii) find the best initial parameters based on meta loss (outer loop): minimizes the cumulative gradient loss for all of the task-specific model (inner loop); iv) use the trained model for fast adaptation to new tasks. Conceptually, we find that the training process of self-supervised learning is similar to meta-learning and includes several similarities:
\begin{itemize}
    \item Both aim to learn generalizable representations and quickly adapt to new tasks: self-supervised learning aims to discriminate between unseen images; meta-learning aims to discriminate between unseen tasks. 
    \item Both learn a fixed amount of information that can be transferred to new tasks: self-supervised learning leverages the similarity and the dissimilarity among multiple views of the samples; meta-learning leverages the similarity of instances within each task. 
    \item Both use batches as units of data processing: self-supervised learning treats all views augmented from the same sample as a batch; meta-learning treats each task which consists of $K$ $n$-way-$m$-shot tasks as a batch. 
\end{itemize}

Inspired by this, we introduce a general paradigm to unify self-supervised learning and meta-learning. The idea is to transform self-supervised learning into a task distribution that is suitable for meta-learning optimization and learn from it. The paradigm consists of the following steps: i) randomly sample a batch of $N$ input images $ \left \{ x_{i} \right \} _{i=1}^{N} \in \mathcal{X} $ from the candidate pool; ii) divide $\mathcal{X}$ into $K$ blocks, each block contains $N/K$ images; iii) apply multiple data augmentations $a \in \mathcal{A}$ on $x \in \mathcal{X}$ of each blocks, obtaining $\{ a^j(x_i)  \}_{i=1,j=1}^{i=N,j=M}$; iv) create an $N$-way classification problem for each block: all data generated from the same $x_{i}$ is regarded as a category, with $a(x)\in\hat{\mathcal{X}} $ as the data and $y\in \mathcal{Y}$ as the pseudo-labels; v) integrate tasks corresponding to $K$ blocks in a batch, obtaining the task distribution $\mathbb{P} (\mathcal{T})$. 
In summary, for each training batch, we can get $K$ $N$-way $N\times M/K$-shot tasks. Similarly, for discrete these classification tasks which constructed based on pseudo-labels, we use a cross-entropy loss for evaluation:
\begin{equation}\label{eq:ssl}
    \begin{aligned}
         \mathcal{L}(\mathcal{D}_i,\theta^i )
         =&\sum_{(a^j(x_{i}), \hat{y} _{i,j}) \sim \tau_i}  \hat{y} _{i,j}\log f_{\theta^i}(a^j(x_{i}))\\
         &+  (1-\hat{y} _{i,j})\log (1-f_{\theta^i}(a^j(x_{i}))),
    \end{aligned}
\end{equation}
The pseudo-code is elucidated in \textbf{Algorithm \ref{alg:SSL}}.

\begin{algorithm}[t]
   \caption{NeuronML for Self-Supervised Learning}
   \label{alg:SSL}
\begin{flushleft}
\textbf{Input:} Candidate pool $\mathcal{D} $, NeuronML model $f_{\theta}$, NeuronML model $f_{\theta_{\mathcal{M}}}$ with randomly initialized $\theta_\mathcal{M}=\mathcal{M}\odot\theta$\\
\textbf{Output:} The trained NeuronML model $f_{\theta}$\\
\end{flushleft}
\begin{algorithmic}[1]
   \WHILE{not done}
   \FOR{task construction}
    \STATE Sample a mini-batch $\mathcal{B}_{\boldsymbol{x}}$ with $N$ samples from $\mathcal{D}$
    \STATE Apply random data augmentations to $\mathcal{B}_{\boldsymbol{x}}$, obtaining $\{ a^j(x_i)  \}_{i=1,j=1}^{i=N,j=M}$
    \STATE Obtain $\mathcal{D}_{\boldsymbol{x}} =(\mathcal{X}, \mathcal{Y})$ where $\mathcal{X}=\{ a^j(x_i) \}_{i=1,j=1}^{i=B,j=2}$ are the samples and $\mathcal{Y}=\{ y_i \}_{i=1}^{ B}$ are the corresponding labels
    \STATE Divide the $\mathcal{D}_{\boldsymbol{x}}$ into $N_{tr}$ blocks
    \STATE Obtain $N_{tr}$ $N$-way classification tasks $\mathcal{T}_{\boldsymbol{x}}$ of each training batch $\mathcal{B}_{\boldsymbol{x}}$
   \ENDFOR
   \FOR{$i=1$ {\bfseries to} $N_{tr}$}
   \STATE Divide the data of $\tau_i$ into $\mathcal{D}_i^s$ and $\mathcal{D}_i^q$
   \STATE Compute the loss $\mathcal{L}_{weight}(\mathcal{D}^s _i,\theta_\mathcal{M})$ via Eq.\ref{eq:ssl}
   \STATE Update $\theta_\mathcal{M}^i$ using $\mathcal{L}_{weight}(\mathcal{D}^s _i,\theta_\mathcal{M})$
   \STATE Compute $\mathcal{L}_{weight}(\mathcal{D}^q _i,\theta_\mathcal{M}^i)$ via Eq.\ref{eq:ssl}
   \STATE Compute $\mathcal{L}_{structure}(\mathcal{D}_i,\theta_\mathcal{M}^i)$ using all the data $\mathcal{D}_i$
   \ENDFOR
   \STATE Update $\theta$ of $\theta_\mathcal{M}$ using $\mathcal{L}_{weight}(\mathcal{D}^q _i,\theta_\mathcal{M}^i)$ of all tasks
   \STATE Update $\mathcal{M}$ of $\theta_\mathcal{M}$ using $\mathcal{L}_{structure}(\mathcal{D}_i,\theta_\mathcal{M}^i)$ of all tasks
   \ENDWHILE
\end{algorithmic}
\end{algorithm}

\subsection{Reinforcement Learning}
In the realm of reinforcement learning (RL), meta-learning aims to empower an agent to rapidly develop a policy for a novel test task with minimal exposure to the test environment \cite{duan2016rl,liu2023simple}. This involves the agent efficiently adapting to a new task, which may entail accomplishing a fresh goal or achieving success on a goal previously encountered but in a different environment \cite{gupta2018meta,beck2023hypernetworks}. For example, an agent learns to quickly grasp objects on a table by interacting with the environment, so that when faced with a new object, it can determine how to reliably achieve the goal of grasping with only a few samples. In this subsection, we discuss how NeuronML be applied in RL, and guide the agent for fast learning and adaptation.

In a more precise formulation, the meta-reinforcement learning setting involves a task distribution denoted as $\mathbb{P} (\mathcal{T})$, where each task $\tau_i$ is represented as a Markov decision process $\langle \mathcal{S}, \mathcal{A}, \mathcal{R}_{\tau_i}, \mathcal{T}_{\tau_i}\rangle$ comprising states $\mathcal{S}$, actions $\mathcal{A}$, a reward function $\mathcal{R}_{\tau_i}$, and plasticity $\mathcal{T}_{\tau_i}$. To characterize the process of learning from a limited number of instances on each task, we formalize it as a ``trial'' following \cite{duan2016rl,liu2023simple}. A trial involves the sampling of a new task from $\mathbb{P} (\mathcal{T})$, executing a policy $\pi^\text{exp}$ for a few ``exploration episodes'', resulting in a trajectory $\tau_i^\text{exp} = (s_0, a_0, r_0, \ldots)$, and running a policy $\pi^\text{task}$ on a concluding ``evaluation episode'' conditioned on the exploration trajectory $\tau_i^\text{exp}$. It's noteworthy that $\pi^\text{exp}$ and $\pi^\text{task}$ can be the same policy. The primary objective in the meta-reinforcement learning setting is to learn policies $\pi^\text{exp}$ and $\pi^\text{task}$ that maximize the expected returns achieved by $\pi^\text{task}$ over trials. The loss can be expressed as:
\begin{equation}\label{eq:rl}
    \begin{array}{l}
         \mathcal{L}(\mathcal{D}_i,\theta^i )=- \mathbb{E}_{\tau_i \sim \mathbb{P} (\mathcal{T}), \tau_i^\text{exp} \sim \pi^\text{exp}(\tau_i)} \left[V_\pi^\text{task} \left(\tau^\text{exp} \right)\right],
    \end{array}
\end{equation}
where $V^\text{task}_\pi(\tau_i^\text{exp})$ denotes the expected returns of $\pi^\text{task}$ conditioned on $\tau_i^\text{exp}$ for a single episode, and $\pi^\text{exp}(\tau_i)$ denotes the distribution over trajectories resulting from $\pi^\text{exp}$ on $\tau_i$.

According to \cite{gupta2018meta,beck2023hypernetworks}, as the expected reward typically lacks differentiability owing to unknown dynamics, we employ policy gradient methods to estimate the gradient for both the model gradient update(s) and meta-optimization following \cite{duan2016rl,maml}. Given that policy gradients are an on-policy algorithm, each additional gradient step during the adaptation of the NeuronML model $f_\theta$ necessitates new samples from the current policy $f_{\theta^i}$. The pseudo-code of this algorithm is provided in \textbf{Algorithm~\ref{alg:RL}}. This algorithm follows the same structure as supervised learning and self-supervised learning, with the primary distinction being that the steps involve sampling trajectories from the environment corresponding to task $\tau_i$. 

\begin{algorithm}[t]
   \caption{NeuronML for Reinforcement Learning}
   \label{alg:RL}
\begin{flushleft}
\textbf{Input:} Task distribution $\mathbb{P} (\mathcal{T})$, NeuronML model $f_{\theta_{\mathcal{M}}}$ with randomly initialized parameters $\theta_\mathcal{M}=\mathcal{M}\odot\theta$\\
\textbf{Output:} The trained NeuronML model $f_{\theta}$\\
\end{flushleft}
\begin{algorithmic}[1]
   \WHILE{not done}
   \STATE Sample $N_{tr}$ tasks from $\mathbb{P} (\mathcal{T})$, $\left \{ \tau_i \right \}_{i=1}^{N_{tr}} \sim \mathbb{P} (\mathcal{T})$
   \FOR{$i=1$ {\bfseries to} $N_{tr}$}
   \STATE Obtain $K_{tr}$ trajectories $\mathcal{D}_i^s=\left \{ \tau_i^j \right \}_{j=1}^{K_{tr}}$ using $f_{\theta_{\mathcal{M}}}$
    \STATE Compute the loss $\mathcal{L}_{weight}(\mathcal{D}^s _i,\theta_\mathcal{M})$ via Eq.\ref{eq:rl}
   \STATE Update $\theta_\mathcal{M}^i$ using $\mathcal{L}_{weight}(\mathcal{D}^s _i,\theta_\mathcal{M})$
   \STATE Obtain $K_{te}$ trajectories $\mathcal{D}_i^q=\left \{ \tau_i^j \right \}_{j=1}^{K_{te}}$ using $f_{\theta_{\mathcal{M}}^i}$
   \STATE Compute $\mathcal{L}_{weight}(\mathcal{D}^q _i,\theta_\mathcal{M}^i)$ via Eq.\ref{eq:rl}
   \STATE Compute $\mathcal{L}_{structure}(\mathcal{D}_i,\theta_\mathcal{M}^i)$ using all the data $\mathcal{D}_i$
   \ENDFOR
   \STATE Update $\theta$ of $\theta_\mathcal{M}$ using $\mathcal{L}_{weight}(\mathcal{D}^q _i,\theta_\mathcal{M}^i)$ of all tasks
   \STATE Update $\mathcal{M}$ of $\theta_\mathcal{M}$ using $\mathcal{L}_{structure}(\mathcal{D}_i,\theta_\mathcal{M}^i)$ of all tasks
   \ENDWHILE
\end{algorithmic}
\end{algorithm}

\begin{figure*}
\centering
    \subfigure[Step=3, Data points=3]{
        \includegraphics[width=0.25\linewidth]{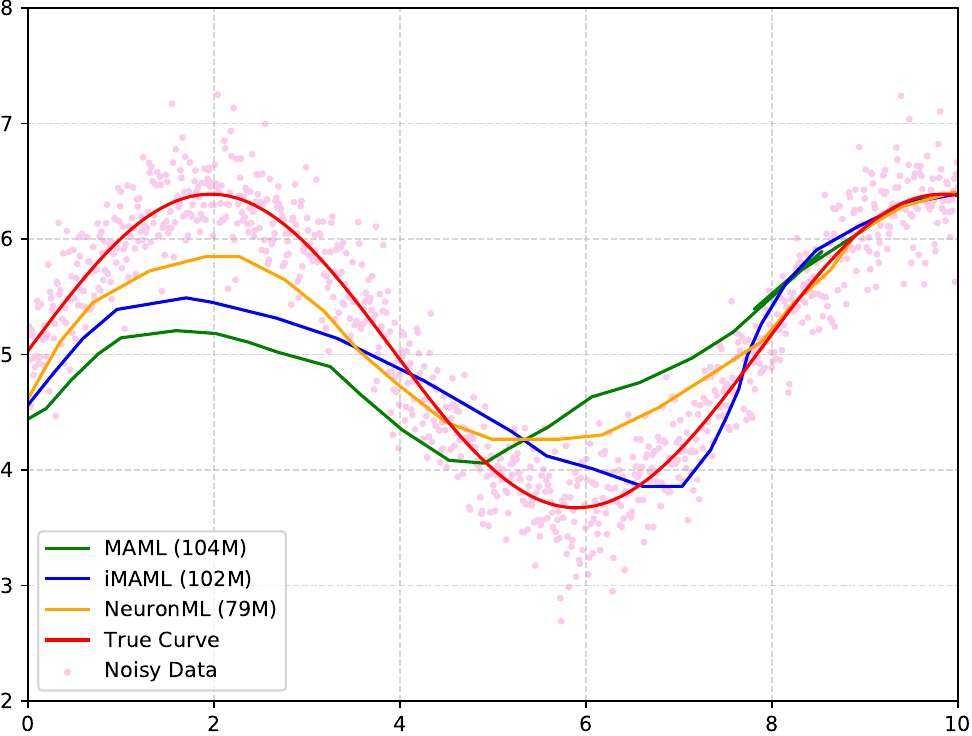}}
        \hfill
    \subfigure[Step=3, Data points=5]{
        \includegraphics[width=0.25\linewidth]{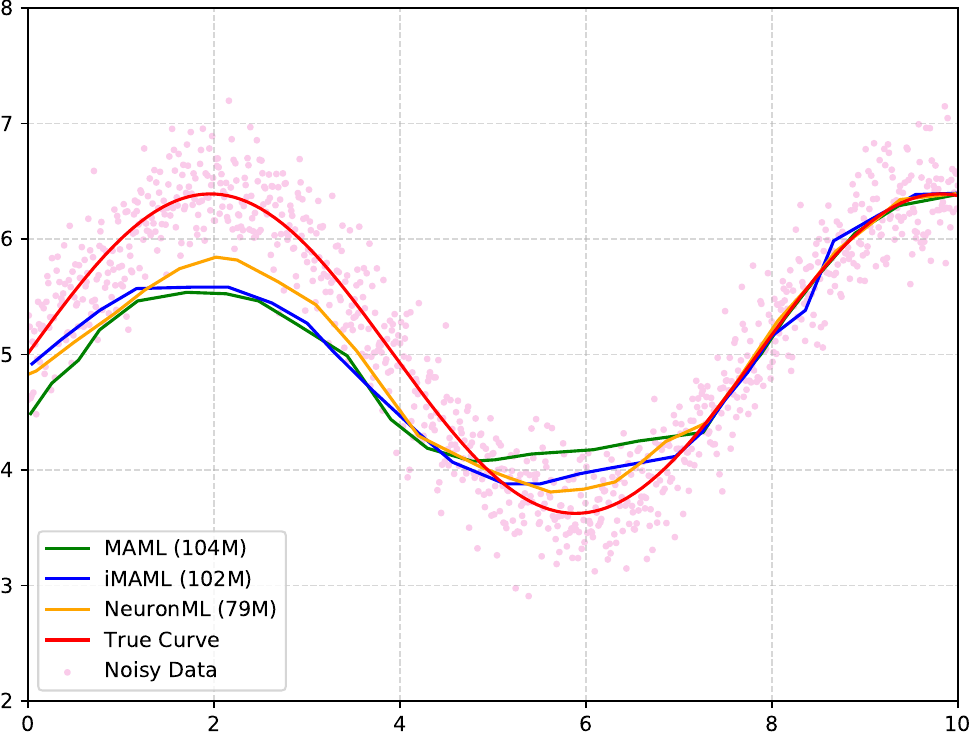}}
        \hfill
    \subfigure[Step=3, Data points=8]{
        \includegraphics[width=0.25\linewidth]{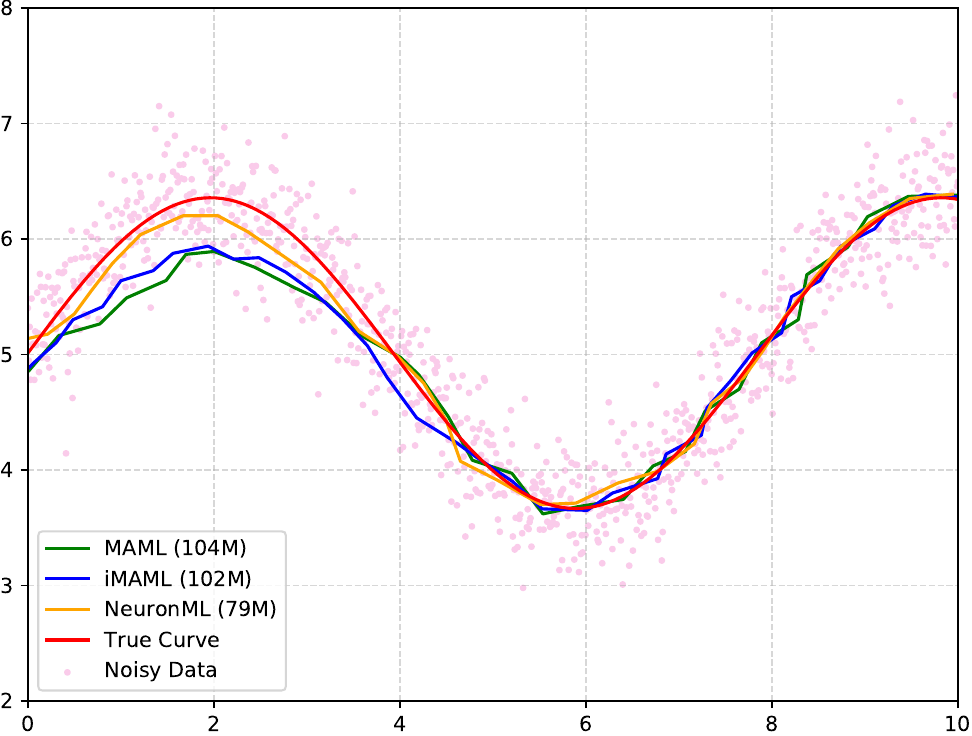}}
        \hfill
    \subfigure[Step=3, Data points=3]{
        \includegraphics[width=0.25\linewidth]{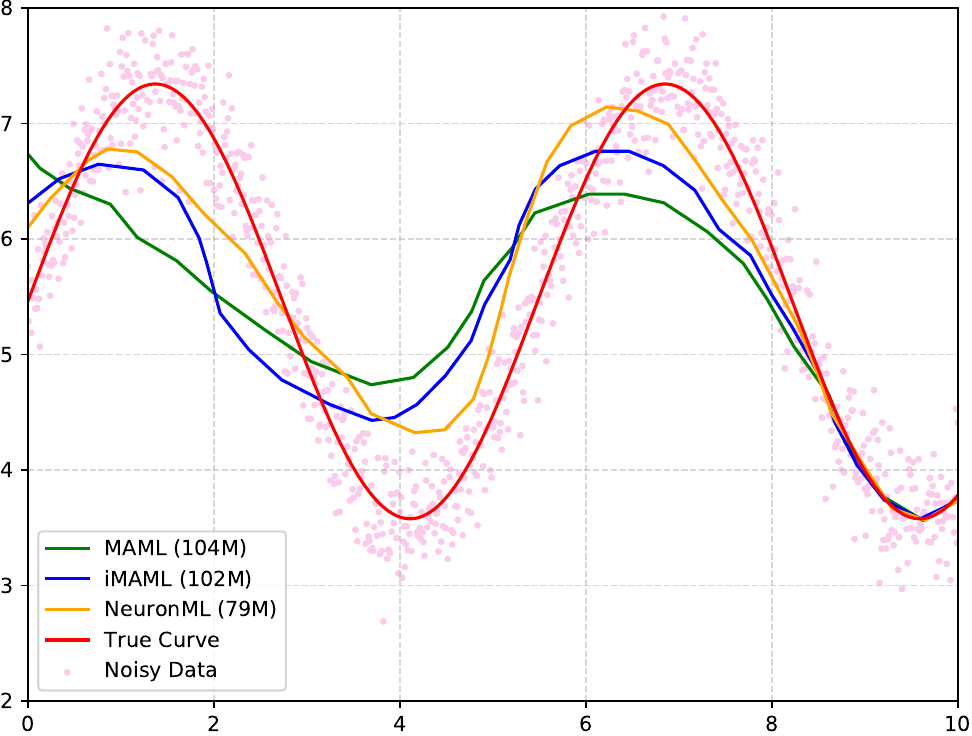}} 
        \hfill
    \subfigure[Step=3, Data points=5]{
        \includegraphics[width=0.25\linewidth]{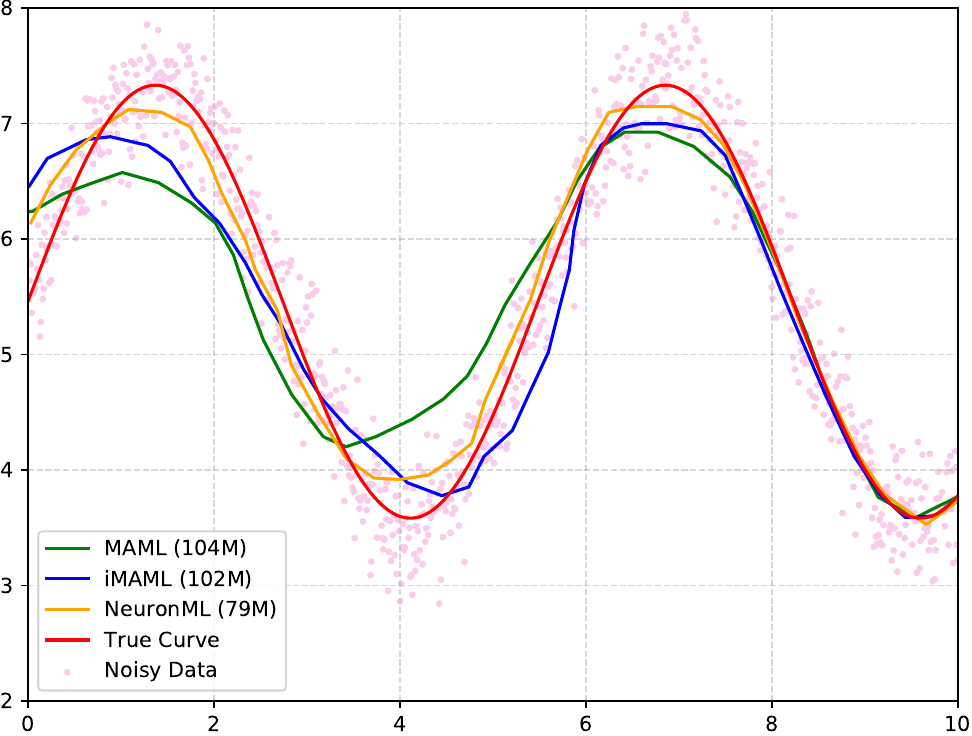}} 
        \hfill
    \subfigure[Step=3, Data points=8]{
        \includegraphics[width=0.25\linewidth]{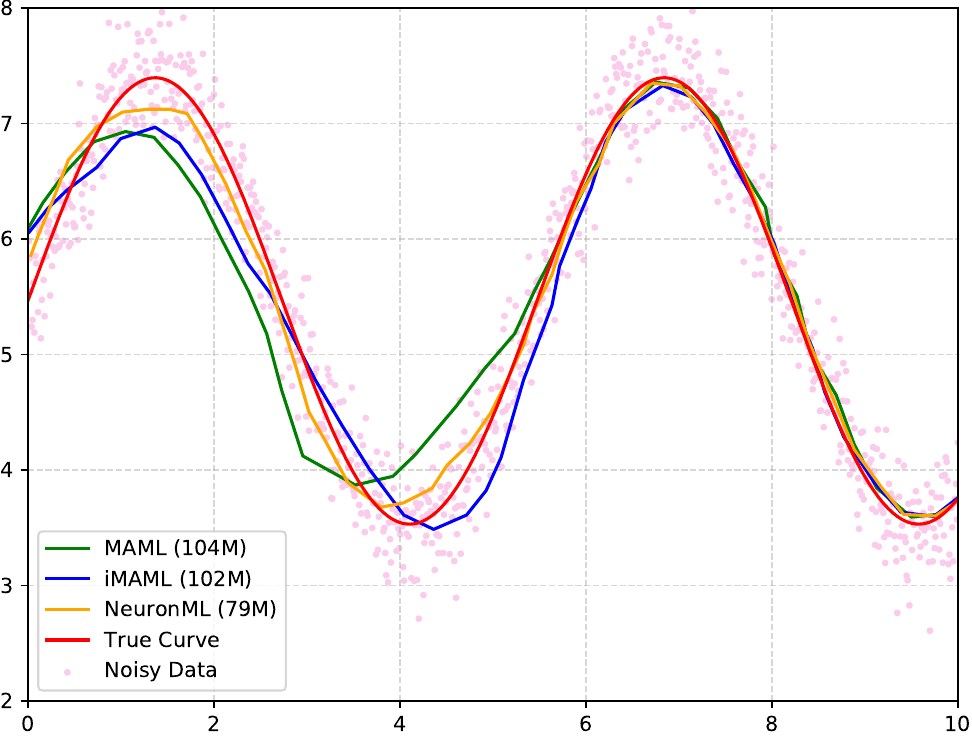}}
        \hfill
    \subfigure[Step=5, Data points=3]{
        \includegraphics[width=0.25\linewidth]{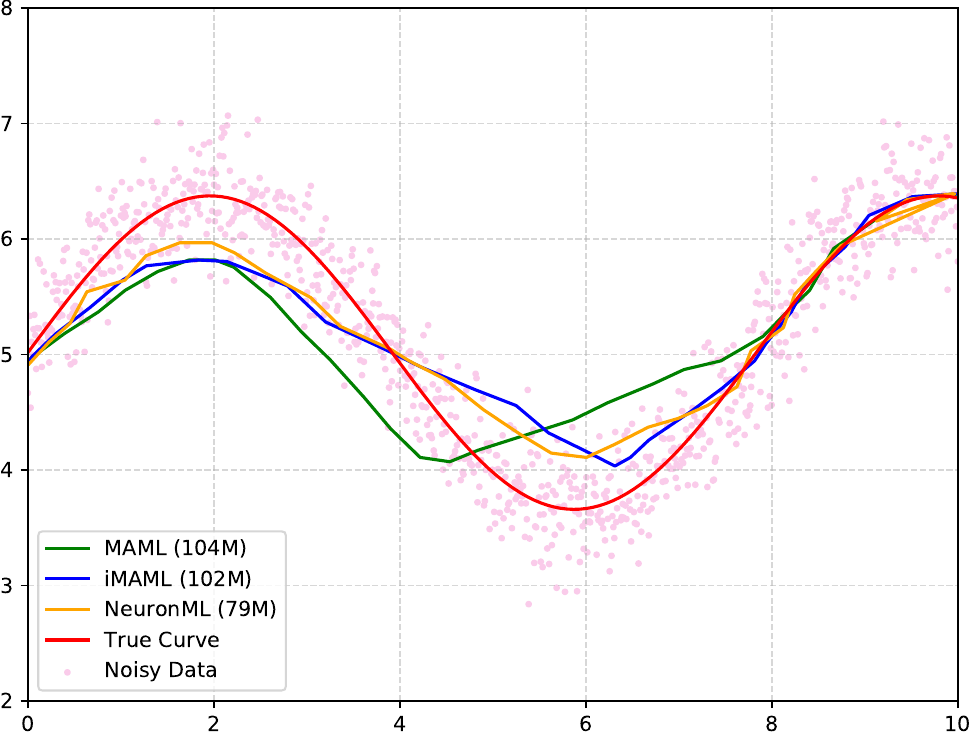}} 
        \hfill
    \subfigure[Step=5, Data points=5]{
        \includegraphics[width=0.25\linewidth]{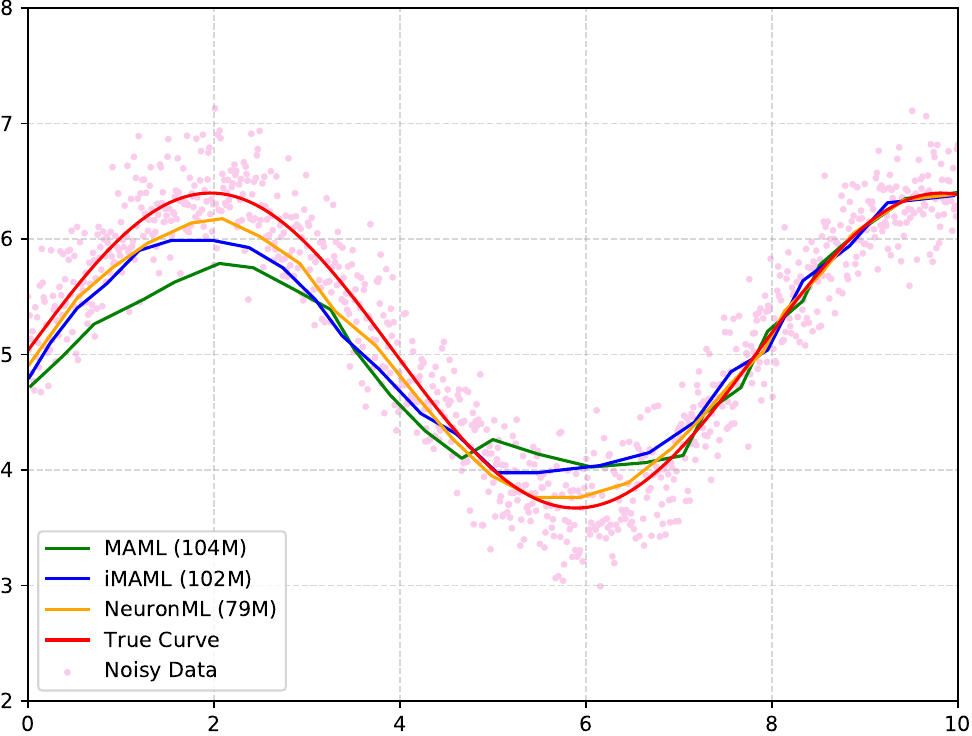}}
        \hfill
    \subfigure[Step=5, Data points=8]{
        \includegraphics[width=0.25\linewidth]{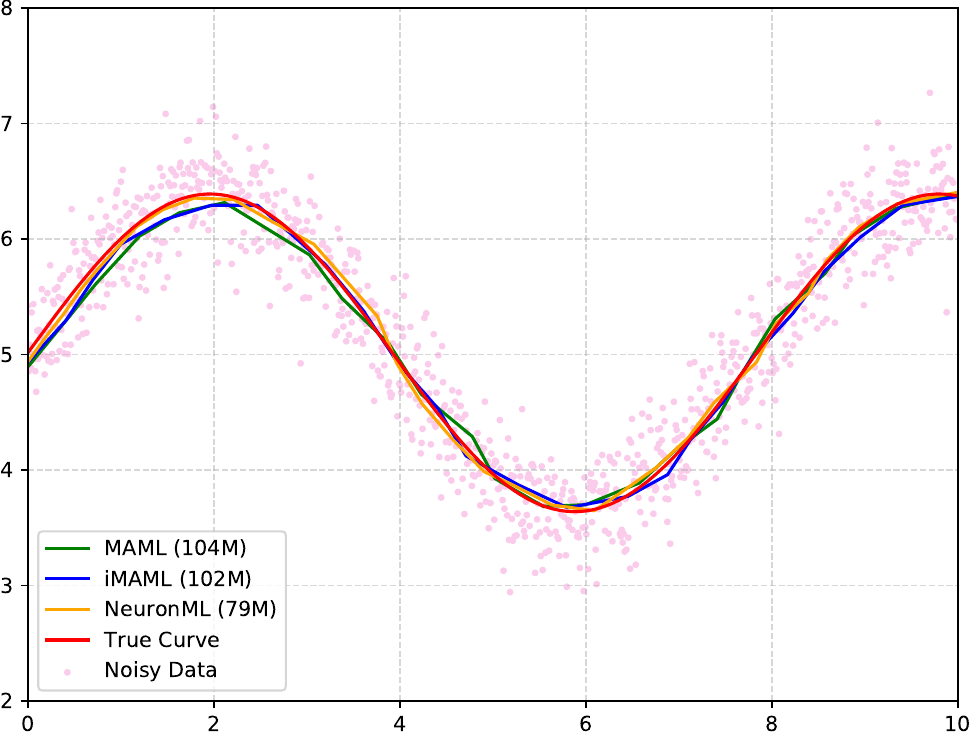}}
    \subfigure[Step=5, Data points=3]{
        \includegraphics[width=0.25\linewidth]{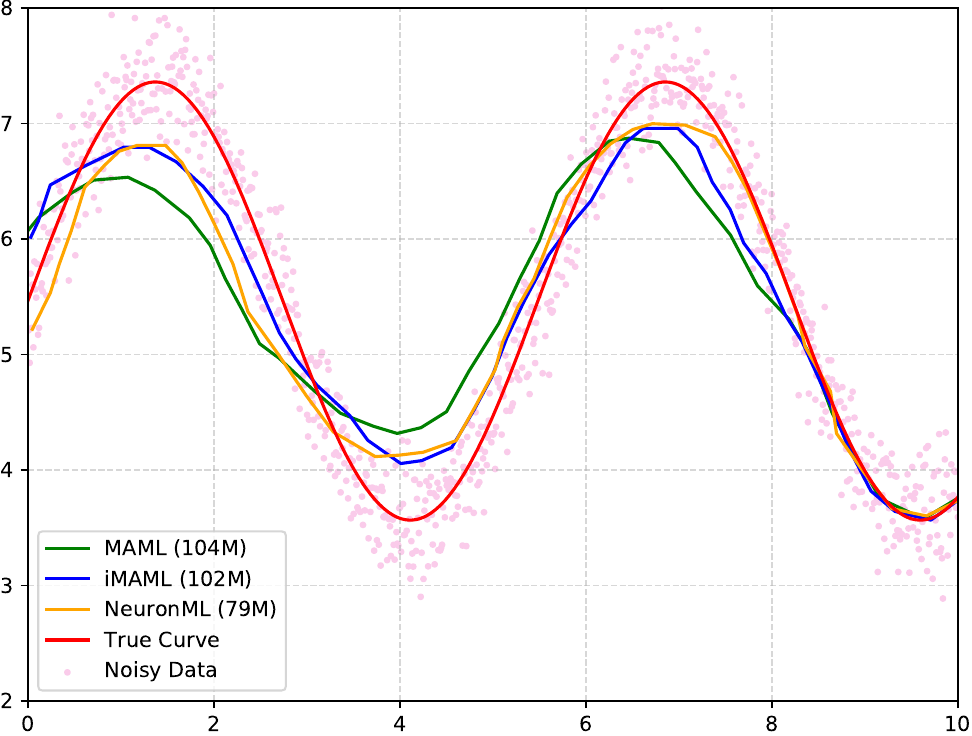}} 
        \hfill
    \subfigure[Step=5, Data points=5]{
        \includegraphics[width=0.25\linewidth]{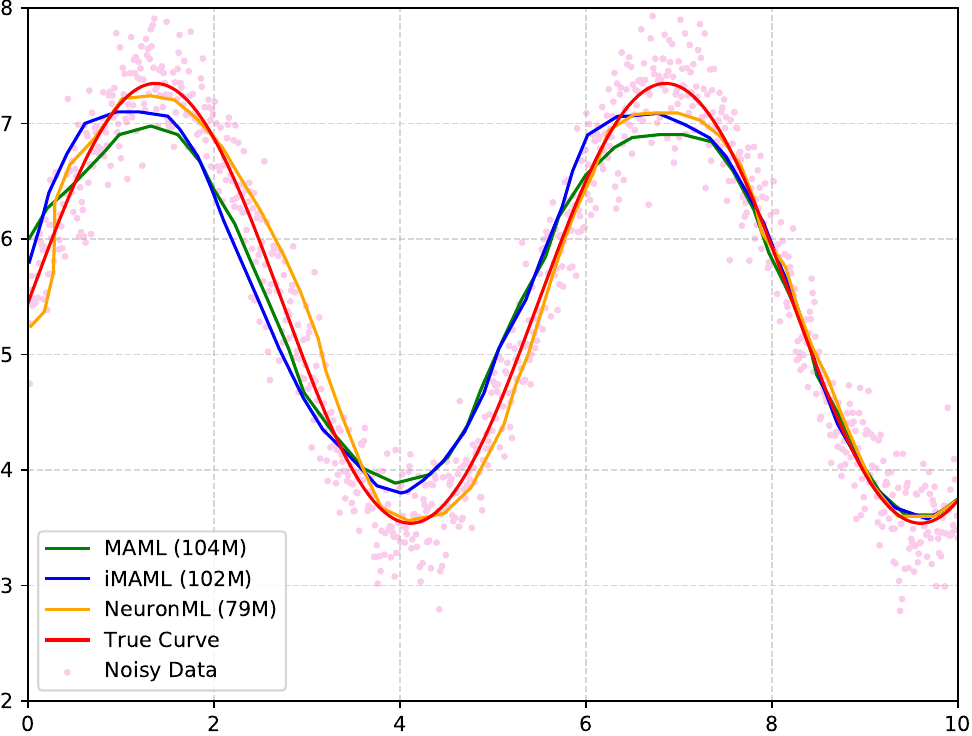}}
        \hfill
    \subfigure[Step=5, Data points=8]{
        \includegraphics[width=0.25\linewidth]{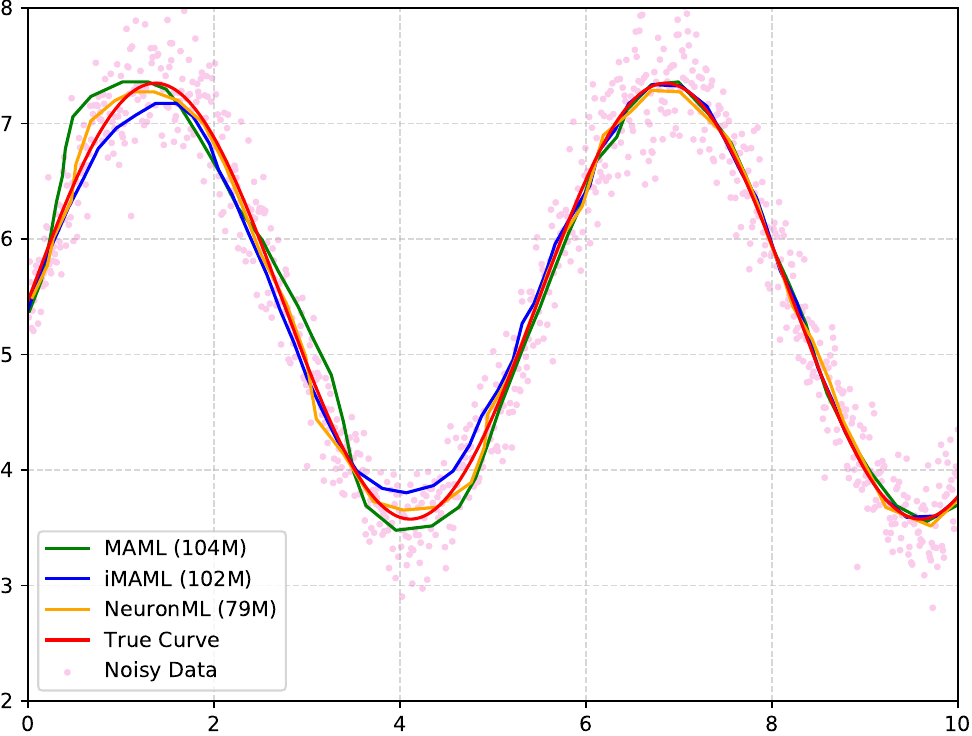}}
    \caption{Adaptation curves of different models (steps = 3, 6) on sinusoid regression. Left: the adaptation curves with 3 data points which are all located at 1/4 of the input range. Middle: the adaptation curves with 5 data points. Right: the adaptation curves with 10 data points. }
    \label{fig:app_ex1}
\end{figure*}

%%%%%%%%%%%%%%%%%%%%%%%%%%%%%%%%%%%%%%%%%%%%%%%%%%%%%%%%%%%%%%%%%

%%%%%%%%%%%%%%%%%%%%%%%%%%%%%%%%%%%%%%%%%%%%%%%%%%%%%%%%%%%%%%%%%
%\newpage
\section{Additional Details and Full Results for Regression}
\label{sec:app_3}

\subsection{Additional Details}
\label{sec:app_3_1}

\paragraph{Datasets}
For regression, we consider two benchmark dataset to illustrate the principles of NeuronML, i.e., sinusoid regression and pose prediction.
\begin{itemize}
    \item Sinusoid regression dataset \cite{jiang2022role}: It is a standard test in regression tasks for measuring the performance of machine learning models in understanding and forecasting periodic data. It features simulated sinusoidal waveforms with varying amplitudes, frequencies, and phases, tailored for meta-learning and few-shot learning scenarios. In this work, we conduct 680 tasks for training and testing, and each task involves regressing from the input to the output of a wave in the form of $A\sin w \cdot x + b + \epsilon $, where $A \in \left [ 0.1, 5.0 \right ] $, $w \in \left [ 0.5, 2.0 \right ]$, and $b \in \left [ 0,2\pi \right ] $. Thus, $\mathbb{P} (\mathcal{T})$ is continuous, and the amplitude and phase of the wave are varied between tasks. We add Gaussian noise with $\sigma = 0$ and $\epsilon = 0.3$ to each data point sampled from the target task.
    \item Pose prediction: we conduct tasks for training and testing from the Pascal 3D dataset \cite{xiang2014beyond}. Pascal 3D dataset \cite{xiang2014beyond}: It is a well-known resource in the computer vision community. It extends the Pascal VOC challenge, which is a benchmark for object detection and image segmentation, with 3D pose annotations for objects. This dataset provides a set of images with annotations that include 3D orientations of objects across a variety of categories, which can be used for a range of computer vision tasks, including 3D object reconstruction, pose prediction, and so on. We randomly select 50 objects for meta-training and 15 additional objects for meta-testing.
\end{itemize}

\paragraph{Baselines} We choose various classic and outstanding methods in few-shot regression as baselines for comparison, including MAML~\cite{maml}, Reptile \cite{reptile}, ProtoNet ~\cite{protonet}, Meta-AUG ~\cite{rajendran2020meta}, MR-MAML ~\cite{yin2019meta}, ANIL ~\cite{optimization_rapid}, MetaSGD ~\cite{Meta-sgd}, T-NET ~\cite{tnet}, TAML ~\cite{jamal2019task}, iMAML ~\cite{rajeswaran2019meta}. 

\paragraph{Experimental Setup} For fair comparison (settings of all the models), in regression experiments, we employ the Mean Squared Error (MSE) as the evaluation metric and adopt the ResNet-50 as the backbone for assessing model size. During training with NeuronML, we execute one gradient update with 10 examples, in line with~\cite{maml}, employing a fixed step size of $\beta=0.02$, 
and utilizing Adam as the optimizer (the ablation study of the optimization methods are provided in \textbf{Subsection \ref{sec:7.4}}, 
and utilizing Adam as the optimizer, all settings and experiments for this study are provided in the CODE). 

\subsection{Full Results}
\label{sec:app_3_2}
Besides the results provided in the main text, we assess performance by conducting fine-tuning on the models acquired through multiple methods, e.g., MAML, iMAML, and NeuronML, using different quantities of datapoints, specifically $\{3, 5, 8\}$ with different fine-tuning steps $\{3,6\}$. Throughout the fine-tuning process, each gradient step is calculated using the same data points. The results are provided in \textbf{Figure \ref{fig:app_ex1}}. As expected, NeuronML demonstrated excellent performance while illustrating its potential for inference.

%\newpage
\section{Additional Details and Full Results for Classification}
\label{sec:app_4}

\subsection{Additional Details}
\label{sec:app_4_1}

\paragraph{Datasets} For classification, we select various problems to evaluate the performance of NeuronML, i.e., standard few-shot learning (SFSL), cross-domain few-shot learning (CFSL), and multi-domain few-shot learning (MFSL). 
\begin{itemize}
    \item SFSL: We select four benchmark datasets for evaluation, i.e., miniImagenet \cite{miniImagenet}, Omniglot ~\cite{Omniglot}, tieredImagenet \cite{tieredImagenet}, and CIFAR-FS \cite{CIFAR-FS}. 
    Each dataset is described below: i) miniImagenet consists of 100 classes with 50,000/10,000 training/testing images, split into 64/16/20 classes for meta-training/validation/testing; ii) Omniglot contains 1,623 characters from 50 different alphabets; iii) tieredImagenet contains 779,165 images and 391/97/160 classes for meta-training/validation/testing; iv) CIFAR-FS contains 100 classes with 600 images per class, split into 64/16/20 classes for meta-training/validation/testing.
    
    \item CFSL: We select two benchmark datasets, i.e., CUB ~\cite{cub} and Places \cite{places}. Each dataset is described below: i) CUB contains 11,788 images of 200 categories, split into 100/50/50 classes for meta-training/validation/testing; ii) Places contains more than 2.5 million images across 205 categories, split into 103/51/51 classes for meta-training/validation/testing. 
    The models in this experiment are trained on miniImagenet, and evaluated on CUB and Places.

    \item MFSL: We choose the challenging Meta-dataset ~\cite{metadataset}. Meta-Dataset stands as a comprehensive benchmark for few-shot learning, encompassing 10 datasets spanning various domains. It is intentionally crafted to emulate a more realistic scenario by avoiding restrictions on fixed ways and shots for few-shot tasks. The dataset comprises 10 diverse domains, with the initial 8 in-domain (ID) datasets, namely ILSVRC, Omniglot, Aircraft, Birds, Textures, Quick Draw, Fungi, and VGG Flower, utilized for meta-training. The remaining 2 datasets, Traffic Signs and MSCOCO, are reserved for evaluating out-of-domain (OOD) performance.
\end{itemize}
Specifically, we entails the swift acquisition of knowledge in $N$-way classification tasks, using either 1 or 5 shots. The procedure is established as such: we select $N$ classes that have not been previously encountered, introduce the model to $K$ examples from each of these classes, and then evaluate the model's ability to accurately classify new samples from the $N$ selected classes. As for the architecture of the classifier, we employ a variety of networks such as Conv4, ResNet, and DenseNet to serve as the foundational backbones.

\paragraph{Baselines} 
We choose various classic and outstanding methods in few-shot classification for comparison, covering all types of meta-learning methods mentioned in \textbf{Section \ref{sec:2}}. For standard few-shot learning, the comparison baselines include MAML~\cite{maml}, ProtoNet ~\cite{protonet}, Meta-AUG ~\cite{rajendran2020meta}, MR-MAML ~\cite{yin2019meta}, ANIL ~\cite{optimization_rapid}, MetaSGD ~\cite{Meta-sgd}, T-NET ~\cite{tnet}. For cross-domain few-shot learning and multi-domain few-shot learning, the comparison baselines are based on standard few-shot learning and additionally introduce SCNAP ~\cite{bateni2020improved}, MetaQDA ~\cite{zhang2021shallow}, Baseline++ ~\cite{chen2019closer}, S2M2 ~\cite{mangla2020charting}, MatchingNet ~\cite{vinyals2016matching}, CNAPs ~\cite{requeima2019fast}.

\paragraph{Experimental Setup} For fair comparison (settings of all the models), in classification experiments, the computation of each gradient involves a batch size of $N_i^s$ examples. Specifically, for Omniglot, both the 5-way convolutional and non-convolutional MAML models underwent training with 1 gradient step, utilizing a step size of $0.4$ and a meta batch size of 32 tasks. The evaluation of the network was conducted with 3 gradient steps, maintaining the same step size of $0.4$. In the case of the 20-way convolutional MAML model, training and evaluation were performed with 5 gradient steps and a step size of $0.1$. Throughout the training process, the meta batch size was configured to be 16 tasks. For miniImagenet, both models were trained with 5 gradient steps of size $0.01$ and evaluated using 10 gradient steps during test time, and 15 examples per class were employed for evaluating the post-update meta-gradient. The meta batch size was set to 4 and 2 tasks for 1-shot and 5-shot training, respectively. 
% Similarly to Adam, our BDA uses an adaptive learning rate, which maintains an adaptive learning rate for each parameter and adjusts the learning rate based on the historical gradient information of the parameters to better adapt to the training process of the model. 
All models underwent training for 30000 iterations, utilizing NVIDIA GeForce RTX 4090s.

\begin{figure*}[t]
\centering
    \subfigure[Ant-Goal]{
        \includegraphics[width=0.28\linewidth]{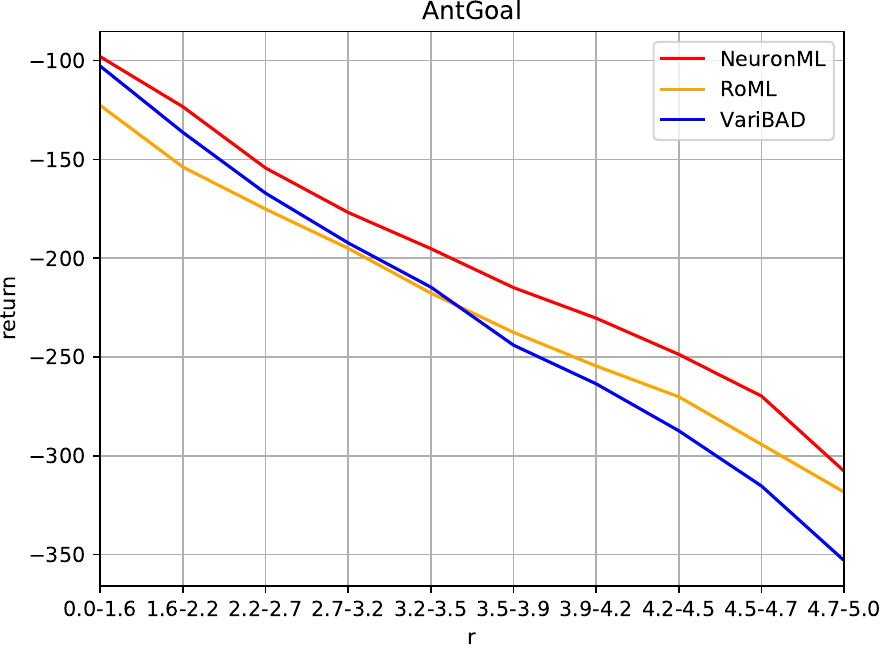}}
        \hfill
    \subfigure[AntAll-Mass]{
        \includegraphics[width=0.28\linewidth]{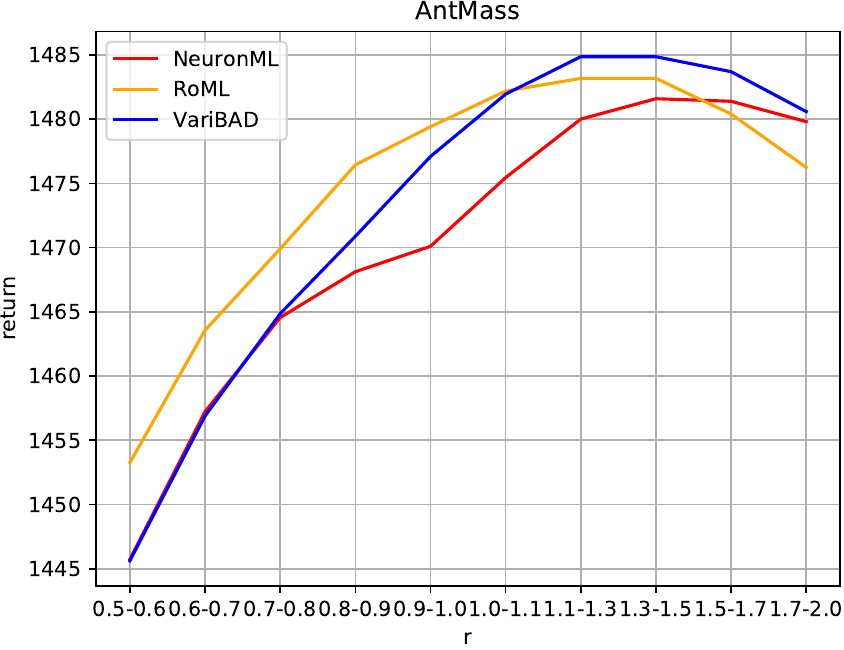}}
        \hfill
    \subfigure[HumanoidAll-Mass]{
        \includegraphics[width=0.28\linewidth]{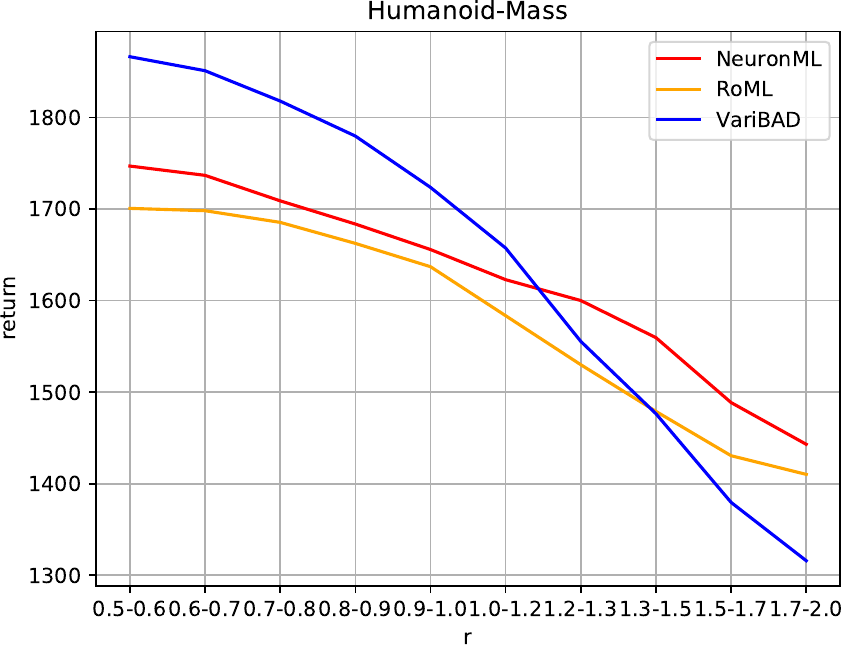}}
        \hfill
    \subfigure[HumanoidBody-damping]{
        \includegraphics[width=0.28\linewidth]{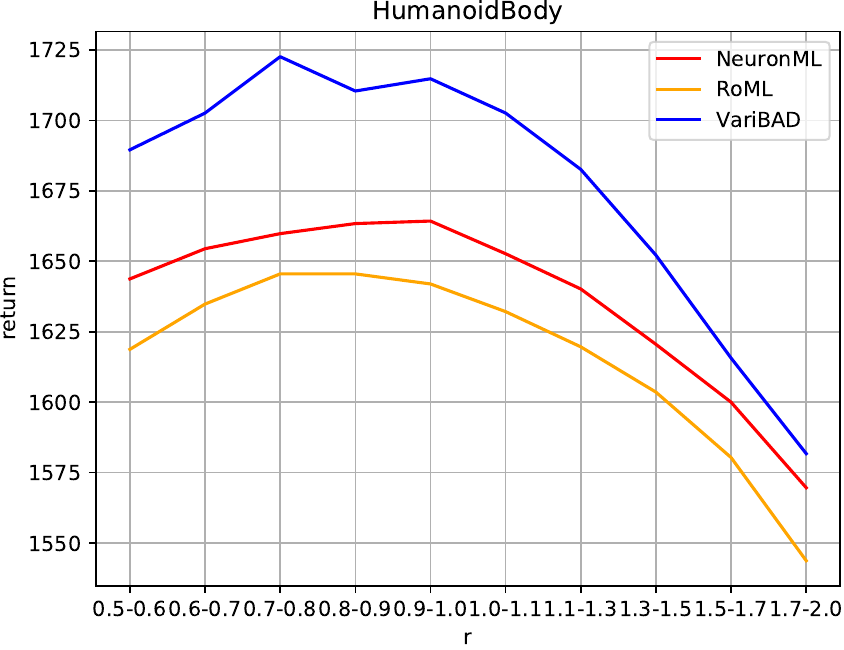}} 
        \hfill
    \subfigure[HumanoidBody-Mass]{
        \includegraphics[width=0.28\linewidth]{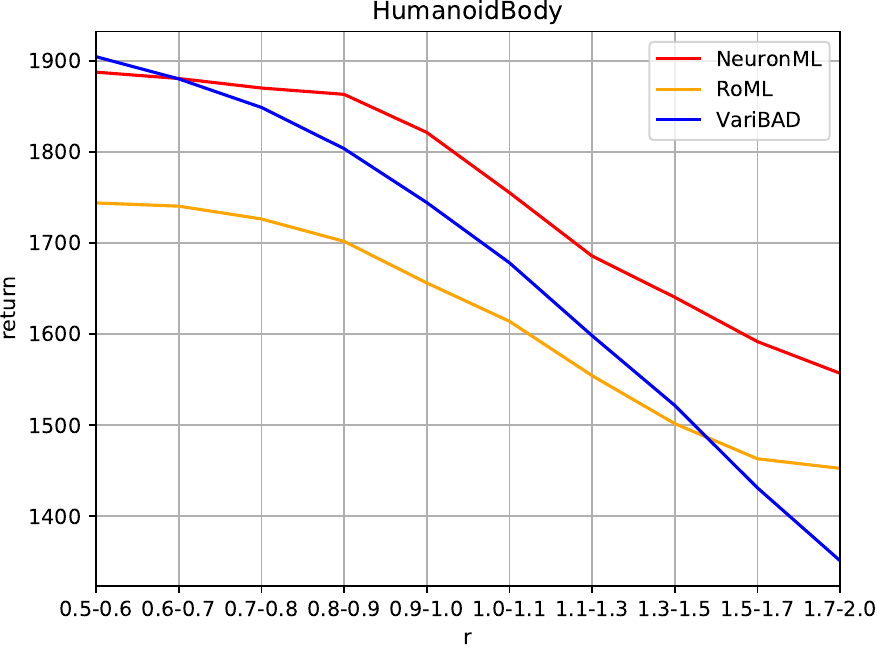}} 
        \hfill
    \subfigure[HumanoidVel]{
        \includegraphics[width=0.28\linewidth]{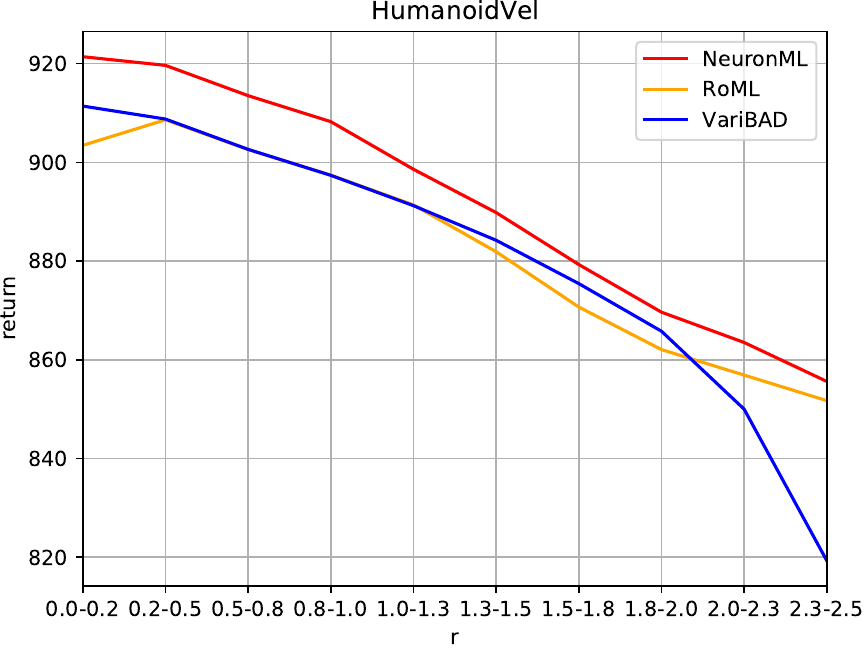}}
        \hfill
    \subfigure[HalfCheetahBody-Damping]{
        \includegraphics[width=0.28\linewidth]{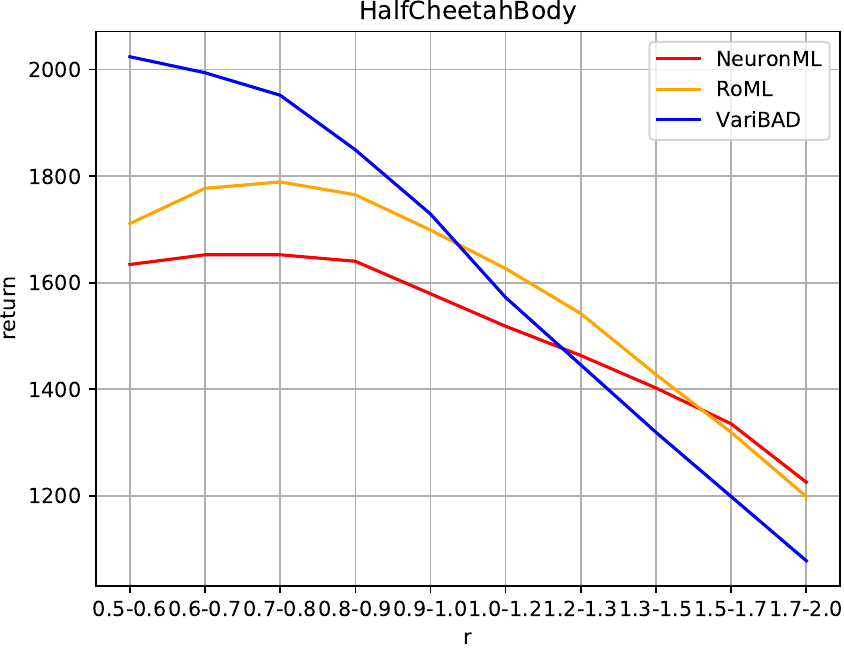}} 
        \hfill
    \subfigure[HalfCheetahBody-Mass]{
        \includegraphics[width=0.28\linewidth]{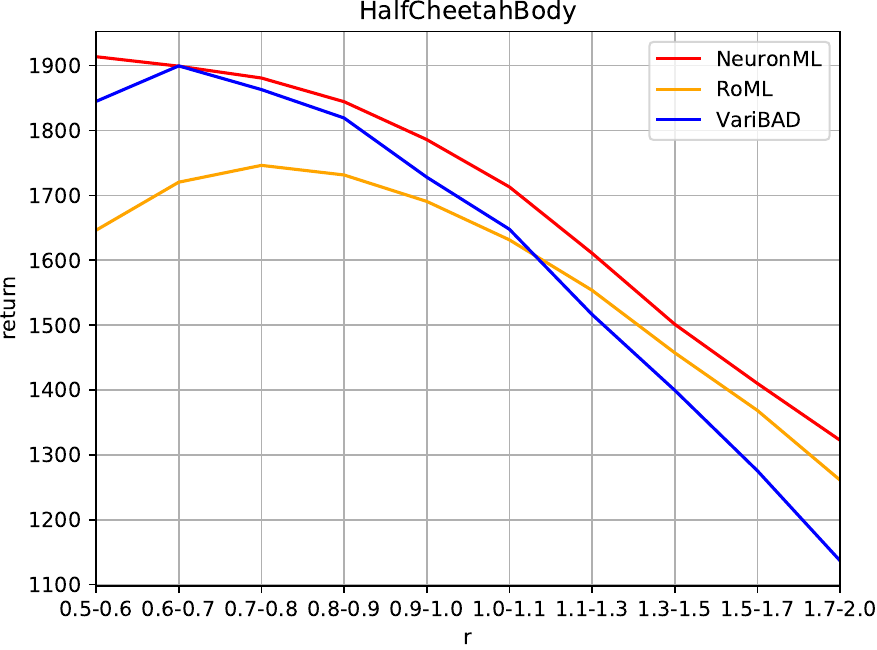}}
        \hfill
    \subfigure[HalfCheetahVel]{
        \includegraphics[width=0.28\linewidth]{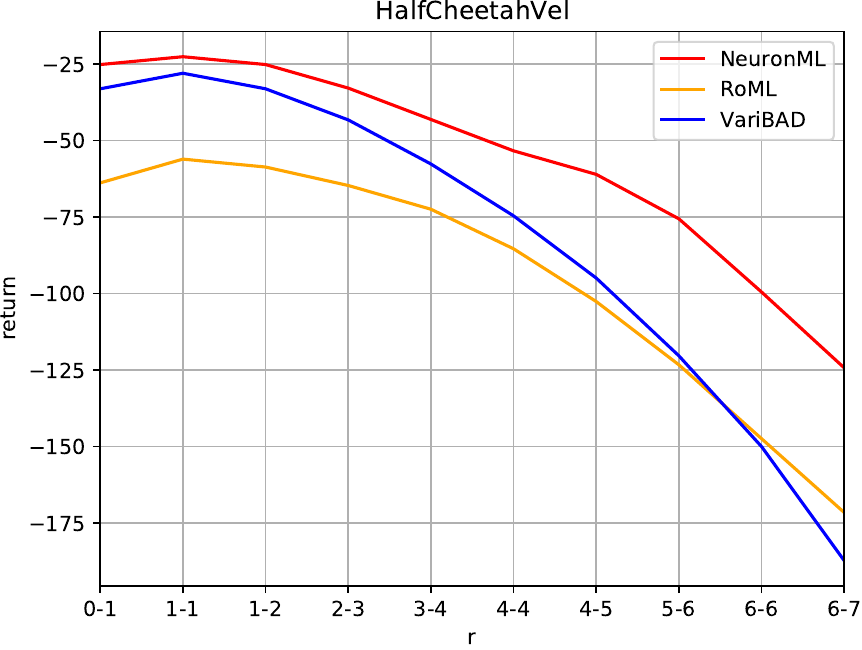}}
    \caption{Average return per range of tasks in the various MuJoCo environments (global average in parentheses). The Mass results (HalfCheetah-Mass, Hum-Mass, and Ant-Mass) are shown in \textbf{Figure \ref{fig:ex3_1}}. 
    Note that in this visualization experiment, we set the upper limit of NeuronML’s training time to 1/2 of other frameworks.
    }
    \label{fig:app_rl}
\end{figure*}

%\newpage
\section{Additional Details and Full Results for Reinforcement Learning}
\label{sec:app_5}

\subsection{Additional Details}
\label{sec:app_5_1}

\paragraph{Datasets} To evaluate the performance of NeuronML on reinforcement learning problems, we adopt two benchmark environments to build the task, i.e., Khazad Dum ~\cite{greenberg2023train} and MuJoCo~\cite{mujoco}. 
\begin{itemize}
    \item Khazad Dum: In the Khazad Dum scenario, NeuronML's performance is assessed within a stochastic discrete action space. Here, the agent commences its journey from a randomly assigned point at the bottom left of the map, aiming to reach a green target zone while avoiding descent into the perilous black chasm. Throughout its traversal, the agent is subjected to motion noise due to simulated rainfall, following an exponential distribution with a hazard rate of 0.01. In the deployment, the agent action step size is fixed to 2 and trained based on $5\cdot10^6$ frames.
    
    \item MuJoCo: This dataset is used to assess NeuronML's capabilities in environments characterized by high-dimensional and continuous action spaces, including HalfCheetah, Humanoid, and Ant. Each of these environments is designed with three distinct tasks: (i) Velocity, targeting various positions or speeds; (ii) Mass, focusing on different body weights; and (iii) Body, involving variations in mass, head size, and physical damping. In addition, within the HalfCheetah environment, we have randomly selected 10 numerical variables from \verb|env.model| and allowed their values to fluctuate across tasks as per the methodology described in ~\cite{greenberg2023train}. These specific datasets are referred to as HalfCheetah 10D-task {a,b,c}.
\end{itemize}

\paragraph{Baselines} 
We choose various outstanding (SOTA) methods in reinforcement learning as baselines for comparison, including CESOR ~\cite{cesor}, PAIRED ~\cite{PAIRED}, VariBAD ~\cite{varibad}, and PEARL ~\cite{pearl}. Meanwhile, we implement NeuronML following the settings of RoML and CVaR-ML mentioned in ~\cite{greenberg2023train}, which are extensions to distinct risk-neutral meta-reinforcement learning baselines.

\paragraph{Experimental Setup}
For Khazad Dum, we follow the setting of ~\cite{greenberg2023train}, where each task is conducted over $K = 4$ episodes, each consisting of $T = 32$ time steps. The return is computed as the undiscounted sum of rewards. At each time step, a cost of $1/T$ points is incurred if the L1 distance between the agent and the target exceeds 5. For distances between 0 and 5, the cost linearly varies between 0 and $1/T$. Upon reaching the destination, the agent receives a reward of $5/T$ and incurs no further costs for the remainder of the episode. However, if the agent falls into the abyss, it becomes incapable of reaching the goal and must endure a cost of $1/T$ for every subsequent step until the episode concludes. During each step, the agent observes its location (represented through a soft one-hot encoding) and decides whether to move left, right, up, or down. If attempting to move into a wall, the agent remains in place. In all the MuJoCo benchmarks introduced in \textbf{Subsection \ref{sec:7.3}} of the main text, each task’s meta-rollout consists of 2 episodes × 200 time-steps per episode. For the experiments, we utilize the official implementations of VariBAD, PEARL, RoML, and CVaR-ML. The NeuronML model is built upon these baselines, and their execution times are indiscernible from the baseline methods. All experiments were conducted on machines equipped with NVIDIA V100 GPUs. The duration for each experiment (both meta-training and testing) ranged from 8 to 106 hours, contingent on the specific environment and baseline algorithm employed.

\subsection{Full Results}
\label{sec:app_5_2}
\textbf{Figure \ref{fig:ex3_1}} of the main text shows the performance of NeuronML on Khazad Dum and MoJoCo, which compares the test return of the model with state-of-the-art (SOTA) and multiple outstanding baselines. 
For the MoJoCo environment, the full results of the reward curves of the model in a variety of different scenarios are provided in \textbf{Figure \ref{fig:app_rl}}.
It is worth noting that our work achieves a lower parameter size compared to other baselines. 
Combined with \textbf{Table \ref{tab:ex3_1}} in the main text, 
we can get the reward curve of NeuronML is the same as or even smoother than that of RoML under multiple tasks, which shows that the NeuronML model is robust and performs better on high-risk tasks.
% ghj-we can get: (i) NeuronML achieves similar results to SOTA work with a smaller model size. (ii) The reward curve of NeuronML is the same as or even smoother than that of RoML under multiple tasks, which shows that the NeuronML model is robust and performs better on high-risk tasks.

\end{appendices}

\end{document}